\documentclass[journal]{IEEEtran}
\newcommand{\method}{VISTA-Net}

\usepackage[pagebackref=true,breaklinks=true,letterpaper=true,colorlinks,bookmarks=false,urlcolor=red]{hyperref}
\usepackage{url}
\usepackage{graphicx}
\usepackage{graphics}
\usepackage{color}
\usepackage{mathtools}
\usepackage{multirow}
\usepackage{subfigure}
\usepackage{tabularx}
\usepackage{wrapfig,lipsum,booktabs}
\usepackage{xcolor}
\usepackage{hyperref}
\usepackage{floatflt}
\usepackage{wrapfig,lipsum,booktabs}
\usepackage{amssymb}
\usepackage{caption}
\usepackage{setspace}

\usepackage[ruled, lined, linesnumbered, commentsnumbered, longend]{algorithm2e}
\usepackage{algpseudocode}

\def\eg{\textit{e.g.}}
\def\ie{\textit{i.e.}}

\def\etal{\textit{et al.}}

\newcommand{\vect}[1]{\mathbf{#1}}

\begin{document}

%
\title{{Variational Structured Attention Networks for \\Deep Visual Representation Learning}}
%
%
%

\author{Guanglei Yang,
        Paolo Rota,
        Xavier Alameda-Pineda,~\IEEEmembership{Senior Member,~IEEE,}
        Dan Xu,
        Mingli Ding$*$,
        Elisa Ricci,~\IEEEmembership{Member,~IEEE}
\thanks{Guanglei Yang and Mingli Ding are with School of Instrument Science and Engineering, Harbin Institute of Technology (HIT), Harbin, China. E-mail:\{yangguanglei,dingml\}@hit.edu.cn. $*$Corresponding author.\protect}
\thanks{Xavier Alameda-Pineda is with the RobotLearn Team, INRIA. E-mail: xavier.alameda-pineda@inria.fr.\protect}
\thanks{Dan Xu is with the Department of Computer Science and Engineering, Hong Kong University of Science and Technology. E-mail: danxu@cse.ust.hk.\protect}
\thanks{{Paolo Rota and Elisa Ricci} are with the Department of Information Engineering and Computer Science, University of Trento, Italy. E-mail:  \{paolo.rota, e.ricci\}@unitn.it.\protect} 
\thanks{{Elisa Ricci} is with Deep Visual Learning group at Fondazione Bruno Kessler, Trento, Italy. \protect}
}

%
%

\markboth{IEEE Transactions on Image Processing}%
{Shell \MakeLowercase{\textit{et al.}}: Variational Structured Attention Networks for Visual Dense Representation Learning}
%



\maketitle

\begin{abstract}
Convolutional neural networks have enabled major progresses in addressing  
pixel-level prediction tasks such as semantic segmentation, depth estimation, surface normal prediction and so on, benefiting from their powerful capabilities in visual representation learning. Typically, state of the art models 
integrate attention mechanisms for improved deep feature representations. Recently, some works have demonstrated the significance of learning and combining both spatial- and channel-wise attentions for deep feature refinement. In this paper, we aim at effectively boosting previous approaches and propose a unified deep framework to jointly learn both spatial attention maps and channel attention vectors in a principled manner so as to structure the resulting attention tensors and model interactions between these two types of attentions. Specifically, we integrate the estimation and the interaction of the attentions within a probabilistic representation learning framework, leading to VarIational STructured Attention networks (\method{}). We implement the inference rules within the neural network, thus allowing for end-to-end learning of the probabilistic and the CNN front-end parameters.
As demonstrated by our extensive empirical evaluation on six large-scale datasets  for dense visual prediction, \method~outperforms the state-of-the-art in multiple continuous and discrete prediction tasks, thus confirming the benefit of the proposed approach in joint structured spatial-channel attention estimation for deep representation learning. The code is available at \url{https://github.com/ygjwd12345/VISTA-Net}. 

\end{abstract}

\begin{IEEEkeywords}
probabilistic deep representation learning, semantic segmentation, depth prediction, surface normal estimation
\end{IEEEkeywords}

%
\IEEEpeerreviewmaketitle

\section{Introduction}

\begin{figure*}
    \begin{minipage}{0.22\textwidth}
    \centering
    \includegraphics[width=\textwidth]{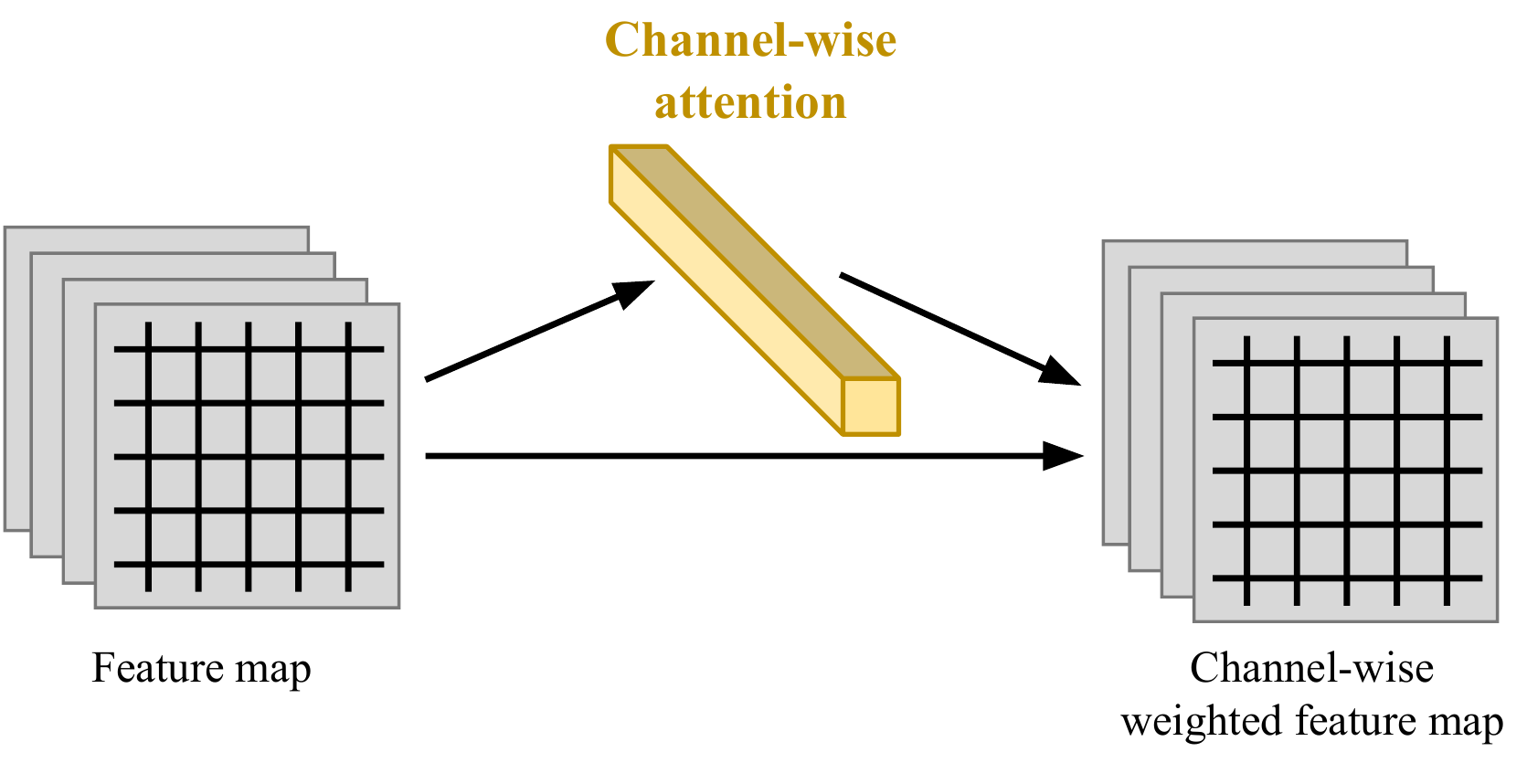}\\
    {\scriptsize (a) Channel-wise attention.}\vspace{3mm}\\
    \includegraphics[width=\textwidth]{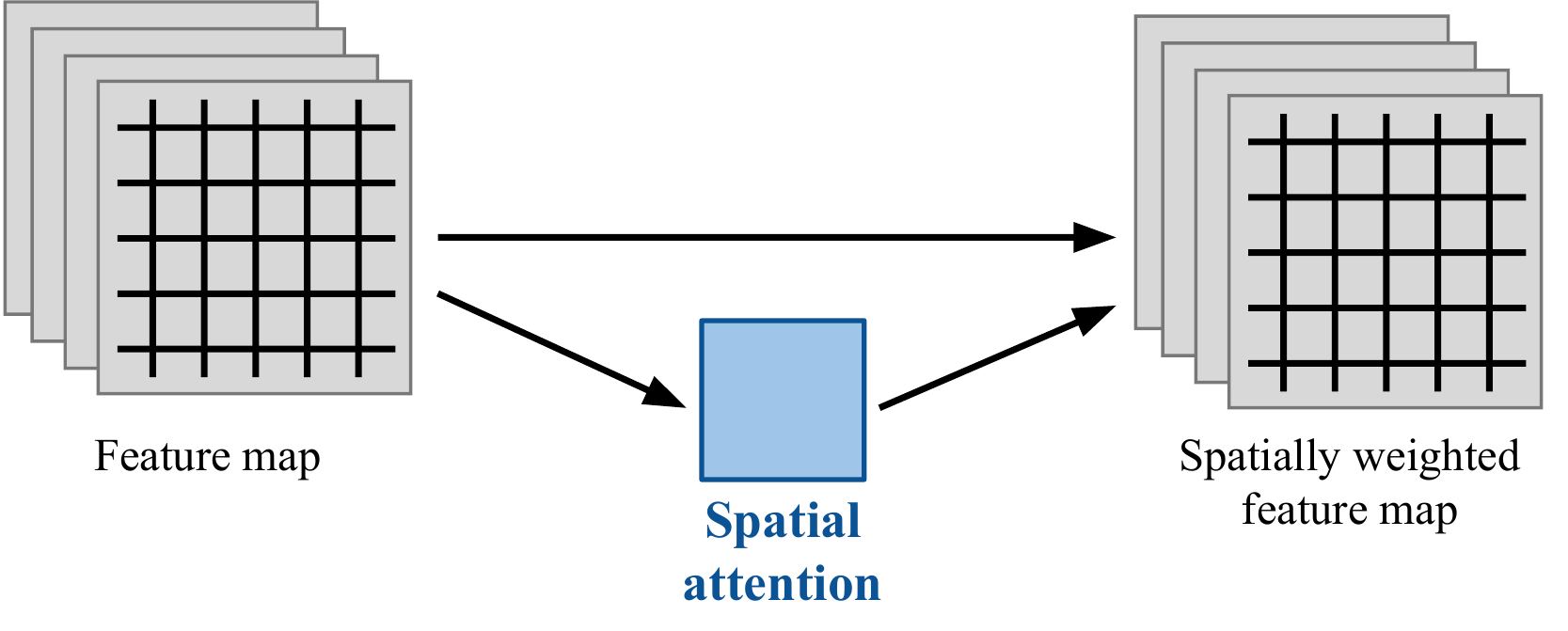}\\
    {\scriptsize (b) Spatial attention.}
    \end{minipage}\hfill
    \begin{minipage}{0.28\textwidth}
    \centering
    \includegraphics[width=\textwidth]{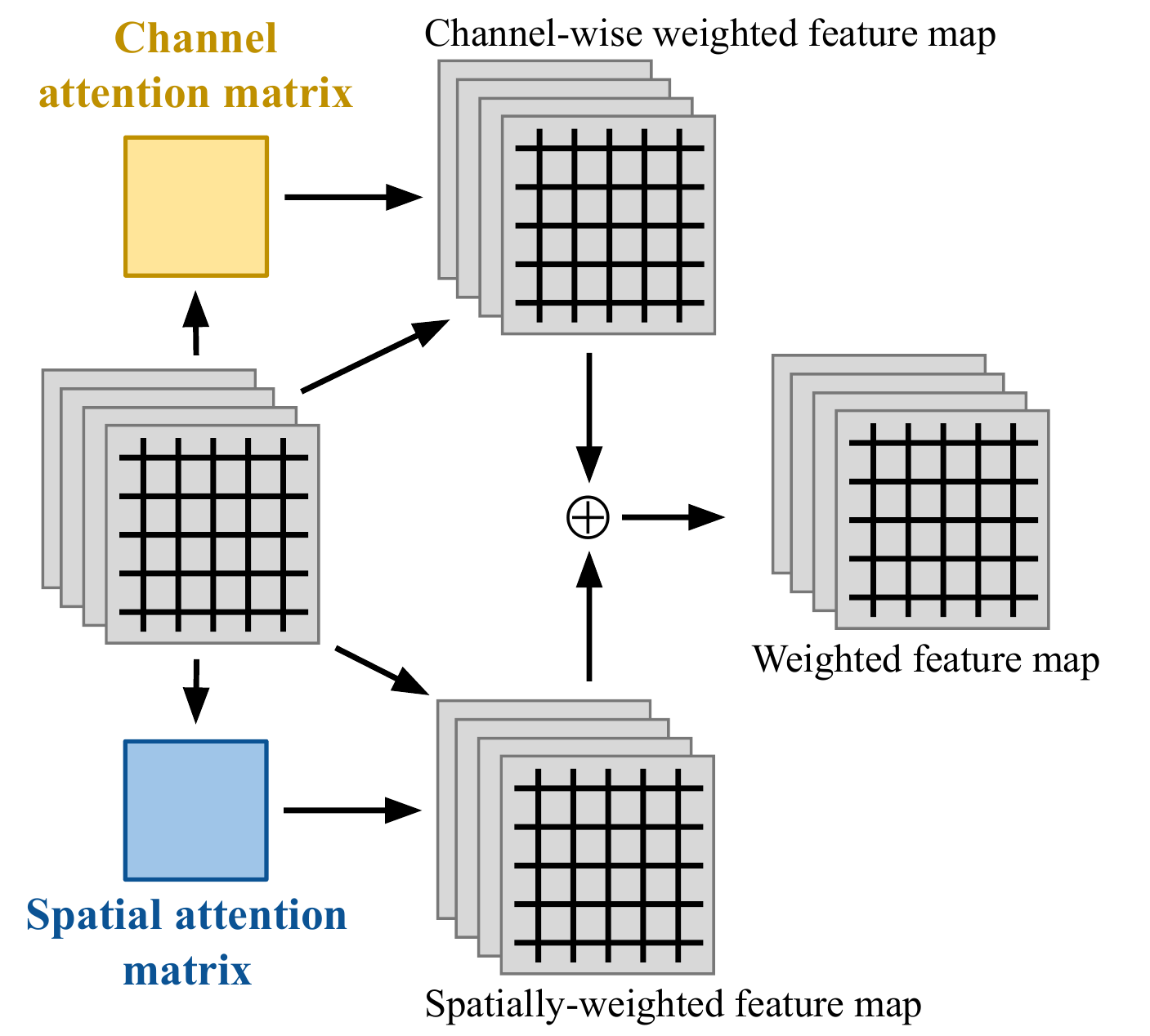}\\
    {\scriptsize (c) Separate spatial and channel attention.}
    \end{minipage}\hfill
    \begin{minipage}{0.32\textwidth}
    \centering
    \includegraphics[width=\textwidth]{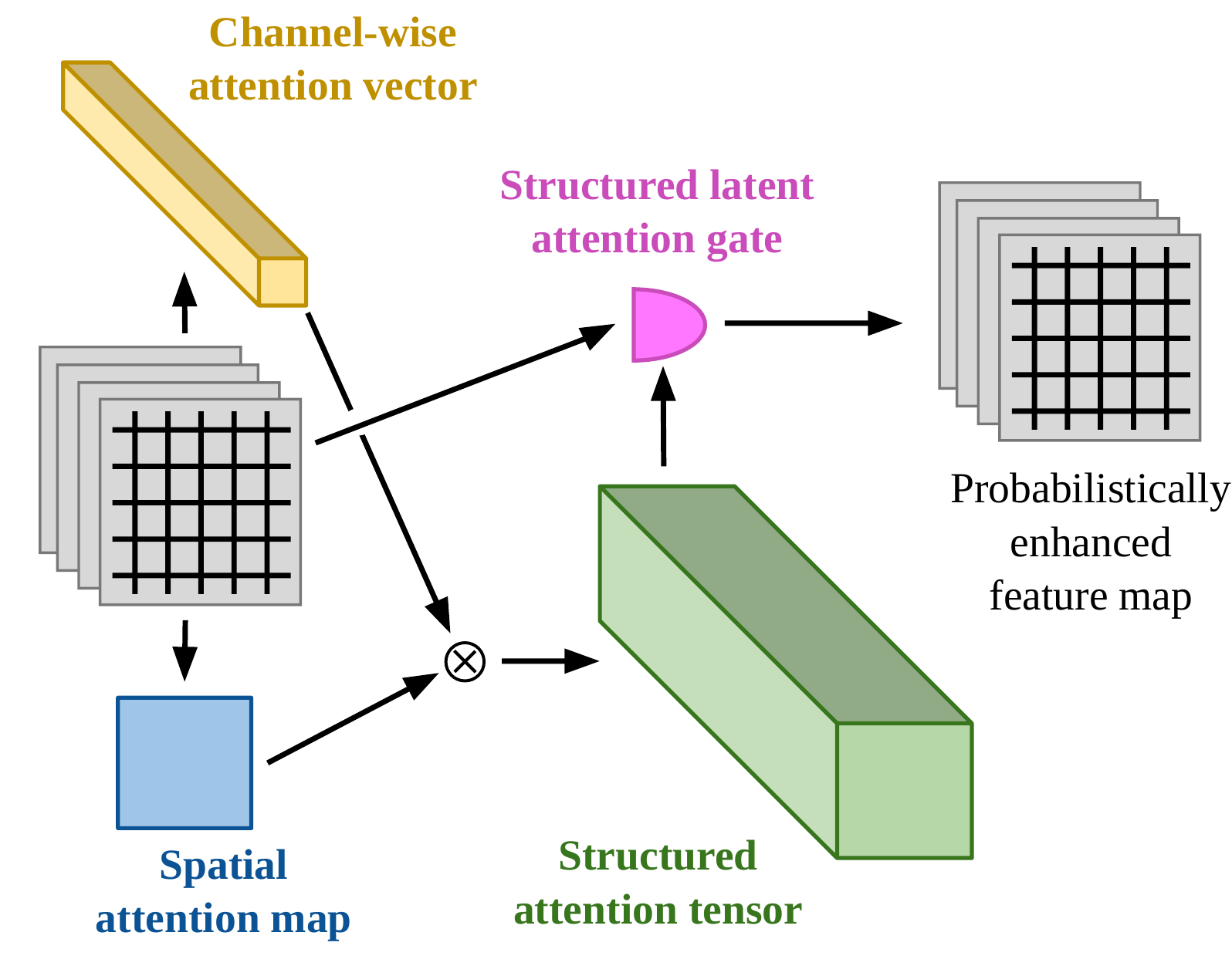}\\
    {\scriptsize (d) Proposed structured attention.}
    \end{minipage}
    \caption{Different attention mechanisms {for deep representation learning}. (a) and (b) correspond to channel-only and spatial-only attention, respectively. (c) corresponds to previous works \cite{fu2019dual} {combining the spatial- and channel-wise attended representations via} {simply applying element-wise addition operation $\oplus$ on} a spatial and a channel tensor. (d) shows the attention mechanism of \method: a channel-wise vector and a spatial map are estimated then tensor-multiplied ($\otimes$) yielding a structured  attention tensor. {The attention tensor acts as a structured latent gate producing a probabilistically enhanced feature map. Attention gates,  introduced in \cite{minka2009gates}, are latent variables which control the message passing in a probabilistic graphical model.}}
    \label{fig:teaser}
\end{figure*}

\IEEEPARstart{O}{ver} the past decade, convolutional neural networks (CNNs) have become the privileged methodology to address computer vision tasks requiring dense pixel-wise prediction, {such as semantic segmentation~\cite{chen2016attention,fu2019dual,ding2021looking}, video segmentation~\cite{wang2020paying,ji2021full}, human  parsing~\cite{wang2019learning,li2020self}, monocular depth prediction ~\cite{liu2015deep,roymonocular,yang2021transformer}, contour detection \cite{xu2017learningdeep} and normal surface computation \cite{eigen2014depth}}.
Recent studies provided clear evidence that attention mechanisms~\cite{mnih2014recurrent} within deep networks are undoubtedly a crucial factor in improving the performance ~\cite{chen2016attention,xu2017learningdeep,fu2019dual,zhan2018unsupervised}, {due to their remarkable effectiveness in enhancing the deep representation learning process.} In particular, previous works 
demonstrated that deeply learned attentions acting as soft weights to interact with different deep features at each channel {\cite{zhong2020squeeze,zhang2018context,song2020channel}} and at each pixel location {\cite{li2020spatial,johnston2020self,tay2019aanet}} permits to improve the pixel-wise prediction accuracy (see Fig.\ref{fig:teaser}.(a) and Fig.\ref{fig:teaser}.(b)). 
Recently, Fu et al. \cite{fu2019dual} proposed the Dual Attention Network (DANet), embedding in a fully convolutional network (FCN) two complementary attention modules, specifically conceived to model separately the semantic dependencies associated to the spatial and to the channel dimensions (Fig.\ref{fig:teaser}.(c)). 

Concurrently, other approaches have considered the use of attention models integrated within a graph network framework~\cite{zhang2020dynamic,chen2019graph,xu2017learningdeep}, showing the empirical advantage of adopting a graphical model to effectively capture the structured information present in the hidden layers of the neural network and thus enabling the learning of better deep feature representations.  
{Notably, Xu et al.~\cite{xu2017learningdeep,xu2020probabilistic} first introduced attention-gated conditional random fields (AG-CRFs), a convolutional {neural network implementing a probabilistic graphical model} that considers latent attention variables, denoted as gates and previously introduced in~\cite{minka2009gates}, in order to learn improved deep features and effectively fuse multi-scale information. However, their attention model is \emph{only} learned at the spatial level, while channel-wise dependencies are not accounted in their model.}

{In this paper we propose to combine these two lines of research by introducing a novel approach for (i) learning deep representations within a probabilistic framework and for (ii) jointly accounting for spatial- and channel-level dependencies (Fig.\ref{fig:teaser}.(d)). In particular, differently from~\cite{fu2019dual}, we demonstrate the benefit of a probabilistic formulation, integrating attention as latent variables in a graphical model. Differently from~\cite{xu2017learningdeep,xu2020probabilistic}, we show the importance of considering both spatial- and channel-wise attention, and inferring them jointly. 
More precisely, we propose a deep network for pixel-level prediction where the attention model consists of latent gates. Both the features and the gates are modeled as latent variables to be inferred, and the gating mechanism allows to model the information flow between hidden features. In plain words, the gates regulate which hidden features are linked (open gate) and which ones are not (closed gate). Additionally, we enforce structure within these gates, by imposing a low-rank tensor decomposition on the attention tensor. Our intuition is that by jointly considering channel and spatial dependencies, better feature representations can be learned. Our experimental results, reported in Section \ref{sec:experiments}, demonstrate the validity of our idea. Finally, in this paper we propose to cast the inference problem into a maximum-likelihood estimation formulation that is made computationally tractable thanks to a variational approximation. We implement the maximum likelihood update rules within a neural network, so that they can be jointly learned with the preferred CNN front-end.} We called our approach based on structured attention and variational inference VarIational STructured Attention Networks or \method. We evaluate our method on multiple pixel-wise prediction problems, \ie~monocular depth estimation, semantic segmentation and surface normale prediction, considering six publicly available datasets, \ie~NYUD-V2~\cite{silberman2012indoor}, KITTI~\cite{Geiger2013IJRR}, Pascal-Context~\cite{mottaghi2014role}, Pascal {VOC2012}~\cite{everingham2010pascal}, Cityscape~\cite{cordts2016cityscapes} and ScanNet~\cite{dai2017scannet}. Our results demonstrate that VISTA-Net is able to learn rich deep representations thanks to the proposed structured attention and to our probabilistic formulation, performing comparably or surpassing state-of-the-art methods.

To summarize, the contribution of this paper is threefold:
\begin{itemize}

\item First, we introduce a novel structured attention mechanism for {effectively learning deep representations}, jointly {modeling spatial-wise and channel-wise semantic dependencies and their interactions}.

\item Second, we propose to use our structured attention tensor within a probabilistic framework, thus introducing a principled {manner} of modeling the statistical relationships between channel-wise and spatial-wise attention.

\item Third, extensive experiments are conducted on three distinct pixel-wise prediction tasks and on six different challenging datasets, demonstrating that the proposed framework is competitive or outperforms previous methods while being task-agnostic, i.e. applicable to different continuous and discrete pixel-level prediction problems.
\end{itemize}

\section{Related Work}
{In this section we review previous works on learning deep representations for pixel-level prediction tasks within a probabilistic framework. As one of our key contributions is the introduction of a novel variational structured attention mechanism, we also discuss previous works considering attention models for deep representation learning. Finally, we also briefly review the state of the art on three important pixel-wise prediction tasks, \ie~monocular depth estimation, semantic segmentation and surface normal prediction, on which the effectiveness of our approach is extensively demonstrated.}

\subsection{Learning deep representations with CRFs}
{Since the seminal work of Zhang \etal \cite{zheng2015conditional}, where they showed that mean-field approximate inference of CRFs can be implemented as Recurrent Neural Networks, many other works have considered the integration of probabilistic graphical models within convolutional networks for improving the performance in pixel-level prediction tasks.}
For instance, Wang~\etal~\cite{wang2015towards} introduced a two-layer hierarchical CRF to fuse global and region-wise local predictions for depth map prediction.
Liu et al.~\cite{liu2015deep} proposed an end-to-end trainable network implementing a continuous CRF which estimate depth information. Xu~\etal~\cite{xu2017multi} improved over~\cite{liu2015deep} by presenting a multi-scale continuous CRF model to learn the multi-scale features and optimally fuse them. {However, no attention mechanism is considered in this work. More recently, Xu~\etal~\cite{xu2020probabilistic} proposed the AG-CRF model, incorporating spatial dependencies within a structured probabilistic framework. Our approach significantly differ from \cite{xu2020probabilistic} and from all these previous methods as it incorporates spatial- and channel-wise dependencies within a single attention tensor.}

{Our work is also closely related to previous studies on dual graph convolutional network~\cite{zhang2019dual} and dynamic graph message passing networks~\cite{zhang2020dynamic}. However, while they also resort on message passing for learning refined deep feature representations, they lack a probabilistic formulation.}

\subsection{Attention Models}
Several works have considered integrating attention models within deep architectures to improve performance in several tasks such as image categorization~\cite{xiao2015application}, 
speech recognition~\cite{chorowski2015attention}, image search~\cite{yu2016deep,yu2017multi}, image generation \cite{tang2021attentiongan,liu2021cross,tang2020dual,tang2020xinggan,tang2019multi,ding2020cross,tang2019attention}, audio-visual analysis \cite{duan2021audio,duan2021cascade}, and machine translation~\cite{vaswani2017attention,kim2017structured,luong2015effective}. Focusing on pixel-wise prediction, Chen \etal~ \cite{chen2016attention} first described an attention model to combine multi-scale features learned by a FCN for semantic segmentation. Zhang \etal~\cite{zhang2018context} designed EncNet, a network equipped with a channel
attention mechanism to model global context. Zhao \etal~
\cite{zhao2018psanet} proposed to account for pixel-wise dependencies introducing relative position information in spatial dimension within the convolutional layers. Huang \etal~\cite{huang2019ccnet} described CCNet, a deep architecture that embeds a criss-cross attention module with the idea of modeling contextual dependencies using sparsely-connected graphs, such as to achieve higher computational efficiency. Fu \etal~\cite{fu2019dual} proposed to model semantic dependencies associated with spatial and channel dimensions by using two separate attention modules. Zhong \etal~\cite{zhong2020squeeze} introduced a squeeze-and-attention network (SANet) specialized to pixel-wise prediction taking into account spatial and channel inter-dependencies in an efficient way.

Attention was first adopted within a CRF framework by \cite{xu2017learningdeep}, which introduced gates to control the message passing between latent variables and showed that this strategy is effective for contour detection. PGA-Net~\cite{xu2020probabilistic} improved over this work proposing feature dependant conditional kernels.
Our work significantly departs from these approaches, as we introduce a novel structured attention mechanism, jointly handling spatial- and channel-level dependencies within a probabilistic framework. Notably, we also prove that our model can be successfully employed in case of several challenging dense pixel-level prediction tasks, {where it significantly outperforms both AG-CRF~\cite{xu2017learningdeep} and PGA-Net~\cite{xu2020probabilistic}.}

\subsection{Pixel-wise Prediction}
\noindent\textbf{Monocular Depth Estimation.} 
Most recent works on monocular depth estimation are based on CNNs~\cite{eigen2015predicting,liu2015deep,wang2015towards,roymonocular,laina2016deeper, fu2018deep, gan2018monocular, lee2019big,xu2018structuredattentionguided}. For instance,
Eigen~\etal~\cite{eigen2014depth} introduced a two-streams deep network to take into account both coarse global prediction and local information. 
Fu~\etal~\cite{fu2018deep} proposed a discretization strategy to treat monocular depth estimation as a deep ordinal regression problem. They also employed a multi-scale network to capture relevant multi-scale information. 
Lee~\etal~\cite{lee2019big} introduced local planar
guidance layers in the network decoder module to learn more effective features for depth estimation.
More recently, PackNet-SfM~\cite{guizilini20203d} used 3D convolutions with self-supervision to learn detail-preserving representations. {Multi-scale representations where also considered within a CRF model in \cite{xu2017multi,liu2015deep}.
Our approach also adopts a probabilistic graphical model for learning better feature representations. However, different from these previous works it integrates an attention mechanism.}

\par\noindent\textbf{Semantic Segmentation.} As for depth estimation, nowadays CNNs are the mainstream approach for semantic segmentation \cite{chen2016deeplab,yu2015multi,ding2020lanet,long2015fully,yuan2018ocnet}.
For instance, Long~\etal~\cite{long2015fully} were the first to introduce fully convolutional networks (FCNs) for semantic segmentation, achieving significant improvements over previous models. Dilated convolutions~\cite{chen2016deeplab,yu2015multi} were designed in order to increase the receptive field while learning deep representations, further boosting performances. OCNet~\cite{yuan2018ocnet} introduced a context aggregation strategy, \ie  \ object context pooling, for robust segmentation. In APCNet \cite{he2019adaptive}, multi-scale contextual representations were constructed with multiple Adaptive Context Modules. 
Other works focused on multi-scale feature representation learning, designing appropriate convolutional encoder-decoder network structures~\cite{noh2015learning, badrinarayanan2015segnet} or considering end-to-end trainable architectures modeling CRFs~\cite{liu2015semantic, arnab2016higher, zheng2015conditional}. 
{HRNet~\cite{wang2020deep} constructs the paralleling connection between the high resolution convolution streams an low resolution convolution streams.
More recently, Wang~\etal~\cite{wang2021exploring} proposed a pixel-wise contrastive algorithm for semantic segmentation to capture both local and global content.}
Our approach adopts a probabilistic graphical model formulation but it is the first which jointly models structured spatial- and channel-wise semantic dependencies.

\noindent\textbf{Surface Normal.} 
Extracting 3D geometry from a single image is a longstanding problem in computer vision. Surface normal estimation is a classical task in this context, which requires modeling both
global and local features. Typical approaches leverage on networks with high capacity to achieve accurate predictions at high resolution. For instance, FrameNet \cite{huang2019framenet} employed the DORN \cite{fu2018deep} architecture, a modification of DeepLabv3 \cite{chen2016deeplab} that removes multiple spatial reductions (2$\times$2 max pool layers), to generate high resolution surface normal maps. A different strategy consists in designing appropriate loss terms. For instance, UprightNet \cite{xian2019uprightnet} considered an angular loss and showed its effectiveness for the task.  
Unlike previous works focusing on designing \textit{ad hoc} network structures or proposing new loss terms, here we show that introducing a structured attention module into a deep network is effective for surface normal prediction.

\begin{figure*}[t]
\centering
\includegraphics[width=1\textwidth]{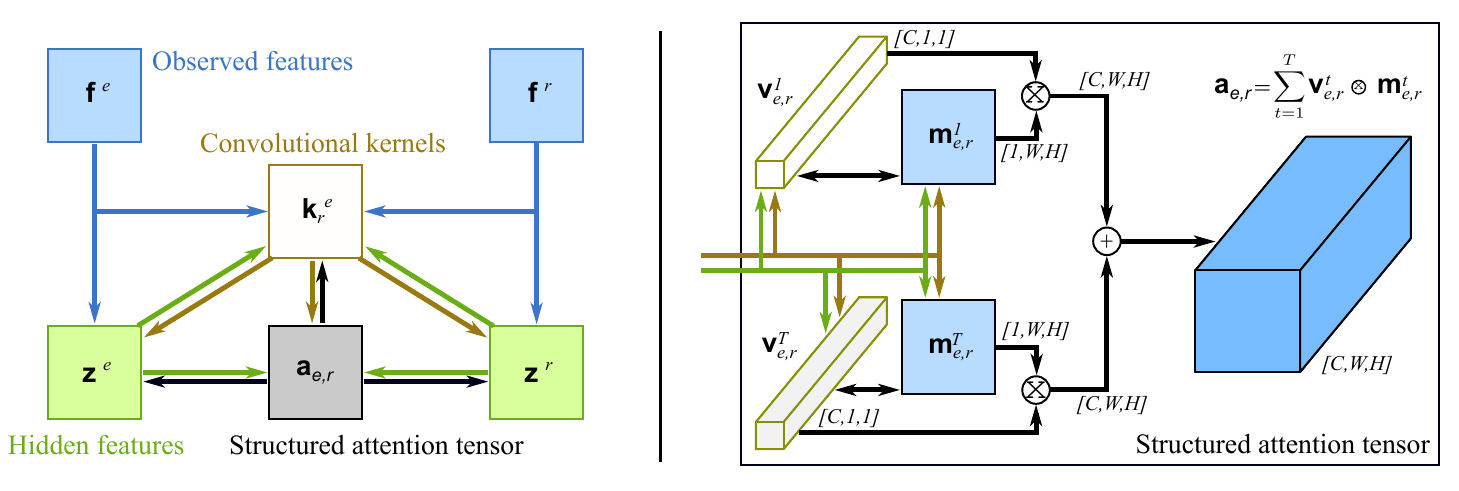}
\caption{ Schematic representation of the various hidden variables in VISTA-Net. (left) For each pair of emitting $e$ and receiving $r$ scales, their respective convolutional features $\vect{f}$ are shown in blue, their hidden variables $\vect{z}$ in green, the associated learned kernel $\vect{k}$ in yellow, and the channel-wise and spatial-wise attention tensor and matrix $\vect{v}$ and $\vect{m}$ in red. Arrows of the corresponding color denote the flow of information when updating the variable.
(right) The computational relationships between the channel-wise and spatial-wise attention variables is shown, as well as the operations required to compute the final structured attention tensor~$\vect{a}$. {The updates of the variables $\vect{k}$, $\vect{z}$, $\vect{v}$ and $\vect{m}$ as indicated by the arrows correspond to the K-step, Z-step, V-step and M-step.} }
\label{fig:inference-vista-net}
\end{figure*}

\begin{table}[t]
\centering

\caption{Notation used in the paper}
\label{tab:symbol}
\begin{tabular}{c|c}
\toprule
 Symbol & Description \\
\midrule
 \textbf{I}      & Input  image \\
 \textbf{F}      & Feature map / observed feature  \\
 $\theta_c$     & Parameters of a  generic front-end CNN model   \\
 $\Theta$  & Parameters of the entire deep network \\ 
 \textbf{Z}        & Hidden feature   \\
 \textbf{v}      & Channel-wise attention vector  \\
 \textbf{m}      & Spatial-wise attention map  \\ 
 \textbf{a}      & Structured attention tensor \\
 \textbf{K}      & Pair-wise CRF kernel   \\ 
 $S$       & Number of scales    \\
 $T$  & Number of spatial- or channel-wise attentions \\   
\bottomrule  

\end{tabular}
\end{table}

\section{Variational Structured Attention Networks}\label{sec:method}
As previously discussed, the goal of our work is to enhance the learned representation by structuring the attention within a probabilistic formulation. On the one side, inducing structure in the attention mechanisms has been proven to be successful~\cite{fu2019dual,zhong2020squeeze}. On the other side, probabilistic formulations combined with deep architectures are interesting for pixel-level prediction tasks~\cite{xu2017learning}. Up to our knowledge, we are the first to bring together recent advances in pixel-wise prediction by formulating a novel structured attention mechanism within a probabilistic CRF-like inference framework. In the following, we first describe the problem formulation, and specifically how to structure the attention within a CRF formulation. Secondly, we derive the energy function as well as the variational approximation. Finally, the derived inference formulae and associated algorithm are detailed.

\subsection{Problem Formulation}

Given an input image $\vect{I}$, we consider a generic front-end CNN model with parameters $\vect{\theta}_c$, which outputs a set of $S$ multi-scale feature maps $\vect{F} = \{\vect{f}_{s}\}_{s=1}^{S}$ {(in Table~\ref{tab:symbol}, we provide the exhaustive list of notation).} To each of these feature maps (or scale) $s$, we can associate a hidden feature map $\vect{z}_s$ of the same size of $\vect{f}_s$, and that needs to be inferred within the CRF formulation. These hidden variables correspond to refined convolutional futures that incorporate information and attention from other feature maps, so as to better represent the key information for the pixel-level task at hand. For each pair of \textit{emitting} $e$ and \textit{receiving} $r$ feature maps, we associate the usual CRF pair-wise kernel, denoted by $\vect{k}_{r}^{e}$ as well as an attention tensor $\vect{a}_{r}^{e}$. This attention tensor should encode the entries of the emitting feature map that better help the inference of the receiving hidden features. Indeed, we inspire from the CRF formulation with gating variables proposed in~\cite{xu2017learningdeep}, so that each entry of the attention tensor is a binary variable indicating whether or not the emitting feature map entry should be used to infer the receiving feature map entry.
 In Table~\ref{tab:symbol}, we provide an exhaustive list of symbols used in the paper and the corresponding description.

Inspired by~\cite{fu2019dual}, where a spatial- and a channel-wise \textit{full-rank} tensors are computed, we opt to infer different spatial and channel attention variables. Differently from~\cite{fu2019dual}, we propose to structure a generic attention tensor $\vect{a}$ (we drop the emitting and receiving scale indices for now) of dimension $W\times H\times C$ (widht, height, channels), as the sum of $T$ one-rank tensors, as shown in Fig. \ref{fig:inference-vista-net} (right), by writing:
\begin{equation}
    \vect{a} = \sum_{t=1}^T \vect{m}^t \otimes \vect{v}^t \in \{0,1\}^{W\times H\times C},\label{eq:structured-attention}
\end{equation}
where $\vect{m}^t$ is a binary image of $P=W\times H$ pixels, $\vect{m}_t\in\{0,1\}^{P}$, and $\vect{v}^t$ is a stochastic vector of dimension $C$, $\vect{v}^t\in\{0,1\}^C$, $\sum_{c=1}^C v^{t,c} = 1$, and $\otimes$ denotes the tensor product, in the case above leading to a 3-way binary tensor of dimensions $W\times H\times C$. {Each of the tensor products within the sum yields a tensor of rank-1, consequently limiting the rank of $\vect{a}$ to be at maximum $T$. In this way, we reduce ambiguity and ease the learning. The rationale behind this choice arises from the original definition of the rank of a tensor, adapted to our case. Indeed, the CANDECOMP/PARAFAC (CP) rank of a tensor $\mathbf{a}$ is defined as the smaller number of rank-one tensors needed to reconstruct $\mathbf{a}$, see for instance~\cite{zhou2017tensor}. However, in our case this is not appropriate, since we would loose the spatial structure (each one-way tensor would be the outer product of a width-way, a height-way and a channel-way vectors). Instead, we adapted this definition to have the outer product of an image-way tensor (replacing the width-way and height-way) and a channel-way vector.} 

Constraining the rank of the attention tensor means that the model is conceived to pay attention to only $T$ channels of the feature map. While this could seem limiting at first glance we remark that: 

(i) The model learns which are the \textit{optimal} $T$ channels among the possible $C$ that have to be used to refine the hidden variables. (ii) the a posteriori distribution of $\vect{m}^t$ boils down to a convex combination of all channels, as it will appear clear when discussing the inference procedure. (\ref{eq:structured-attention}) is the algebraic expression of the proposed structured attention mechanism, and is the methodological foundation of VISTA-Net. Intuitively, the structured attention tensor should help refining the hidden variables $\vect{z}_s$ to allow better performance at various pixel-level prediction tasks.

For each emitting-receiving pair of scales, $e$ and $r$, we thus propose to infer the $T$ associated attention maps $\{\vect{m}^t_{e,r}\}_{t=1}^T$ and vectors $\{\vect{v}^t_{e,r}\}_{t=1}^T$, structuring the attention tensors. In addition, we also propose to learn the pair-wise binary kernels of the CRF, $\vect{k}_{r}^{e}$. We believe learning the kernels is important because it allows the CRF to weight the information flow depending on the content rather than keeping the same weights for all images. 

Summarizing, in addition to the set of hidden CRF feature maps, $\vect{Z}=\{\vect{z}_s\}_{s=1}^S$, we propose to infer the set of pair-wise CRF kernels $\vect{K}=\{\vect{k}_{r}^e\}_{e,r}^{S,S}$ and the sets of spatial-wise and channel-wise attention maps and stochastic vectors, denoted as $\vect{M}=\{\vect{m}_{e,r}^t\}_{e,r,t=1}^{S,S,T}$ and $\vect{V}=\{\vect{v}_{e,r}^t\}_{e,r,t=1}^{S,S,T}$, respectively. In the following section, we describe the energy function associated to our formulation and propose a variational approximation that allows us to derive closed-form solutions for the a posterior distributions of all the aforementioned random variables.

\subsection{Energy Function and Variational Approximation}
\label{sec:var_approx}
Our model consists on three different latent variables: the hidden features $\vect{Z}$, and the hidden attention maps $\vect{M}$ and vectors $\vect{V}$. In addition, we also consider inferring the CRF kernels, denoted by $\vect{K}$ from the data. More precisely, the energy function associated to the proposed models writes:
\begin{align}
 & -E(\vect{Z}, \vect{M}, \vect{V}, \vect{K} , \vect{F}, \Theta) \nonumber \\ & = \sum_{s} \sum_{p,c} \phi_z ({z}_{r}^{p,c}, {f}_{r}^{p,c})\nonumber\\ 
  & + \sum_{e,r} \sum_{p,c,p'\!\!,c'} \sum_{t} m_{e,r}^{t,p} v_{e,r}^{t,c} \psi ({z}_{r}^{p,c}, {z}_{e}^{p'\!\!,c'},{k}_{r,p,c}^{e,p'\!\!,c'}) \nonumber\\
 & +  \phi_k ({f}_{r}^{p,c}, {f}_{e}^{p'\!\!,c'}, {k}_{r,p,c}^{e,p'\!\!,c'}).
\end{align}
where $\phi_z$, $\phi_k$ and $\psi$ are potentials to be defined and ${k}_{r,p,c}^{e,p'\!\!,c'}$ denotes the kernel value weighting the information flow from the $(p'\!\!,c')$-th value of the feature map of scale $e$ to the $(p,c)$-th value of the feature map of scale $r$.

Since the exact a posteriori distribution is not computationally tractable, we opt to approximate it with the following family of separable distributions:
\begin{align}
    &p(\vect{Z}, \vect{M}, \vect{V}, \vect{K} | \vect{F}, \Theta) \nonumber \\
    & \approx q(\vect{Z}, \vect{M}, \vect{V}, \vect{K})\nonumber\\
    &= q_z(\vect{Z}) q_m(\vect{M}) q_v(\vect{V}) q_k(\vect{K}).
    \label{eq:var-approx}
\end{align}

In that case, the optimal solution for each of the factors of the distribution is to take the expectation w.r.t.\ to all the others, for instance:
\begin{equation}
    q_z(\vect{Z}) \propto \exp\Big( - \mathbb{E}_{q_m(\vect{M}) q_v(\vect{V}) q_k(\vect{K})} \Big\{ E(\vect{Z}, \vect{M}, \vect{V}, \vect{K} , \vect{F}, \Theta) \Big\} \Big).
\end{equation}

It can be shown that the optimal variational factors write:

\begin{small}
\begin{equation}
\begin{aligned}
q_z({z}_{r}^{p,c}) &\propto \exp\Big(\phi_z({z}_r^{p,c},{f}_r^{p,c}) \nonumber\\
&+ \sum_{e\neq r}\sum_{t}  \bar{m}_{e,r}^{t,p} \bar{v}_{e,r}^{t,c} \sum_{p'\!\!,c'} \mathbb{E}_{q_z q_k}\{\psi ({z}_{r}^{p,c}, {z}_{e}^{p'\!\!,c'},{k}_{r,p,c}^{e,p'\!\!,c'})\} 
\Big),\\
q_m(m_{e,r}^{t,p}) &\propto \exp\Big( m_{e,r}^{t,p} \sum_c \bar{v}_{e,r}^{t,c}\sum_{p'\!\!,c'} \mathbb{E}_{q_z,q_k}\{\psi({z}_{s}^{p,c}, {z}_{s'}^{p'\!\!,c'},{k}_{r,p,c}^{e,p'\!\!,c'})\} \Big),\\
q_v(v_{e,r}^{t,c}) &\propto \exp\Big( v_{e,r}^{t,c} \sum_p \bar{m}_{e,r}^{t,p}\sum_{p'\!\!,c'} \mathbb{E}_{q_z,q_k}\{\psi({z}_{s}^{p,c}, {z}_{s'}^{p'\!\!,c'},{k}_{r,p,c}^{e,p'\!\!,c'})\} \Big),\\
q_k({k}_{r,p,c}^{e,p'\!\!,c'}) &\propto \exp\Big( 
\phi_k({f}_r^{p,c},{f}_{e}^{p'\!\!,c'},{k}_{r,p,c}^{e,p'\!\!,c'}) \nonumber \\
&+ \sum_{t}  \bar{m}_{e,r}^{t,p} \bar{v}_{e,r}^{t,c} \mathbb{E}_{q_z}\{\psi ({z}_{s}^{p,c}, {z}_{s'}^{p'\!\!,c'},{k}_{r,p,c}^{e,p'\!\!,c'})\} 
\Big).
\end{aligned}
\label{eq:variational-factors}
\end{equation}
\end{small}

where $\bar{m}_{e,r}^{t,p}=\mathbb{E}_{q_m}\{{m}_{e,r}^{t,p}\}$ denotes the a posteriori mean, and analogously for $\bar{v}_{e,r}^{t,c}$. This result also implies that thanks to the variational approximation in~(\ref{eq:var-approx}), the posterior distributions factorise in each of the variables above, \eg\ $q_z(\vect{Z})=\prod_{r,p,c=1}^{S,P,C} q_z(z_r^{p,c})$.The relation between the various hidden variables as for their inference is shown in Fig.\ref{fig:inference-vista-net} (left). In addition, we also show the information flow between the hidden variables using arrows. Finally, in Fig.\ref{fig:inference-vista-net} (right) we show the relation between the channel-wise and spatial attention variables and how the final structured attention tensor is computed.

\subsection{Inference with \method}
\label{sec:inference}
In order to construct an operative model we need to define the potentials $\phi_z$, $\phi_k$ and $\psi$. In our case, the unary potentials correspond to:
\begin{align}
\phi_z({z}_{r}^{p,c},{f}_{r}^{p,c}) = & - \frac{b_r^{p,c}}{2} ({z}_{r}^{p,c}-{f}_r^{p,c})^2,\nonumber \\ 
 \phi_k({f}_r^{p,c},{f}_{e}^{p'\!\!,c'},{k}_{r,p,c}^{e,p'\!\!,c'}) = & -\frac{1}{2} ( {k}_{r,p,c}^{e,p'\!\!,c'} - {f}_r^{p,c} {f}_{e}^{p'\!\!,c'})^2.
 \label{eq:unary-kernel}
\end{align}
 
where $b_s^{p,c}>0$ is a weighting factor. $\psi$ is bilinear in the hidden feature maps:
\begin{equation}
 \psi({z}_r^{p,c},{z}_{e}^{p'\!\!,c'},{k}_{r,p,c}^{e,p'\!\!,c'}) = {z}_r^{p,c} {k}_{r,p,c}^{e\!\!,p'\!\!,c'\!\!} z_{e}^{p'\!\!,c'}.
 \label{eq:binary-kernel}
\end{equation}

Using the over bar notation also for the hidden features and kernels, e.g.\ $\bar{z}_{s}^{p,c} = \mathbb{E}_{q_z}\{z_s^{p,c}\}$, and by combining the kernel definitions~(\ref{eq:unary-kernel}) and~(\ref{eq:binary-kernel}) with the expression of the variational factors~(\ref{eq:variational-factors}), we obtain the following update rules for the latent variables.

\noindent\textbf{Z-step.} It can be seen that the posterior distribution on $q_z$ is Gaussian with mean:
\begin{equation}
\bar{z}_{s}^{p,c} = \frac{1}{b_s^{p,c}}\Big(b_s^{p,c}{f}_s^{p,c} + \sum_{e} \sum_{t} \bar{m}_{s,s'}^{t,p} \bar{v}_{s,s'}^{t,c} \sum_{p'\!\!,c'} \bar{k}_{r,p,c}^{e,p'\!\!,c'} \bar{z}_{s'}^{p'\!\!,c'} \Big) \label{eq:h-update}
\end{equation}
This corresponds to the update rule obtained in~\cite{xu2017learningdeep} with two remarkable differences. First, the posterior of the attention gate corresponds to the posterior of the structured tensor of rank $T$. Second, the impact of the neighboring features is weighted by the expected kernel value $\bar{k}_{r,p,c}^{e,p'\!\!,c'}$.

\noindent\textbf{M-step.} The variational approximation leads to a Bernoulli distribution for $q_m(m_{e,r}^{t,p})$, which boils down to the following a posteriori mean value using the sigmoid function $\sigma$:
\begin{equation}
    \bar{m}_{e,r}^{t,p} = \sigma \Big( \sum_c \bar{v}_{e,r}^{t,c} \sum_{p'\!\!,c'} \bar{z}_{s}^{p,c}\bar{k}_{r,p,c}^{e,p'\!\!,c'}\bar{z}_{s'}^{p'\!\!,c'}\Big).
    \label{eq:m-update}
\end{equation}

\noindent\textbf{V-step.} It can be shown that the approximated posterior distribution is categorical, and that the expected value of each dimension of $\mathbf{v}_{e,r}^t$ can be computed using the softmax operator:
\begin{equation}
    (\bar{v}_{e,r}^{t,c})_{c=1}^C = \textrm{softmax} \Big( \sum_p \bar{m}_{e,r}^{t,p} \sum_{p'\!\!,c'} \bar{z}_{s}^{p,c}\bar{k}_{r,p,c}^{e,p'\!\!,c'}\bar{z}_{e}^{p'\!\!,c'}\Big)_{c=1}^C . \label{eq:v-update}
\end{equation}

\noindent\textbf{K-step.} Finally, we need to derive the update rules for $\vect{K}$. By further deriving the corresponding variational posterior distribution, it can be shown that the a posteriori distribution for the kernels is a Gaussian distribution with the following mean:
\begin{equation}
    \bar{k}_{r,p,c}^{e,p'\!\!,c'} = {f}_r^{p,c}{f}_{e}^{p'\!\!,c'} + \sum_{t}  \bar{m}_{e,r}^{t,p}\bar{v}_{e,r}^{t,c} \bar{z}_{r}^{p,c}\bar{z}_{e}^{p'\!\!,c'}.
    \label{eq:k-update}
\end{equation}

This solution is very straightforward, but since the kernels are estimated independently for each pair of receiving $(r,p,c)$ - emitting $(e,p'\!\!,c')$ pixels, it has two major drawbacks. First, the kernel values are estimated without any spatial context. Second, given the large amount of kernel values, one must find a very efficient way to compute them. We propose to kill two birds with one stone by learning the kernels from the features using convolutional layers. By design, they take spatial context into account, and many popular libraries have efficient implementations of the convolution operation. The estimated kernel corresponding to the input channel $c'$ of scale $e$, $\mathbf{k}_{r}^{e,c'}$ is computed via a convolutional operation. The input of the convolution is a concatenation of the tensor $\mathbf{f}_r + \mathbf{z}_r\sum_{t=1}^T \bar{m}_{r,e}^t\otimes \bar{v}_{r,e}^t$ and the image $\mathbf{z}_{e}^{c'}$ resized to the spatial size of $\mathbf{f}_r$.

\noindent\textbf{Joint Learning.}
We implement the inference procedure described before within the neural network, on the top of the CNN front-end. Indeed, implementing all inference operations using available deep learning operators has two prominent advantages. First, we can perform the inference and learning the CNN front-end at the same time, within the same formalism and for the same aim. Second, this allows direct parallelisation of our method, speeding up training and inference. 

The precise implementation goes as follows. Regarding $\bar{\vect{z}}_r$, we first apply message passing from the $e$-th scale to the $r$-th scale is performed with $\vect{z}_{e\rightarrow r} \leftarrow \bar{\mathbf{k}}_{r}^{e} \circledast \bar{\vect{z}}_{e}$, where $\circledast$ denotes the convolutional operation and $\bar{\mathbf{k}}_{r}^{e}$ denotes the corresponding learned convolution kernel. We then apply element-wise product with the corresponding structured attention tensor $\sum_{t=1}^T \bar{\vect{m}}_{e,r}^t \otimes \bar{\vect{v}}_{e,r}^t$. Finally we compute the element-wise sum with other emiting scales and the feature maps $\vect{f}_r$, see~(\ref{eq:h-update}). Regarding $\bar{\vect{m}}_{e,r}$, we first compute the element-wise product between $\bar{\vect{z}}_r$ and $\vect{z}_{e\rightarrow r}$. The sum over channels weighted by $\bar{\vect{v}}_{e,r}$ is computed previous to applying pixel-wise sigmoid, see~(\ref{eq:m-update}). Regarding $\bar{\vect{v}}_{e,r}$ we operate in a very similar fashion, but weighting each pixel with $\bar{\vect{m}}_{e,r}$ and then summing every channel independently, before applying softmax, see~(\ref{eq:v-update}). Regarding $\bar{\vect{k}}_{r}^{e,c'}$, as discussed before, it is computed via a convolutional operation on the concatenations of $\mathbf{f}_{t_m} + \mathbf{g}_{t_m} $ and the image $\mathbf{z}_{e}^{c'}$ resized to the spatial size of $\mathbf{f}_r$. In terms of initialisation, we draw a random guess for $\vect{M}$ and $\vect{V}$, and set $\vect{Z}$ to $\vect{F}$. This allows us to update the kernels, then the other variables. Our structured attention method is summarised in Algorithm~\ref{alg:vista}.

Once the hidden variables are updated, we use them to address several different pixel-wise prediction tasks involving continuous and discrete variables, including monocular depth estimation, surface normal estimation and semantic segmentation. Following previous works, the network optimization losses for these three tasks are a standard L2 loss~\cite{xu2017multi}, a cosine similarity loss~\cite{eigen2014depth} and a cross-entropy loss~\cite{chen2016deeplab}, respectively. The CNN front-end and \method, are jointly trained end-to-end.

\begin{figure*}[!t]
\centering
    \begin{minipage}{0.24\linewidth}
        \centering
        \includegraphics[width=0.993\textwidth,height=0.7in]{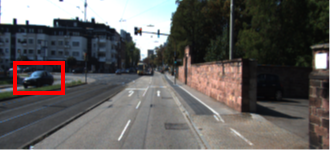}\\
        \includegraphics[width=0.993\textwidth,height=0.7in]{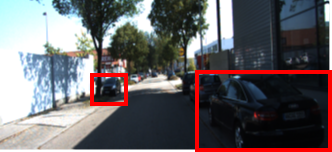}\\
        \includegraphics[width=0.993\textwidth,height=0.7in]{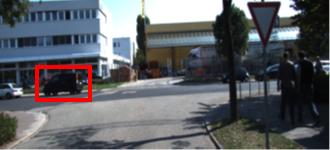}\\
        \includegraphics[width=0.993\textwidth,height=0.7in]{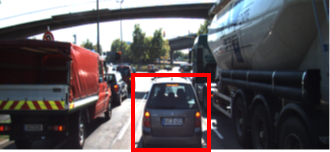}\\
        \includegraphics[width=0.993\textwidth,height=0.7in]{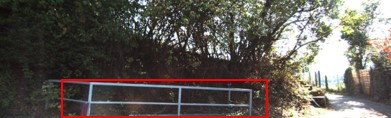}\\
        \footnotesize (a) Image 
    \end{minipage}%
    \begin{minipage}{0.24\linewidth}
        \centering
        \includegraphics[width=0.993\textwidth,height=0.7in]{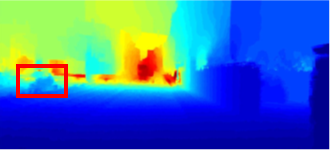}\\
        \includegraphics[width=0.993\textwidth,height=0.7in]{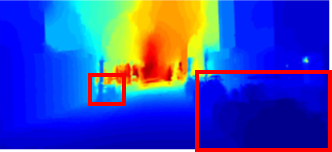}\\

        \includegraphics[width=0.993\textwidth,height=0.7in]{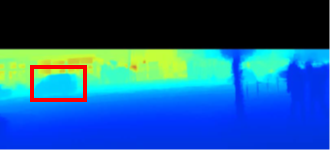}\\

        \includegraphics[width=0.993\textwidth,height=0.7in]{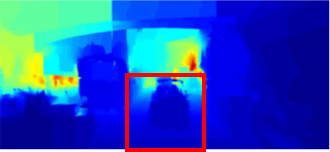}\\
        \includegraphics[width=0.993\textwidth,height=0.7in]{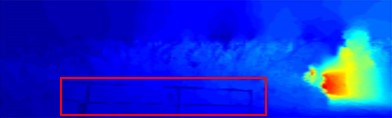}\\  
        \footnotesize (b) GT 
    \end{minipage}%
    \begin{minipage}{0.24\linewidth}
        \centering
        \includegraphics[width=0.993\textwidth,height=0.7in]{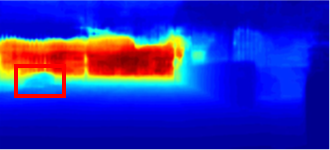}\\
        \includegraphics[width=0.993\textwidth,height=0.7in]{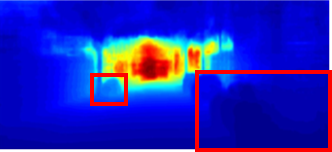}\\
        \includegraphics[width=0.993\textwidth,height=0.7in]{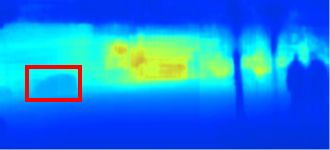}\\

        \includegraphics[width=0.993\textwidth,height=0.7in]{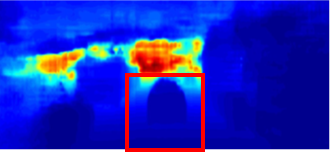}\\
        \includegraphics[width=0.993\textwidth,height=0.7in]{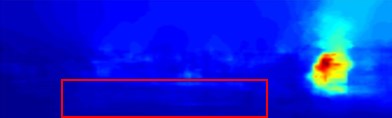}\\ 
        \footnotesize (c) DORN
    \end{minipage}%
    \begin{minipage}{0.24\linewidth}
        \centering
        \includegraphics[width=0.993\textwidth,height=0.7in]{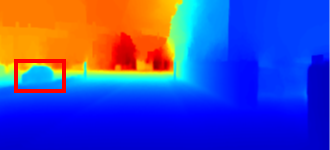}\\
        \includegraphics[width=0.993\textwidth,height=0.7in]{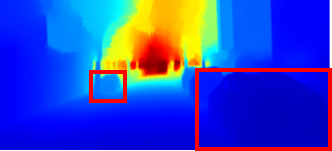}\\
        \includegraphics[width=0.993\textwidth,height=0.7in]{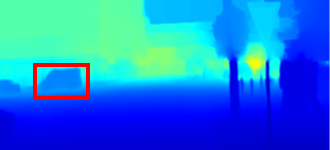}\\
        \includegraphics[width=0.993\textwidth,height=0.7in]{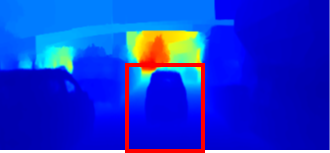}\\
        \includegraphics[width=0.993\textwidth,height=0.7in]{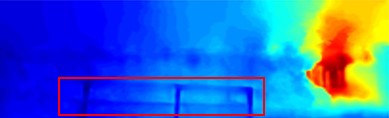}\\ 
        \footnotesize (d) \method\ (ours) 
    \end{minipage}%
 
\centering
\caption{Qualitative examples on the KITTI dataset.}
\label{fig:vis_kitti}
\end{figure*}

\begin{table}[t]
\caption{Depth Estimation: KITTI dataset. 
Only monocular estimation methods are reported. 
}
\centering
\label{tab:overall_kitti}
\scalebox{0.68}{%
\begin{tabular}{lccccccc}
\toprule[1.2pt]
\multirow{2.5}{*}{Method} & \multicolumn{4}{c}{Error (lower is better)} & \multicolumn{3}{c}{Accuracy (higher is better)} \\ \cmidrule{2-8} 
   & abs-rel & sq-rel & rms & log-rms & $\delta \textless 1.25$ & $\delta \textless 1.25^2$ & $\delta \textless 1.25^3$ \\ \midrule
{CC}~\cite{ranjan2019competitive}   & 0.140 & 1.070 & 5.326 & 0.217 & 0.826 & 0.941 & 0.975 \\
Bian~\etal \cite{bian2019unsupervised}   & 0.137 & 1.089 & 5.439 & 0.217 & 0.830 & 0.942 & 0.975 \\
$S^3$Net~\cite{cheng2020s} & 0.124 &	0.826 &	4.981 &	0.200 & 0.846 & 0.955 &	0.982\\
MS-CRF~\cite{xu2017multi}  & 0.125 & 0.899 & 4.685 & - & 0.816 & 0.951 & 0.983 \\
AG-CRF~\cite{xu2017learningdeep}   & {0.126}  & {0.901} & {4.689} & {0.157} & 0.813  & {0.950} & {0.982} \\
DeFeat~\cite{spencer2020defeat} & 0.126 & 0.925 & 5.035 & 0.200 & 0.862 & 0.954 & 0.980 \\
Monodepth2~\cite{godard2019digging}   & 0.115 & 0.903 & 4.863 & 0.193 & 0.877 & 0.959 & 0.981 \\
pRGBD~\cite{tiwari2020pseudo} & 0.113 & 0.793 & 4.655 & 0.188 &	0.874 &	0.960 &	0.983\\
SGDepth~\cite{klingner2020self} & 0.107 & 0.768 & 4.468 & 0.180 &	0.891 &	0.963 & 0.982\\
Johnston~\etal \cite{johnston2020self}   & 0.106 & 0.861 & 4.699 & 0.185 & 0.889 & 0.962 & 0.982 \\
Shu~\etal \cite{shu2020feature} & 0.104 &	0.729 &	4.481 &	0.179 &	0.893 &	0.965 &	0.984\\
DORN~\cite{fu2018deep}   & 0.072 & 0.307 & 2.727 & 0.120 & 0.932 & 0.984 & 0.994 \\
Yin~\etal \cite{yin2019enforcing}   & 0.072 & - & 3.258 & 0.117 & 0.938 & 0.990 & \textbf{0.998} \\
{PackNet-SfM}~\cite{guizilini20203d}   & 0.071 &	0.359 &	3.153 &	0.109 &	0.944 &	0.990 &	0.997\\
PGA-Net~\cite{xu2020probabilistic} & 0.063 & 0.267 & 2.634 & 0.101 & 0.952 & 0.992 & \textbf{0.998}\\ 
Lee~\etal \cite{lee2019big}   & \textbf{0.061} & 0.261 & 2.834 & {0.099} & {0.954} & 0.992 & \textbf{0.998} \\ \midrule
VISTA-Net  & \textbf{0.061} & \textbf{0.211} & \textbf{2.445} & \textbf{0.092} & \textbf{0.960} & \textbf{0.994} & \textbf{0.998} \\ 
\bottomrule[1.2pt]
\end{tabular}}
\end{table}

\begin{algorithm*}[!tb]
\SetKwInOut{Input}{Input}
\SetKwInOut{Output}{Output}
\caption{
Our VISTA-Net for a given receiving scale~$r$.}
\label{alg:vista}
\Input{
\begin{itemize}
\item $\{\vect{f}_e\}_{e\in E}$ -- set of emitting feature map.
\item $\vect{f}_r$ -- receiving feature map.
\end{itemize}
}
\Output{
\begin{itemize}
\item $\hat{\vect{f}}_r$ -- updated receiving feature map.
\end{itemize}
}
\For{$e\in E$}{
$\bar{\vect{z}}_{e}\leftarrow \bar{\mathbf{k}}_{r}^{e} \circledast \vect{f}_{e}$\\

$\bar{\vect{z}}_{e\rightarrow r} \leftarrow \bar{\mathbf{k}}_{r}^{e} \circledast \bar{\vect{z}}_{e}$ -- will replace $\vect{z}_{s'}$ in~(\ref{eq:unary-kernel})-(\ref{eq:k-update}).\\
}
$\bar{\vect{z}}_{r} \leftarrow \vect{b}_r^{-1}\Big(\vect{b}_r \odot \vect{f}_r + \sum_{e} \bar{\vect{z}}_{e\rightarrow r} \odot \sum_{t} \bar{\vect{m}}_{r,e}^t\ \otimes \bar{\vect{v}}_{r,e}^t \Big)$ (where $\vect{b}_r^{-1}$ denotes the element-wise inverse) \algorithmiccomment{{\textbf{Z-step}}}\\

\For{$t \leftarrow 1$ \KwTo $T$}{

$\vect{m}_{r,e}^t\leftarrow \sum_c \bar{v}_{r,e}^{t,c}   \left(\bar{\vect{z}}_{r} \odot \bar{\vect{z}}_{e\rightarrow r}\right)_{c}$ (where $()^c$ extracts the $c$-th channel) \algorithmiccomment{{\textbf{M-step}}}\\
$\bar{\vect{m}}_{r,e}^t\leftarrow \textrm{sigmoid}(\vect{m}_{r,e}^t)$\algorithmiccomment{{\textbf{M-step}}}\\
$\vect{v}_{r,e}^t\leftarrow \sum_p \bar{m}_{r,e}^{t,p} \left(\bar{\vect{z}}_{r} \odot \bar{\vect{z}}_{e\rightarrow r}\right)^{p}$ (where $()^p$ extracts the $p$-th pixel)  \algorithmiccomment{{\textbf{V-step}}}\\
$\bar{\vect{v}}_{r,e}^t\leftarrow \textrm{softmax}(\vect{v}_{r,e}^t)$\algorithmiccomment{{\textbf{V-step}}}\\
}
$\bar{\vect{k}}_{r}^{e} \leftarrow \vect{f}_r \otimes \vect{f}_{e} + \left(\sum_{t} (\bar{\vect{m}}_{r,e}^{t}\otimes \bar{\vect{v}}_{r,e}^{t}) \odot \bar{\vect{z}}_{r} \right) \otimes \bar{\vect{z}}_{e}$ \algorithmiccomment{{\textbf{K-step}}}\\
$\hat{\vect{f}}_r\leftarrow \mathbf{f}_r + \bar{\vect{k}}_{r}^{e}  \circledast   \bar{\mathbf{z}}_r$\\
\KwRet{$\hat{f}_r$}
\end{algorithm*}

\section{Experimental Evaluation}\label{sec:experiments}

\subsection{Datasets}

The {NYU-v2} dataset~\cite{silberman2012indoor} is used to evaluate our approach in the depth estimation task. We use 120K RGB-Depth pairs with a resolution of $480\times640$ pixels, acquired with a Microsoft Kinect device from 464 indoor scenes. We follow the standard train/test split as previous works~\cite{eigen2014depth}, using 249 scenes for training and 215 scenes (654 images) for testing.

The {KITTI} dataset~\cite{Geiger2013IJRR} is a large-scale outdoor dataset created for various autonomous driving tasks. We use it to evaluate the depth estimation performance of our proposed model. Following the standard training/testing split proposed by Eigen~\etal~\cite{eigen2014depth}, we specifically use 22,600 frames from 32 scenes for training, and 697 frames from the rest 29 scenes for testing.

The Pascal-Context dataset~\cite{mottaghi2014role} is used for assessing the performance of VISTA-Net on the semantic segmentation task. It consists of RGB images from Pascal VOC 2010 and annotated semantic labels for more than 400 classes. As in previous works~\cite{chen2016deeplab, zhang2018context}, we consider the most frequent 59 classes plus the background class. The remaining classes are masked during training and testing.

The PASCAL VOC2012 dataset~\cite{everingham2010pascal} is the most widely studied segmentation benchmark, which contains 20 classes and is composed of 10,582 training images, and 1,449 validation images, 1,456 test images. We train the \method{} using augmented data as previous works \cite{zhong2020squeeze,long2015fully}.

The Cityscapes dataset~\cite{cordts2016cityscapes} is tasked for urban segmentation, Only the 5,000 finely annotated images are used in our experiments and are divided into 2,975/500/1,525 images for training, validation, and testing.

The {ScanNet} dataset~\cite{dai2017scannet} is a large RGB-D dataset for 3D scene understanding. We employ it to evaluate the surface normal performance of our proposed model. ScanNet dataset is divided into 189,916 for training and 20,942 for test with file lists provided in \cite{dai2017scannet}.

\subsection{Evaluation Metrics}
\par\noindent\textbf{Evaluation Protocol on Monocular Depth Estimation.} Following the standard evaluation protocol as in previous works~\cite{eigen2015predicting,eigen2014depth,wang2015towards}, the following quantitative evaluation metrics are adopted in our experiments: 
\begin{itemize}
\item Abs relative error (abs-rel): 
 \( \frac{1}{K}\sum_{i=1}^K\frac{|\tilde{d}_i - d_i^\star|}{d_i^\star} \); 
\item  Squared Relative difference (sq-rel): 
 \( \frac{1}{K}\sum_{i=1}^K\frac{||\tilde{d}_i - d_i^\star||^2}{d_i^\star} \); 
\item Root mean squared error (rms): 
 \( \sqrt{\frac{1}{K}\sum_{i=1}^K(\tilde{d}_i - d_i^\star)^2} \);
\item Mean log10 error (log-rms): 
\( \sqrt{\frac{1}{K}\sum_{i=1}^K \Vert \log_{10}(\tilde{d}_i) - \log_{10}(d_i^\star) \Vert^2 }\);
\item  Accuracy with threshold $t$: percentage (\%) of $d_i^\star$, subject to $\max (\frac{d_i^\star}{\tilde{d}_i}, \frac{\tilde{d}_i}{d_i^\star}) = 
\delta < t~(t \in [1.25, 1.25^2, 1.25^3])$.
\end{itemize}
Where $\tilde{d}_i$ and $d_i^\star$ is the ground-truth depth and the estimated depth at pixel $i$ respectively; $K$ is the total number of pixels of the test images.

\par\noindent\textbf{Evaluation Protocol on Semantic Segmentation.} As for semantic segmentation, 
we consider two metrics~\cite{zhou2017scene, zhang2018context}, \ie~pixel accuracy (pixAcc) and mean intersection over union (mIoU), averaged over classes.
The normal prediction performance is evaluated with five metrics. We compute the per-pixel angle distance between prediction and ground-truth, then compute mean and median for valid pixels with given ground-truth normal. 
\par\noindent\textbf{Evaluation Protocol on Surface Normal Estimation.} For the evaluation of surface normal estimation, we utilize five standard evaluation metrics~\cite{fouhey2013data}, \ie~mean and median angle distance between prediction and ground-truth for valid pixels, and the fraction of pixels with angle difference with ground-truth less than $t$ ($t\in[11.25^{\circ}, 22.5^{\circ}, 30^{\circ}]$. 

\subsection{Implementation Details} 
The proposed VISTA-Net is implemented in~\textit{Pytorch}. The experiments are conducted on four Nvidia Quadro RTX 6000 GPUs, each with 24~GB memory. The ResNet-101 architecture pretrained on ImageNet~\cite{deng2009imagenet} is considered in the experiments for initializing the backbone network of {VISTA-Net}. Our model can be used for effective deep feature learning in  multi-scale contexts. To boost the performance, following previous works~\cite{xie2015holistically,xu2017learningdeep}, we also consider multi-features produced from different convolutional blocks of a backbone CNN (\eg~{res3c}, {ref4f}, {ref5d} of a ResNet-50). 
{In detail, {ref5d} is chosen as the receiving feature, $f_r$, while {res3c}, {ref4f}, {ref5d} are taken up as emitting features, $f_e$, in all tasks.}

For the semantic segmentation task, we use a learning rate of 0.001 with a momentum of 0.9 and a weight decay of 0.0001 using a polynomial learning rate scheduler as previously done in~\cite{zhang2018context,chen2016deeplab}. For the the monocular depth estimation task, the learning rate is set to $10^{-4}$ with weight decay of 0.01. The Adam optimizer is used in all our experiments with a batch size of 8 for monocular depth estimation and 16  for semantic segmentation and surface normal. 
The total training epochs are set to 50 for depth prediction, to 150 for semantic segmentation, 20 for surface normal and to 500 for the Cityscapes dataset. {The default value of the rank is 1 in all tasks, and an ablation study on $T$ is shown in Table~\ref{tab:pcontext efficiency}.}

\begin{figure}[!t]
\centering
    \begin{minipage}{0.24\linewidth}
        \centering
        \includegraphics[width=0.993\textwidth,height=0.5in]{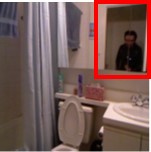}\\
        \includegraphics[width=0.993\textwidth,height=0.5in]{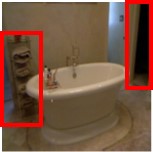}\\
        \includegraphics[width=0.993\textwidth,height=0.5in]{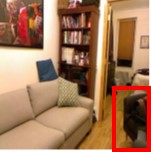}\\  
        \includegraphics[width=0.993\textwidth,height=0.5in]{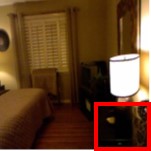}\\
        \includegraphics[width=0.993\textwidth,height=0.5in]{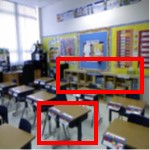}\\   
    \footnotesize (a) Image    
    \end{minipage}%
    \begin{minipage}{0.24\linewidth}
        \centering
        \includegraphics[width=0.993\textwidth,height=0.5in]{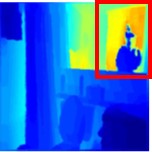}\\
        \includegraphics[width=0.993\textwidth,height=0.5in]{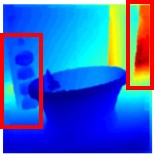}\\
        \includegraphics[width=0.993\textwidth,height=0.5in]{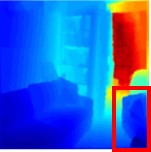}\\  
        \includegraphics[width=0.993\textwidth,height=0.5in]{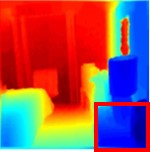}\\
        \includegraphics[width=0.993\textwidth,height=0.5in]{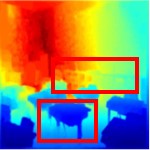}\\  
        \footnotesize (b) GT
    \end{minipage}%
    \begin{minipage}{0.24\linewidth}
        \centering
        \includegraphics[width=0.993\textwidth,height=0.5in]{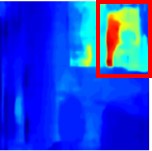}\\
        \includegraphics[width=0.993\textwidth,height=0.5in]{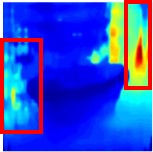}\\
        \includegraphics[width=0.993\textwidth,height=0.5in]{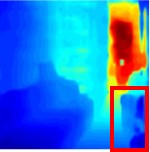}\\  
        \includegraphics[width=0.993\textwidth,height=0.5in]{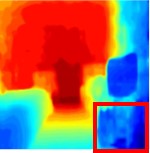}\\
        \includegraphics[width=0.993\textwidth,height=0.5in]{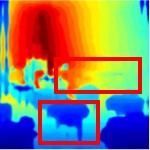}\\ 
        \footnotesize (c) DORN
    \end{minipage}%
    \begin{minipage}{0.24\linewidth}
        \centering
        \includegraphics[width=0.993\textwidth,height=0.5in]{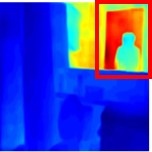}\\
        \includegraphics[width=0.993\textwidth,height=0.5in]{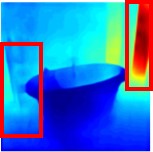}\\
        \includegraphics[width=0.993\textwidth,height=0.5in]{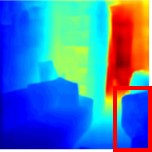}\\  
        \includegraphics[width=0.993\textwidth,height=0.5in]{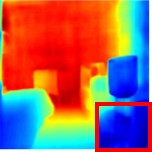}\\
        \includegraphics[width=0.993\textwidth,height=0.5in]{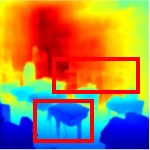}\\ 
        \footnotesize (d) \method

    \end{minipage}%
 
\centering
\caption{ Qualitative examples on NYU dataset.}
\label{fig:vis_nyu}
\end{figure}

\begin{table}[]
\centering
\caption{Depth Estimation: NYU dataset.}
\resizebox{1\linewidth}{!}{
\label{tab:overall_nyu}
\begin{tabular}{lcccccc}
\toprule[1.2pt]
\multirow{2.5}{*}{Method} & \multicolumn{3}{c}{Error (lower is better)} & \multicolumn{3}{c}{Accuracy (higher is better)} \\ \cmidrule{2-7} 
 & rel & log10 & rms  & $\delta \textless 1.25$ & $\delta \textless 1.25^2$ & $\delta \textless 1.25^3$ \\ 
 \midrule
    PAD-Net~\cite{xu2018pad} & 0.214 & 0.091 & 0.792 & 0.643 & 0.902 & 0.977 \\
    Li~\etal~\cite{li2017two}& 0.152 & 0.064 & 0.611 & 0.789 & 0.955 & 0.988\\
    CLIFFNet~\cite{wang2020cliffnet} & 0.128 &	0.171 &	0.493 &	0.844 &	0.964 &	0.991\\
    Laina~\etal \cite{laina2016deeper} & 0.127 & 0.055 & 0.573 & 0.811 & 0.953 & 0.988\\
    MS-CRF~\cite{xu2017multi} & 0.121 & 0.052 & 0.586 & 0.811 & 0.954 & 0.987\\
    Lee~\etal \cite{lee2020multi} & 0.119 & 0.050 &	- &	0.870 &	0.974 & 0.993\\
    AG-CRF~\cite{xu2017learningdeep} & 0.112 & 0.051 & 0.526 & 0.818 & 0.960 & 0.989\\
    DORN~\cite{fu2018deep} & 0.115 & 0.051 & 0.509 & 0.828 & 0.965 & 0.992 \\
    Xia~\etal \cite{xia2020generating} & 0.116 & - & 0.512 & 0.861 & 0.969 & 0.991 \\
    Yin~\etal \cite{yin2019enforcing} & \textbf{0.108} & {0.048} & 0.416 & 0.875 & 0.976 & 0.994 \\
    Lee~\etal \cite{lee2019big} & 0.113 & 0.049 & 0.407 & 0.871 & 0.977 & 0.995 \\
    \midrule
    VISTA-Net  & 0.111 & {0.048} & {0.393} & {0.881} & {0.979} & \textbf{0.996}\\
\bottomrule[1.2pt]
\end{tabular}}
\end{table}

\subsection{Experimental Results and Analysis}
\label{sec:results}
\noindent \textbf{Monocular Depth Estimation.} 
Comparative results on KITTI dataset are shown in Table \ref{tab:overall_kitti}. We propose a comparison with state of the art models such as \cite{eigen2014depth,ranjan2019competitive,bian2019unsupervised,godard2019digging,fu2018deep,yin2019enforcing,lee2019big,guizilini20203d,xu2020probabilistic}. In addition we demonstrate the effectiveness of our \method{} comparing with MS-CRF~\cite{xu2017multi}, a previous approach which exploit a probabilistic framework for multi-scale feature learning but does not consider an attention mechanisms. Our approach is superior, thus demonstrating the effectiveness of the proposed structured attention model. 
We also compare with AG-CRF~\cite{xu2017learningdeep} and PGA-Net \cite{xu2020probabilistic}. Also in this case \method{} outperforms the competitors confirming the importance of having a joint structured spatial- and channel-wise attention model.
Note that AG-CRF~\cite{xu2017learningdeep}, PGA-Net \cite{xu2020probabilistic} and \method{} are compared using the same backbone.
In order to demonstrate the competitiveness of our approach in an indoor scenario we also report the results on NYUD-V2 dataset in Table~\ref{tab:overall_nyu}. Similarly to the experiments on KITTI, \method{} outperforms both state of the art approaches and previous methods based on attention gates and CRFs \cite{xu2017multi,xu2017learningdeep}.
{Both Table~\ref{tab:overall_kitti} and ~\ref{tab:overall_nyu} also prove that our structured attention can merge more low-level information and can make the network learn a more efficient deep representation. It is essential for dense pixel-wise task network to predict better results.}
In Fig.~\ref{fig:vis_kitti} is shown a qualitative comparison of our method with DORN~\cite{fu2018deep}. Results indicate that \method ~generates better depth maps, in particular one can appreciate the opening of the sky and the smoothness of the prediction on the sides. Fig.~\ref{fig:vis_nyu} shows a similar comparison done on NYU dataset. The same accuracy in the prediction is visible also in this case, objects are more distinguishable w.r.t. DORN (e.g. the bathtub in row 2 and the desks in row 5).

\begin{figure}[t]
\centering
    \begin{minipage}{0.191\linewidth}
        \centering
        \includegraphics[width=0.993\textwidth,height=0.7in]{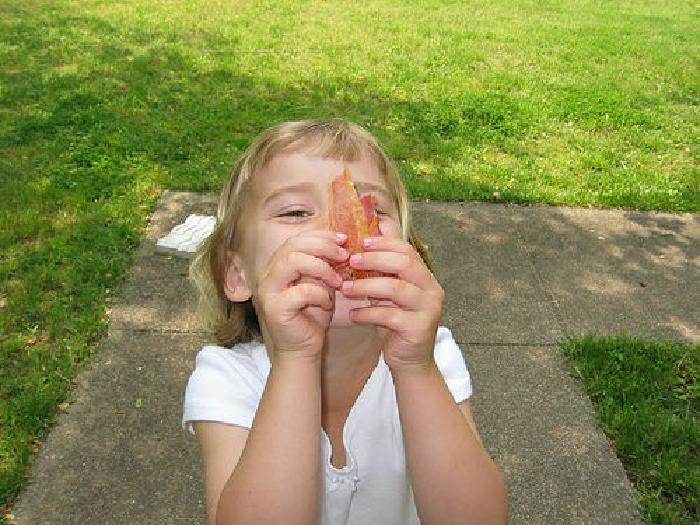}\\
        \includegraphics[width=0.993\textwidth, height=0.7in]{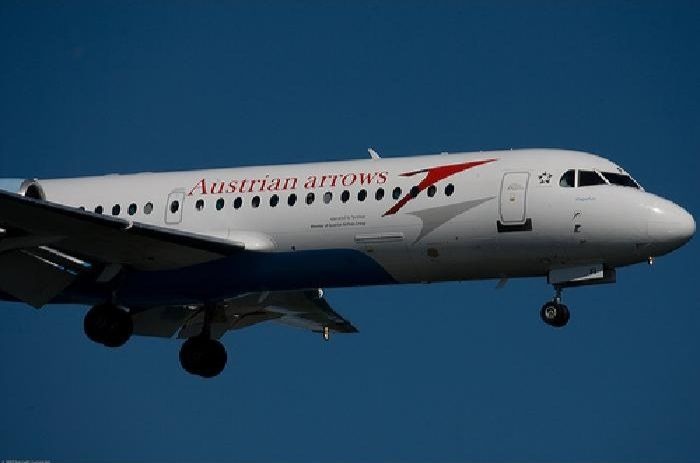}\\
        \includegraphics[width=0.993\textwidth, height=0.7in]{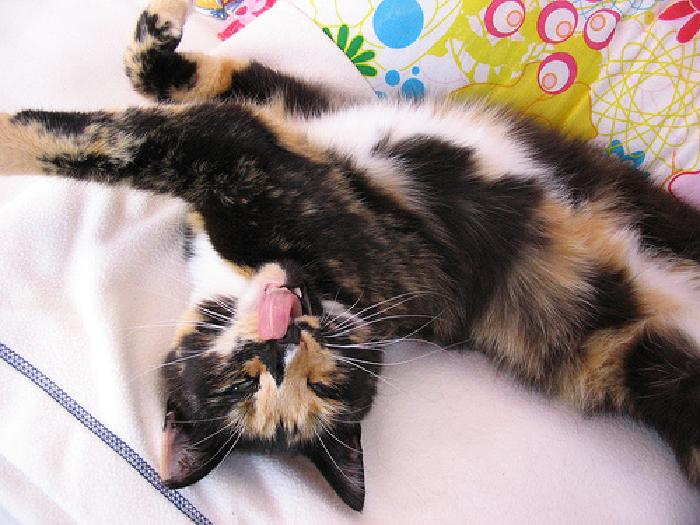}\\
        \includegraphics[width=0.993\textwidth, height=0.7in]{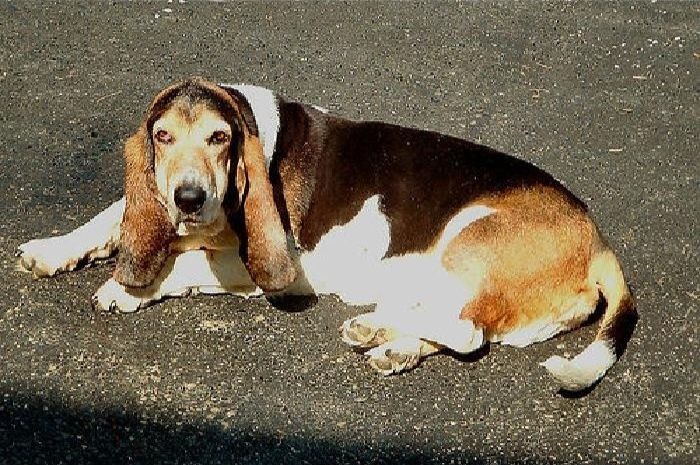}\\
        \includegraphics[width=0.993\textwidth,height=0.7in]{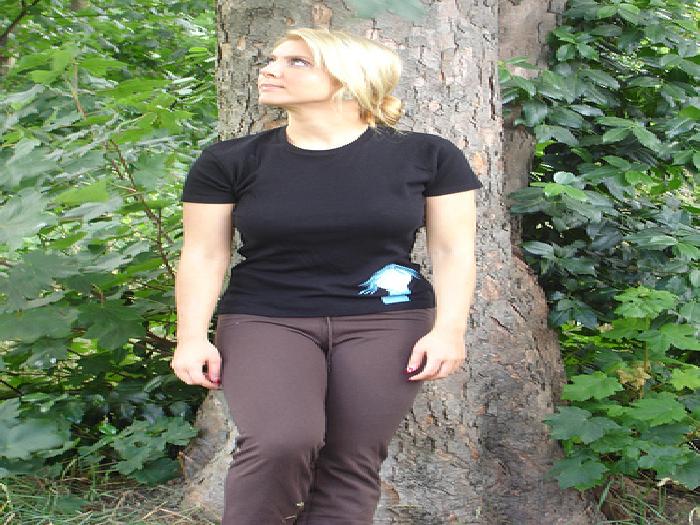}\\        
        \footnotesize (a) Image
    \end{minipage}%
    \begin{minipage}{0.191\linewidth}
        \centering
        \includegraphics[width=0.993\textwidth,height=0.7in]{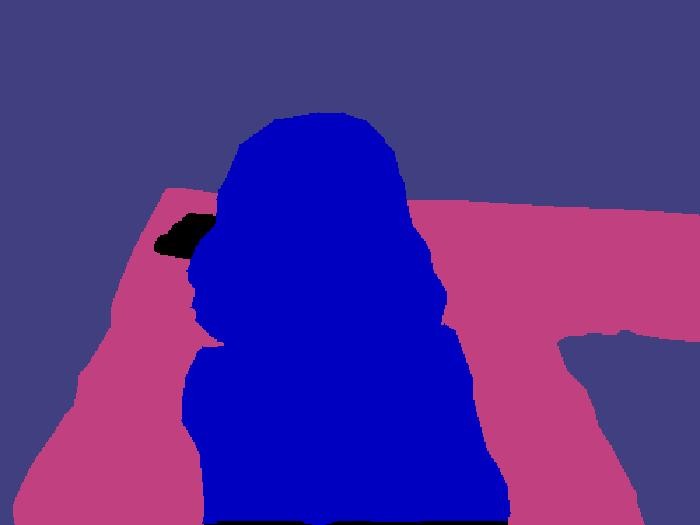}\\
        \includegraphics[width=0.993\textwidth, height=0.7in]{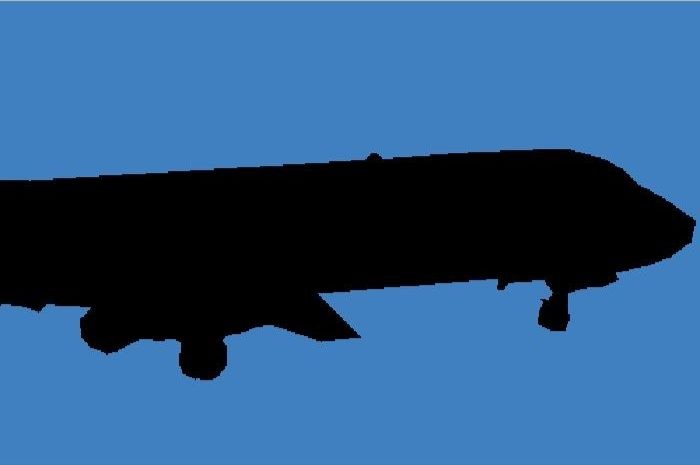}\\
        \includegraphics[width=0.993\textwidth, height=0.7in]{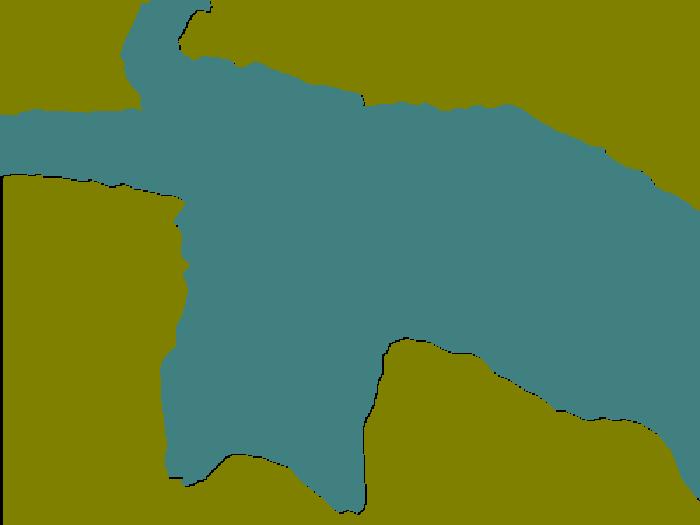}\\
        \includegraphics[width=0.993\textwidth, height=0.7in]{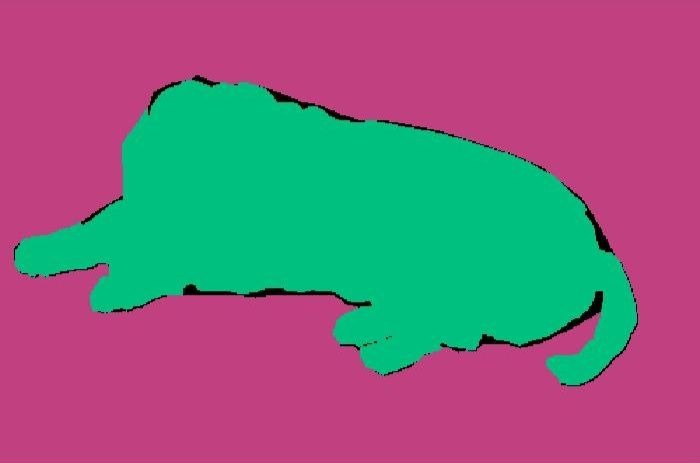}\\
        \includegraphics[width=0.993\textwidth,height=0.7in]{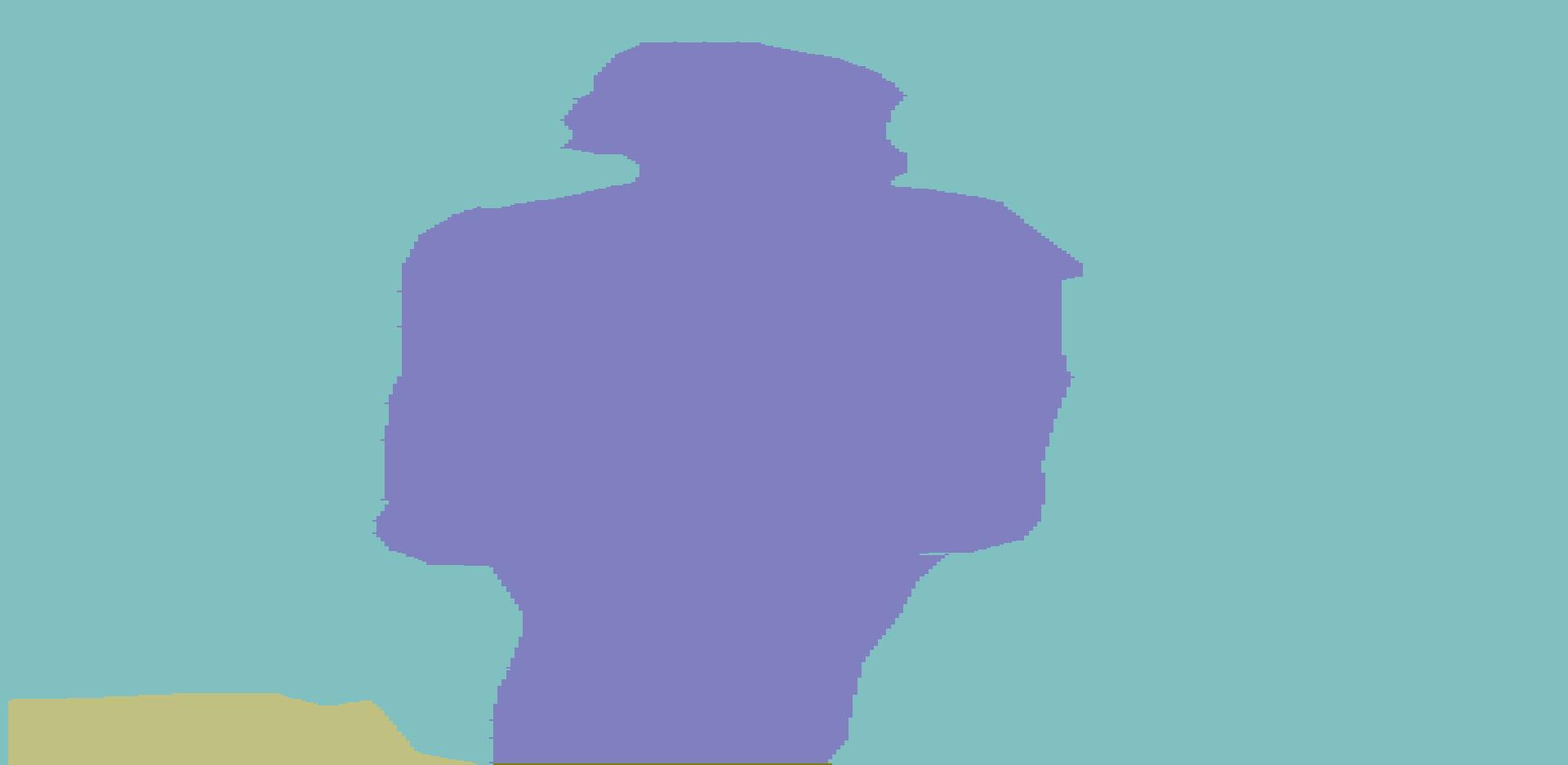}\\  
        \footnotesize (b) GT
    \end{minipage}%
    \begin{minipage}{0.191\linewidth}
        \centering
        \includegraphics[width=0.993\textwidth,height=0.7in]{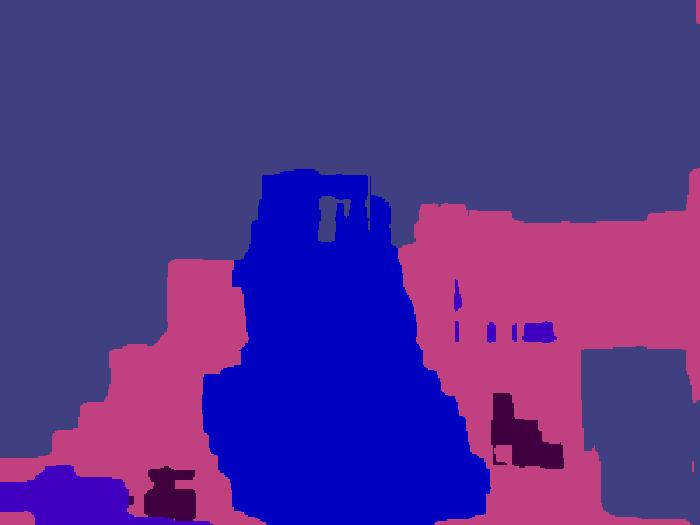}\\
        \includegraphics[width=0.993\textwidth, height=0.7in]{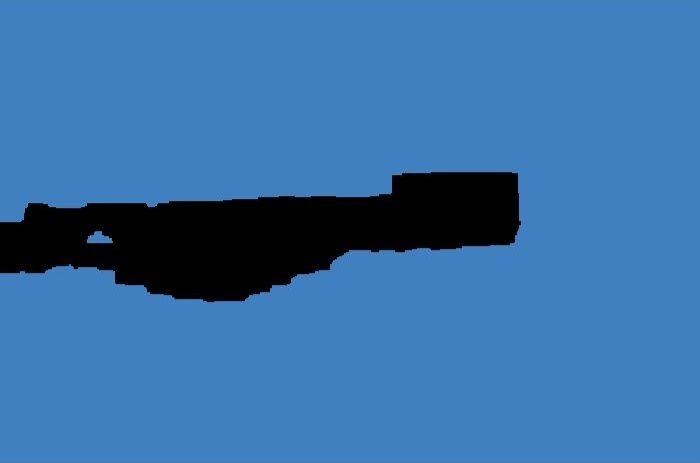}\\
        \vspace{0.02cm}
        \includegraphics[width=0.993\textwidth, height=0.7in]{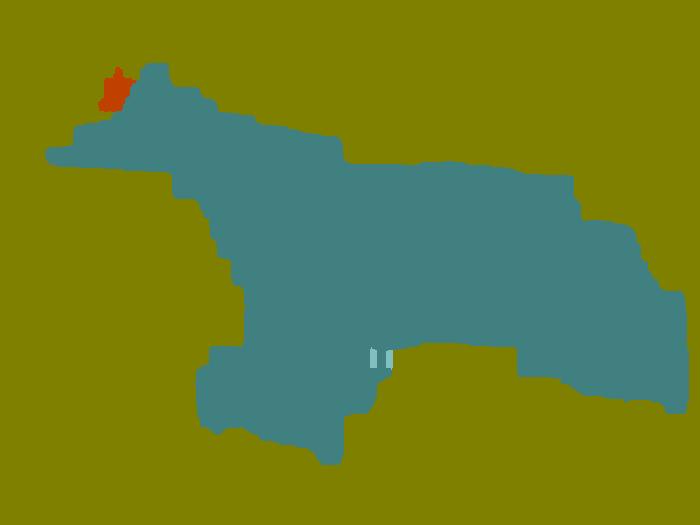}\\
        \includegraphics[width=0.993\textwidth, height=0.7in]{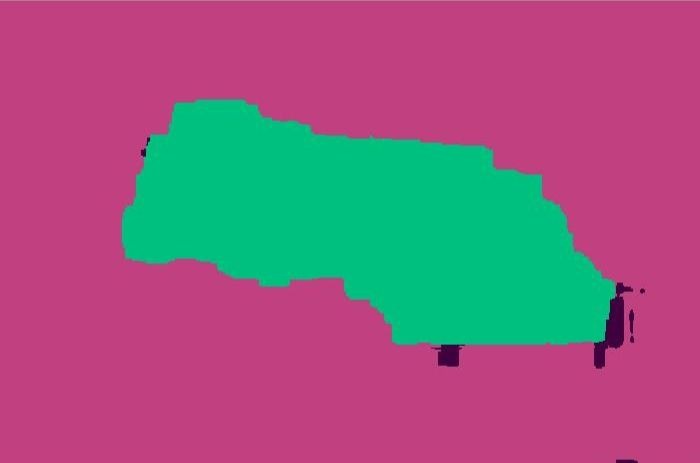}\\
        \includegraphics[width=0.993\textwidth,height=0.7in]{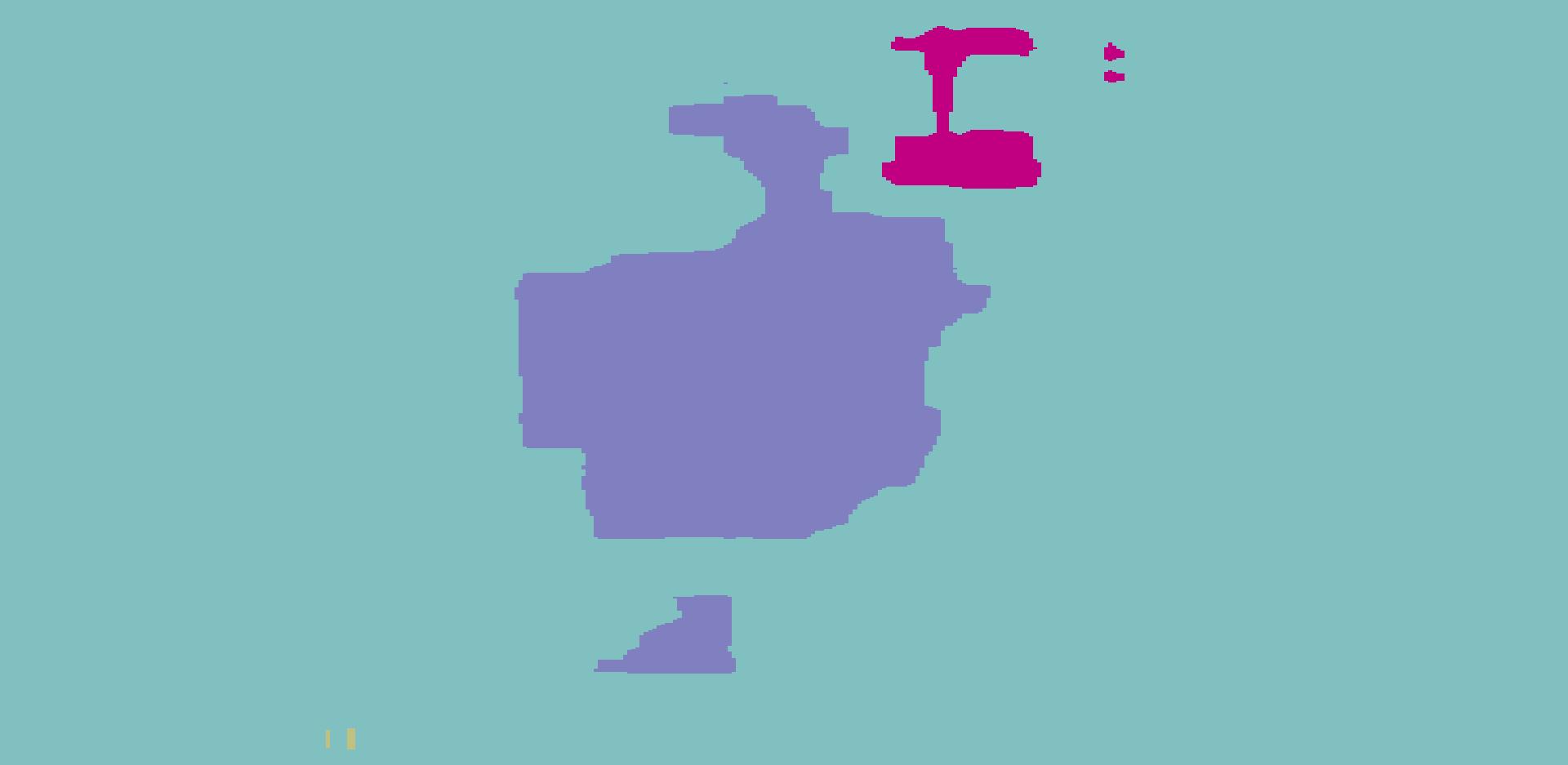}\\   
        \footnotesize (c) rank=0
    \end{minipage}%
    \begin{minipage}{0.191\linewidth}
        \centering
        \includegraphics[width=0.993\textwidth, height=0.7in]{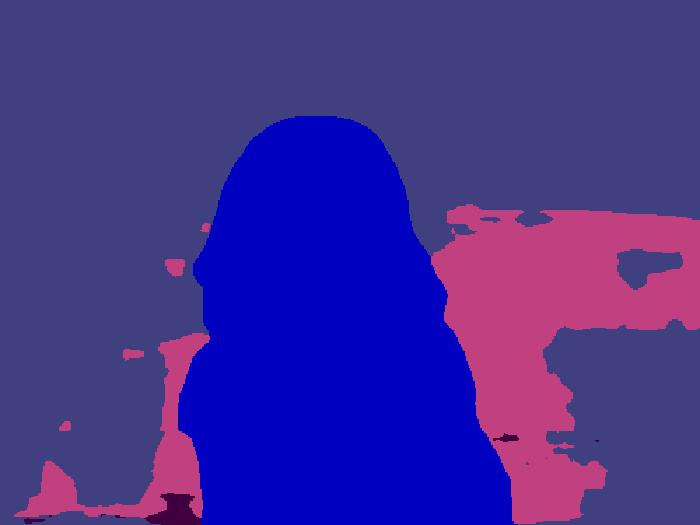}\\
        \includegraphics[width=0.993\textwidth, height=0.7in]{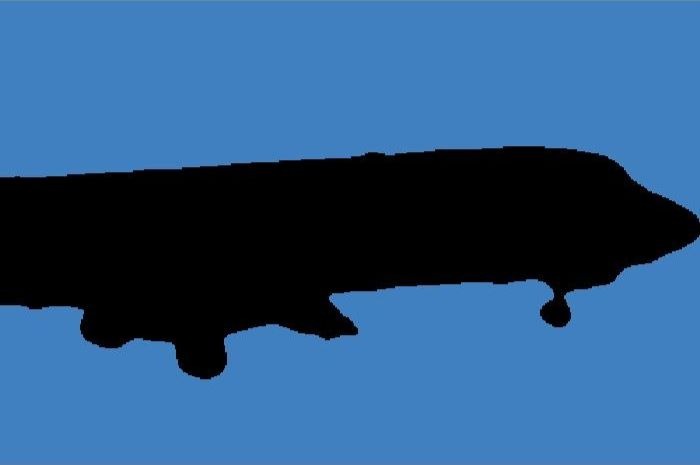}\\
        \includegraphics[width=0.993\textwidth, height=0.7in]{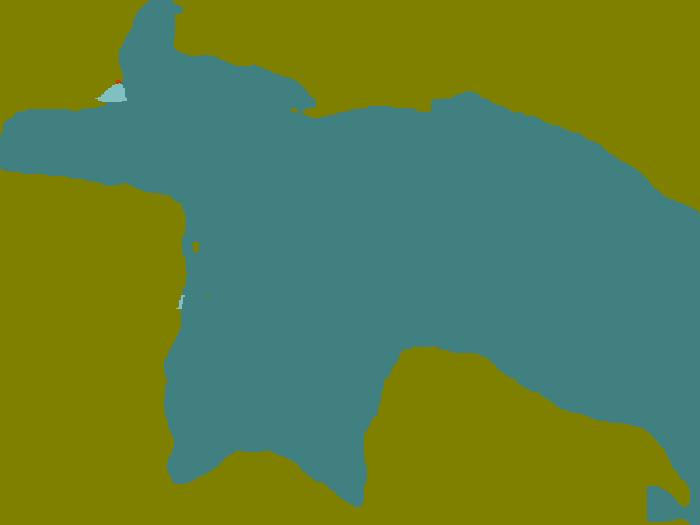}\\
        \includegraphics[width=0.993\textwidth, height=0.7in]{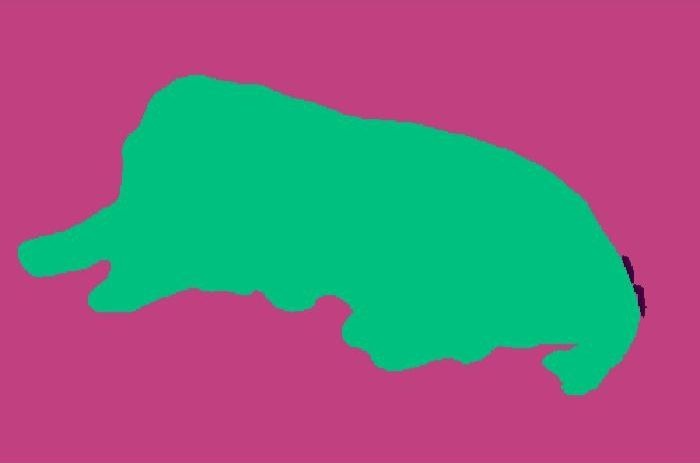}\\
        \includegraphics[width=0.993\textwidth, height=0.7in]{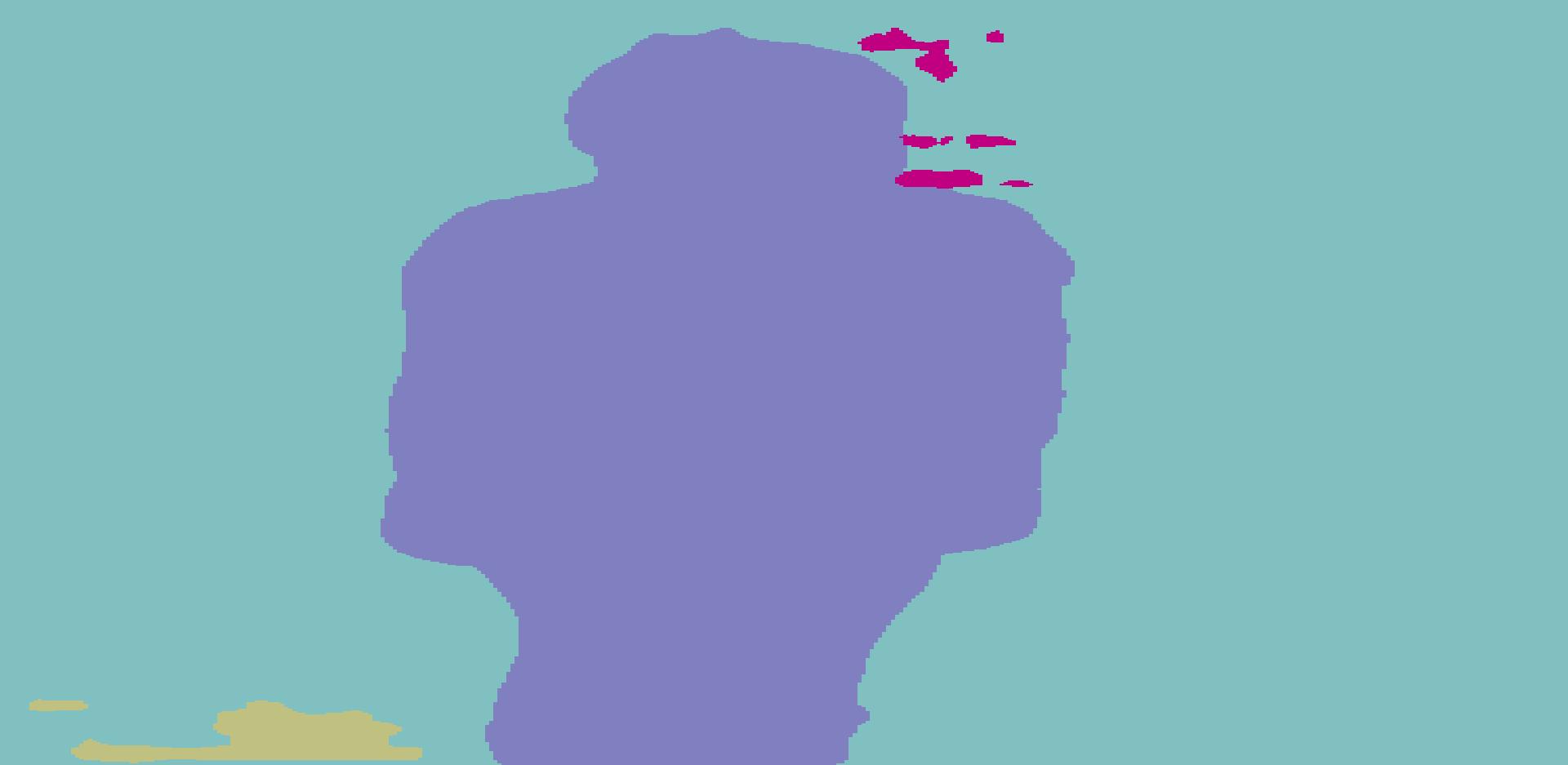}\\   
        \footnotesize (d) rank=1
    \end{minipage}%
    \begin{minipage}{0.191\linewidth}
        \centering
        \includegraphics[width=0.993\textwidth, height=0.7in]{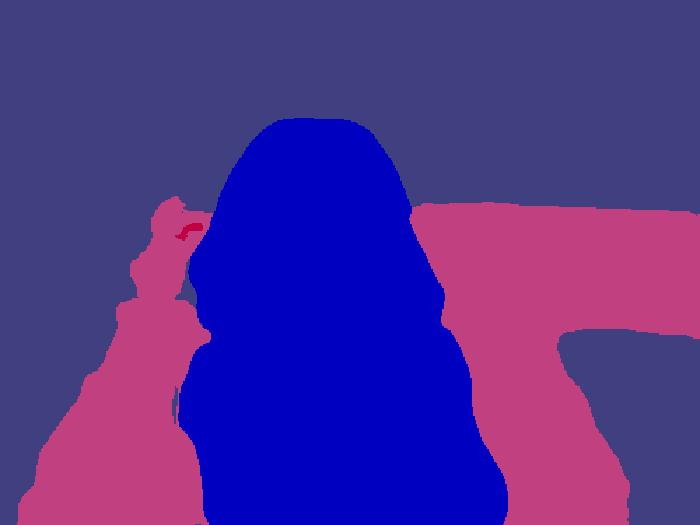}\\
        \includegraphics[width=0.993\textwidth, height=0.7in]{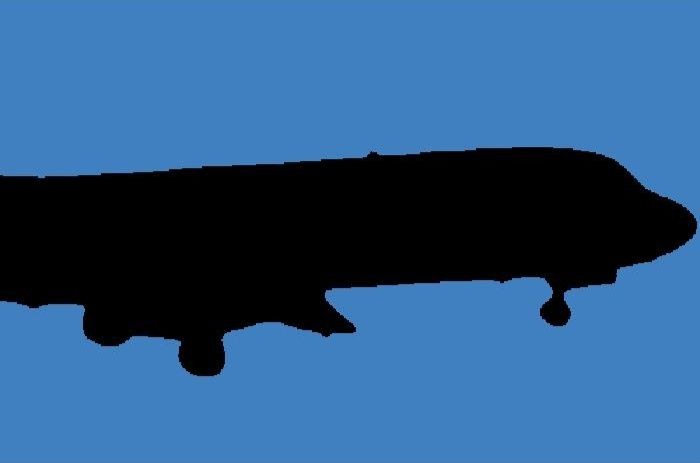}\\
        \includegraphics[width=0.993\textwidth, height=0.7in]{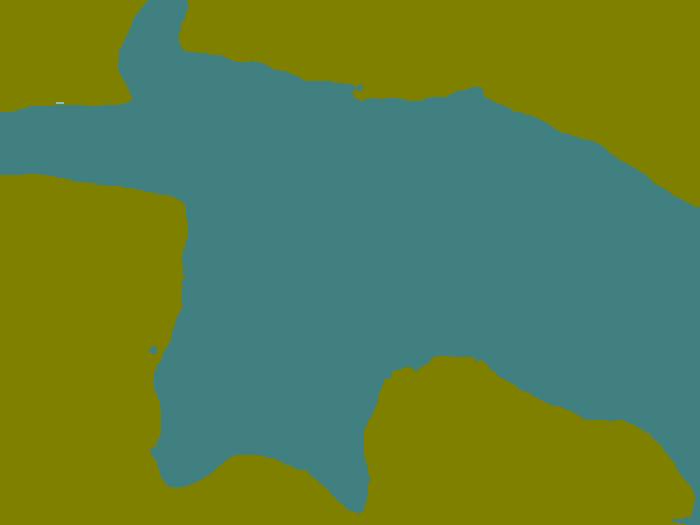}\\
        \includegraphics[width=0.993\textwidth, height=0.7in]{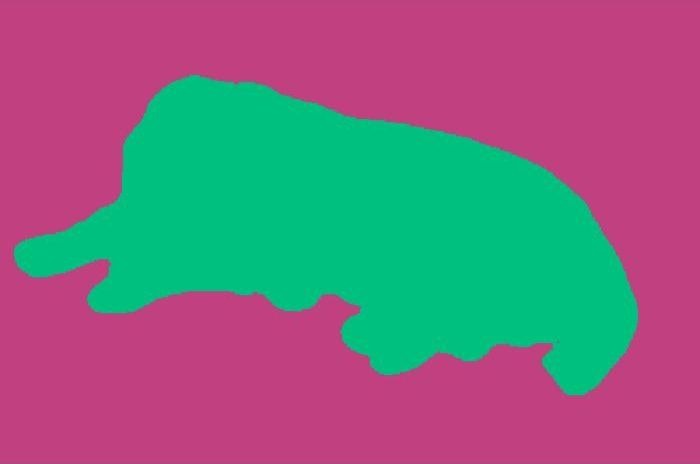}\\
        \includegraphics[width=0.993\textwidth, height=0.7in]{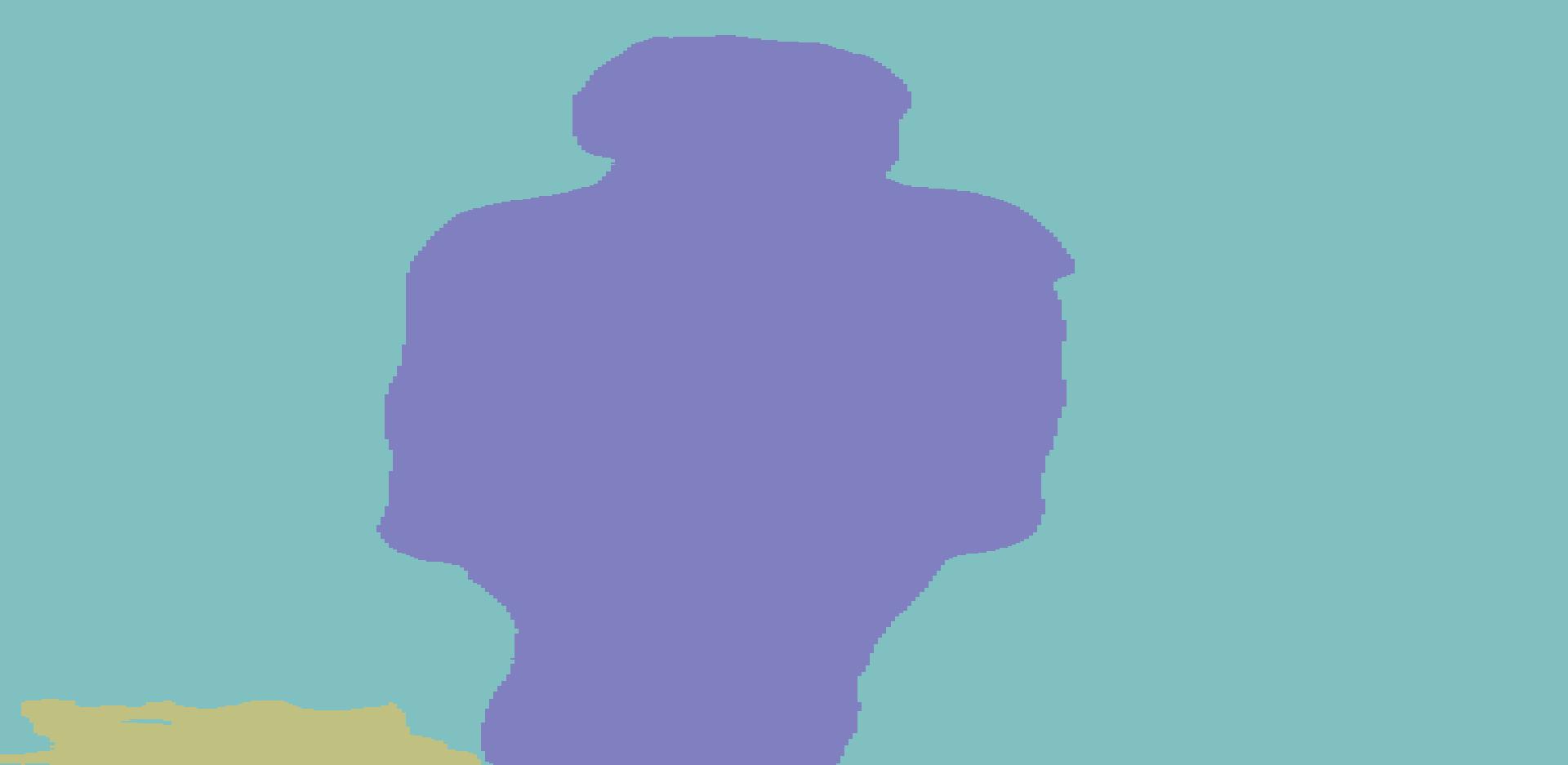}\\   
        \footnotesize (e) rank=9
    \end{minipage}%
\centering
\caption{Qualitative examples on the Pascal-Context dataset.}
\label{fig:vis_pcontext}
\end{figure}

\begin{table}[t]
   \caption{
   Semantic Segmentation: PASCAL-Context. D-ResNet-101 denotes Dilated ResNet-101.
 }
\centering
\resizebox{0.45\textwidth}{!}{%
\begin{tabular}{lccc}
\toprule[1.2pt]
Method & Backbone & pixAcc\% & mIoU\% \\
\midrule
CFM (VGG+MCG)~\cite{dai2015convolutional} & VGG-16 & - & 34.4 \\
DeepLab-v2 ~\cite{chen2016deeplab} & VGG-16 & - & 37.6 \\
FCN-8s~\cite{long2015fully} & VGG-16 & 50.7 & 37.8 \\
BoxSup~\cite{dai2015boxsup} & VGG-16 & - & 40.5 \\
ConvPP-8s~\cite{xie2016top} & VGG-16 & - & 41.0 \\
PixelNet~\cite{bansal2017pixelnet} & VGG-16 & 51.5 & 41.4 \\
\midrule
HRNetV2~\cite{wang2020deep} & HRNetV2-W48 & - & 54.0 \\
{Wang~\etal~\cite{wang2021exploring}} & HRNetV2-W48& - & 55.1\\
{\method{}  }& HRNetV2-W48 &  \bf 82.1 & {\bf 56.7} \\ 
\midrule
EncNet~\cite{zhang2018context} & D-ResNet-101 & 79.2 & 51.7 \\
DANet~\cite{fu2019dual} & D-ResNet-101 & - & 52.6 \\
ANN~\cite{zhu2019asymmetric} &D-ResNet-101 & - & 52.8\\
SpyGR~\cite{li2020spatial} & ResNet-101 & - & 52.8\\
SANet~\cite{zhong2020squeeze} & ResNet-101 & 80.6 & 53.0\\
SVCNet~\cite{ding2019semantic} & ResNet-101 & - & 53.2 \\
CFNet~\cite{zhang2019co} & ResNet-101 & - & 54.0 \\
APCNet~\cite{he2019adaptive} & D-ResNet-101 & - & 54.7 \\
{OCR~\cite{YuanCW19}} & ResNet-101 & - & 54.8\\
PGA-Net~\cite{xu2020probabilistic} & D-ResNet-101 & {81.2} & 55.1 \\ 
\midrule
\method{}  & D-ResNet-101 &  81.1 & {55.4} \\ 
\bottomrule[1.2pt]
\end{tabular}}%
\label{tab:overall-pcontext}
\end{table}

\begin{figure}[t]
\centering
    \begin{minipage}{0.24\linewidth}
        \centering
        \includegraphics[width=0.993\textwidth,height=0.4in]{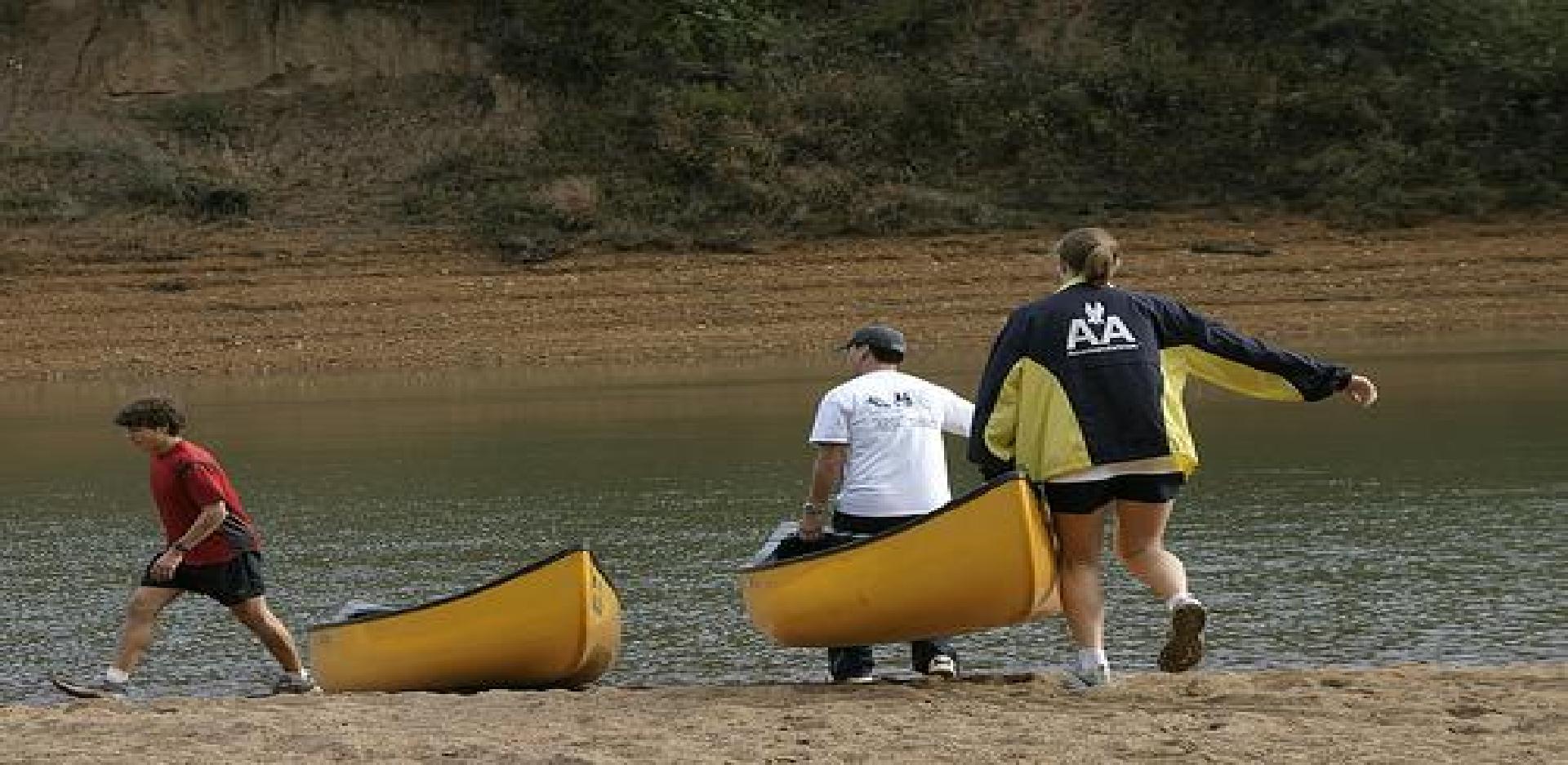}\\
        \includegraphics[width=0.993\textwidth, height=0.4in]{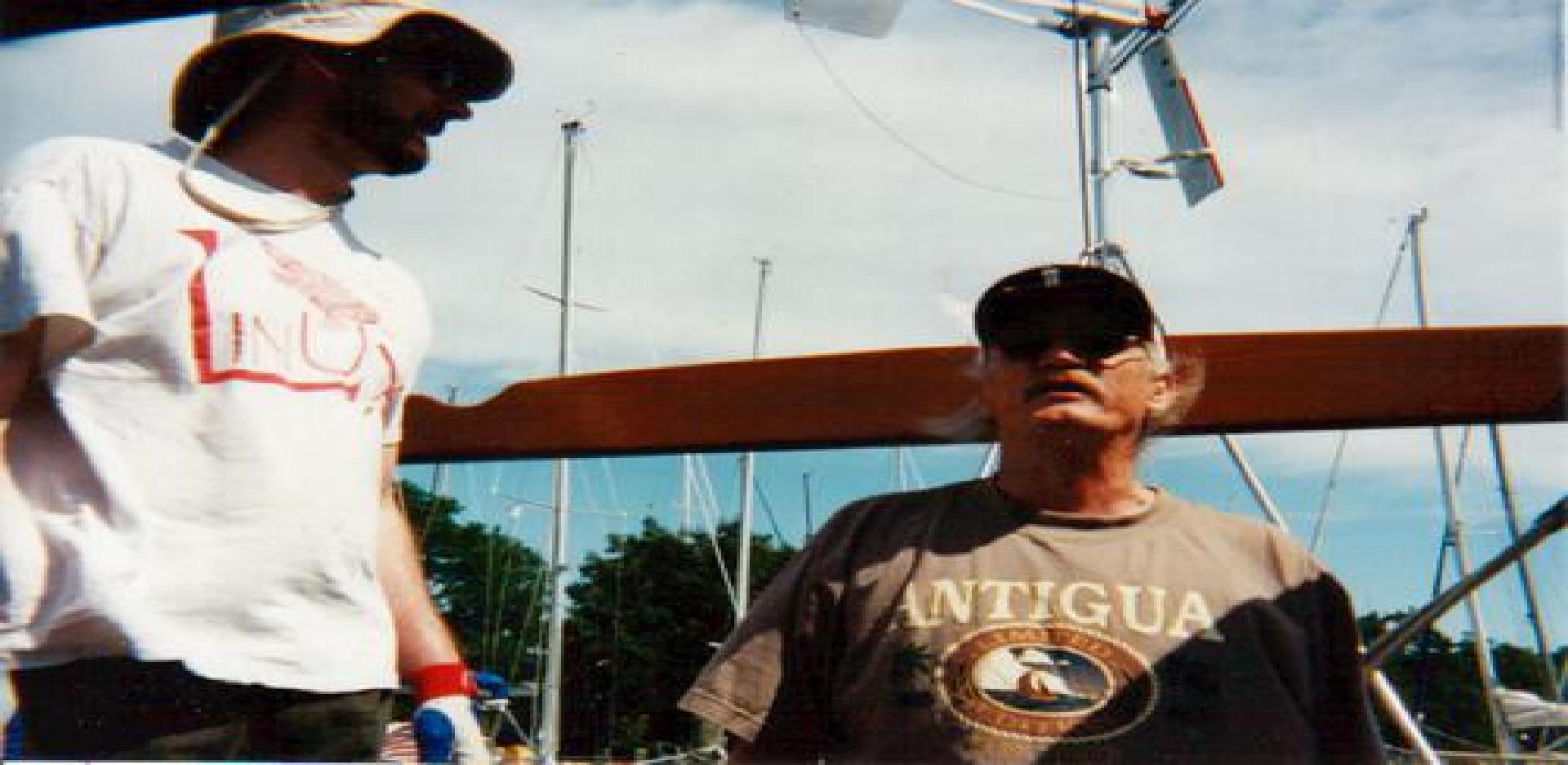}\\
        \includegraphics[width=0.993\textwidth, height=0.4in]{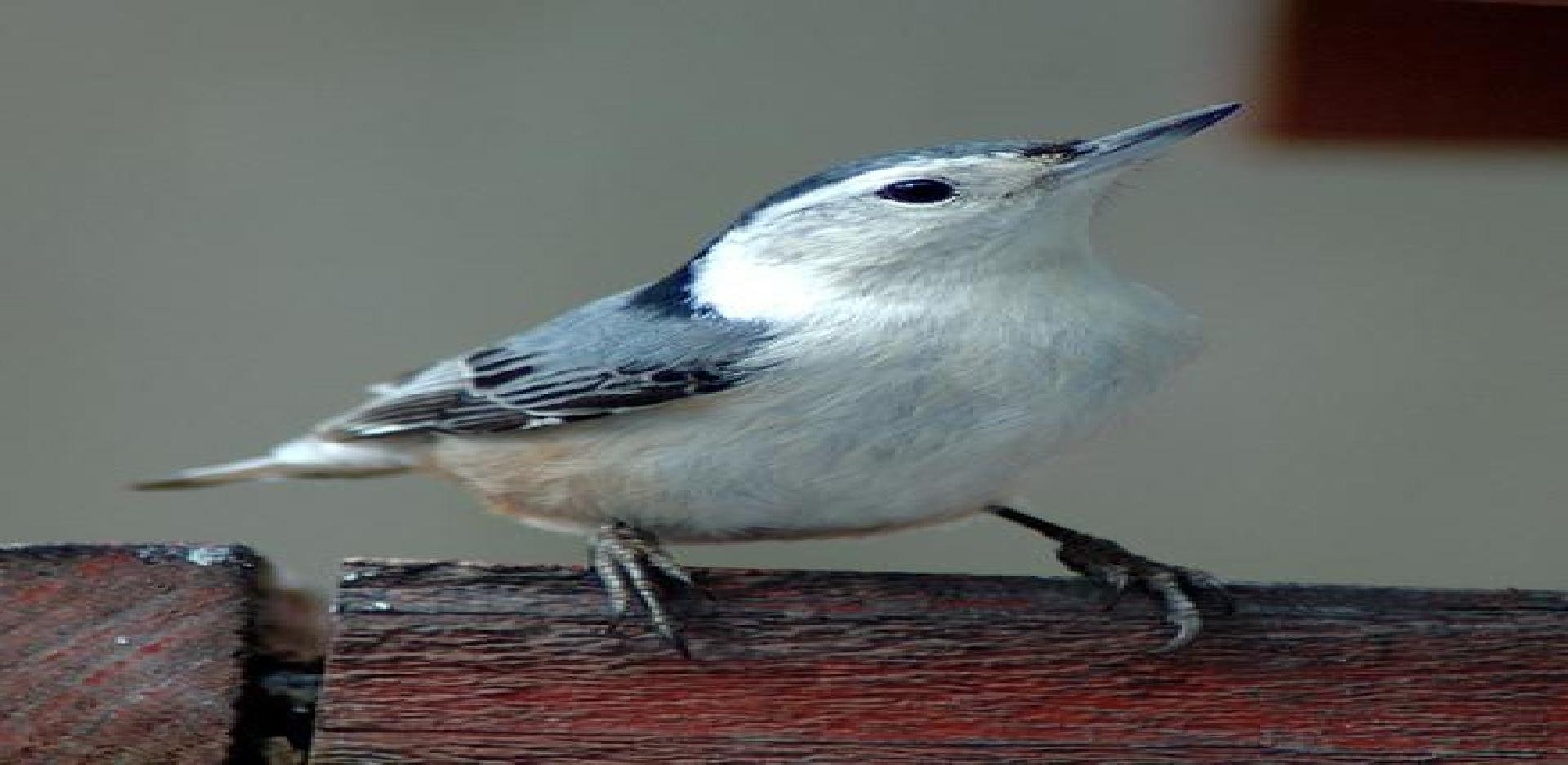}\\
        \includegraphics[width=0.993\textwidth, height=0.4in]{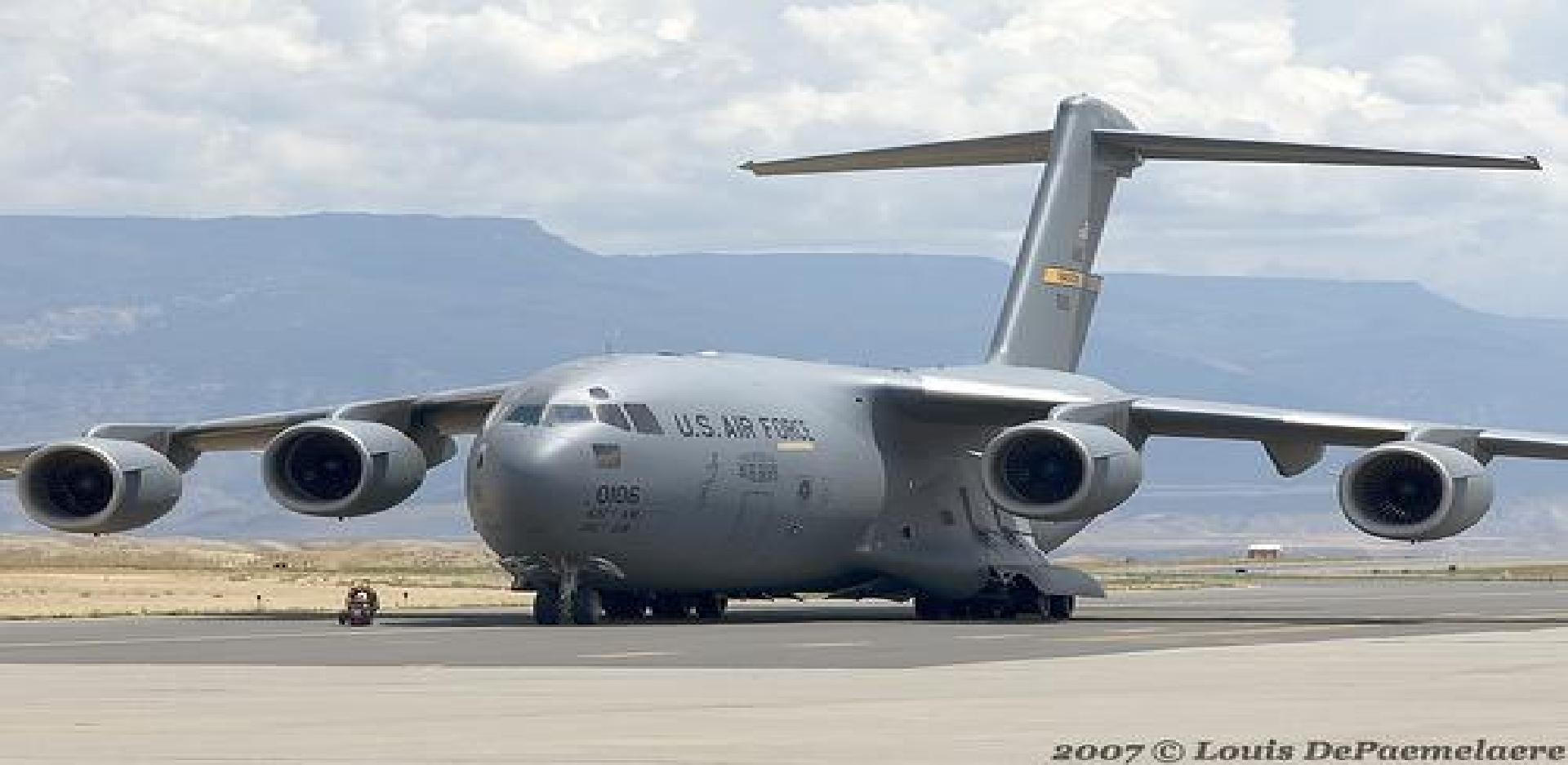}\\
        \footnotesize (a) Image
    \end{minipage}%
    \begin{minipage}{0.24\linewidth}
        \centering
        \includegraphics[width=0.993\textwidth,height=0.4in]{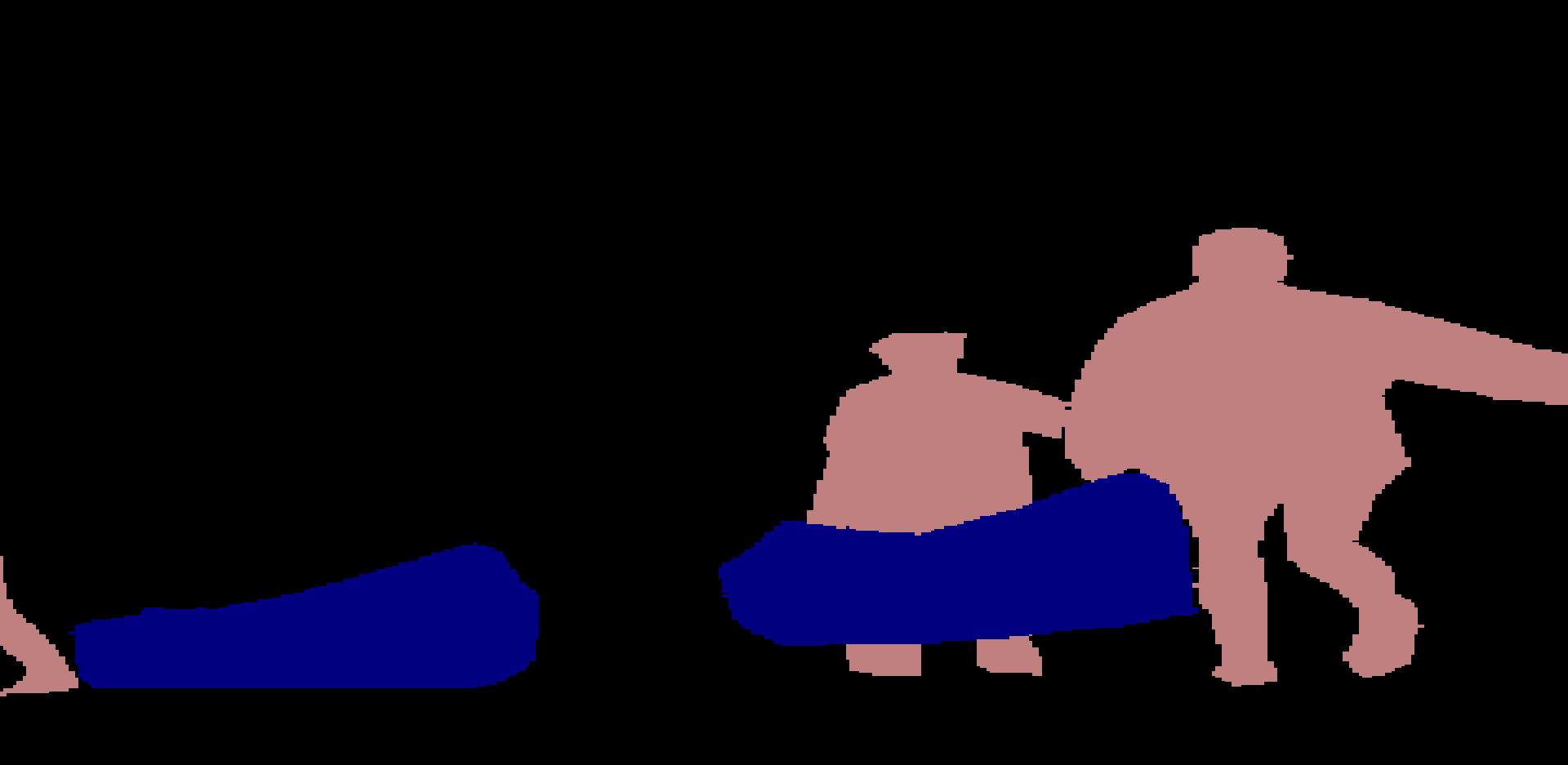}\\
        \includegraphics[width=0.993\textwidth, height=0.4in]{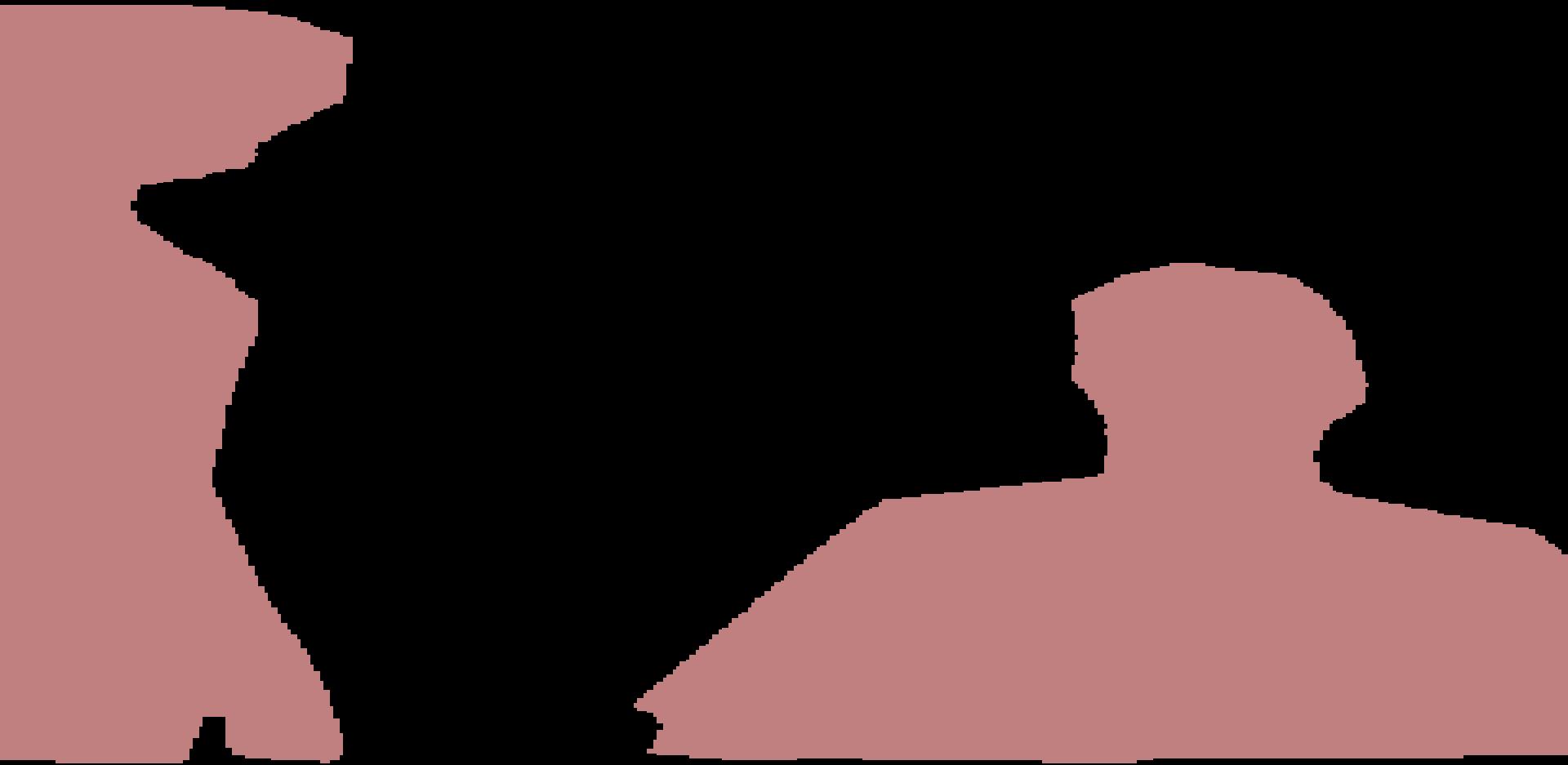}\\
        \includegraphics[width=0.993\textwidth, height=0.4in]{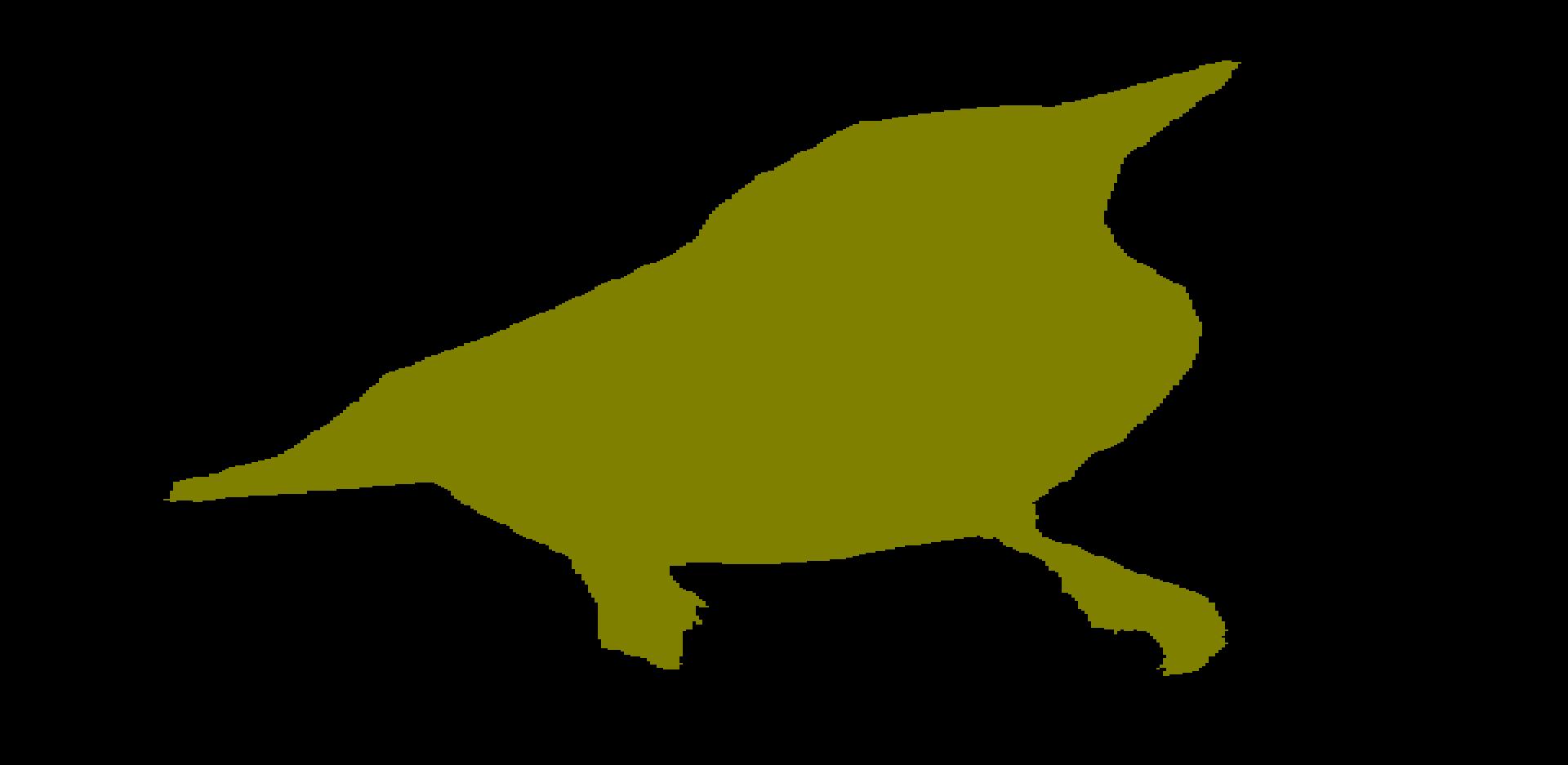}\\
        \includegraphics[width=0.993\textwidth, height=0.4in]{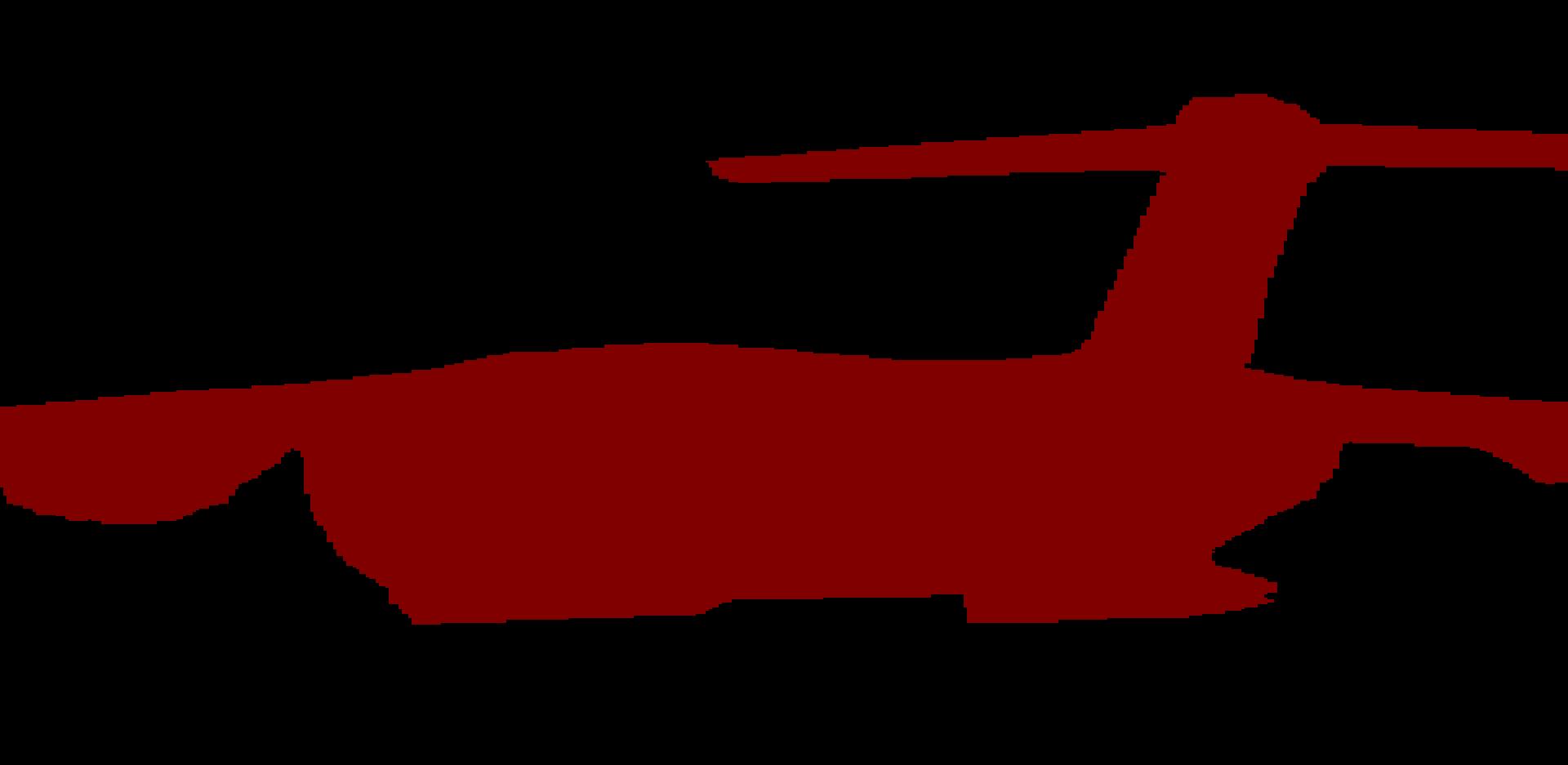}\\
        \footnotesize (b) Image

    \end{minipage}%
    \begin{minipage}{0.24\linewidth}
        \centering
        \includegraphics[width=0.993\textwidth,height=0.4in]{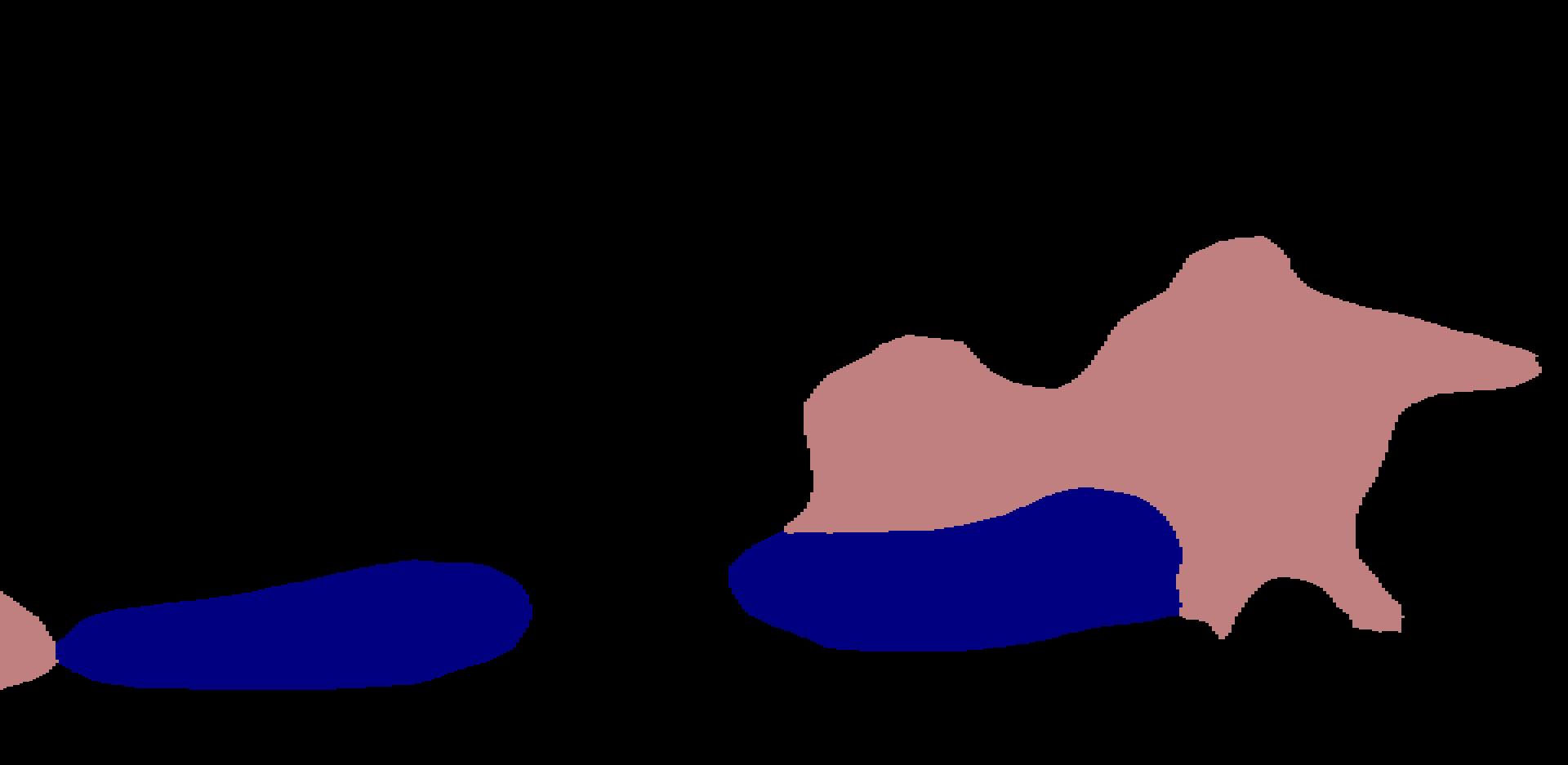}\\
        \includegraphics[width=0.993\textwidth, height=0.4in]{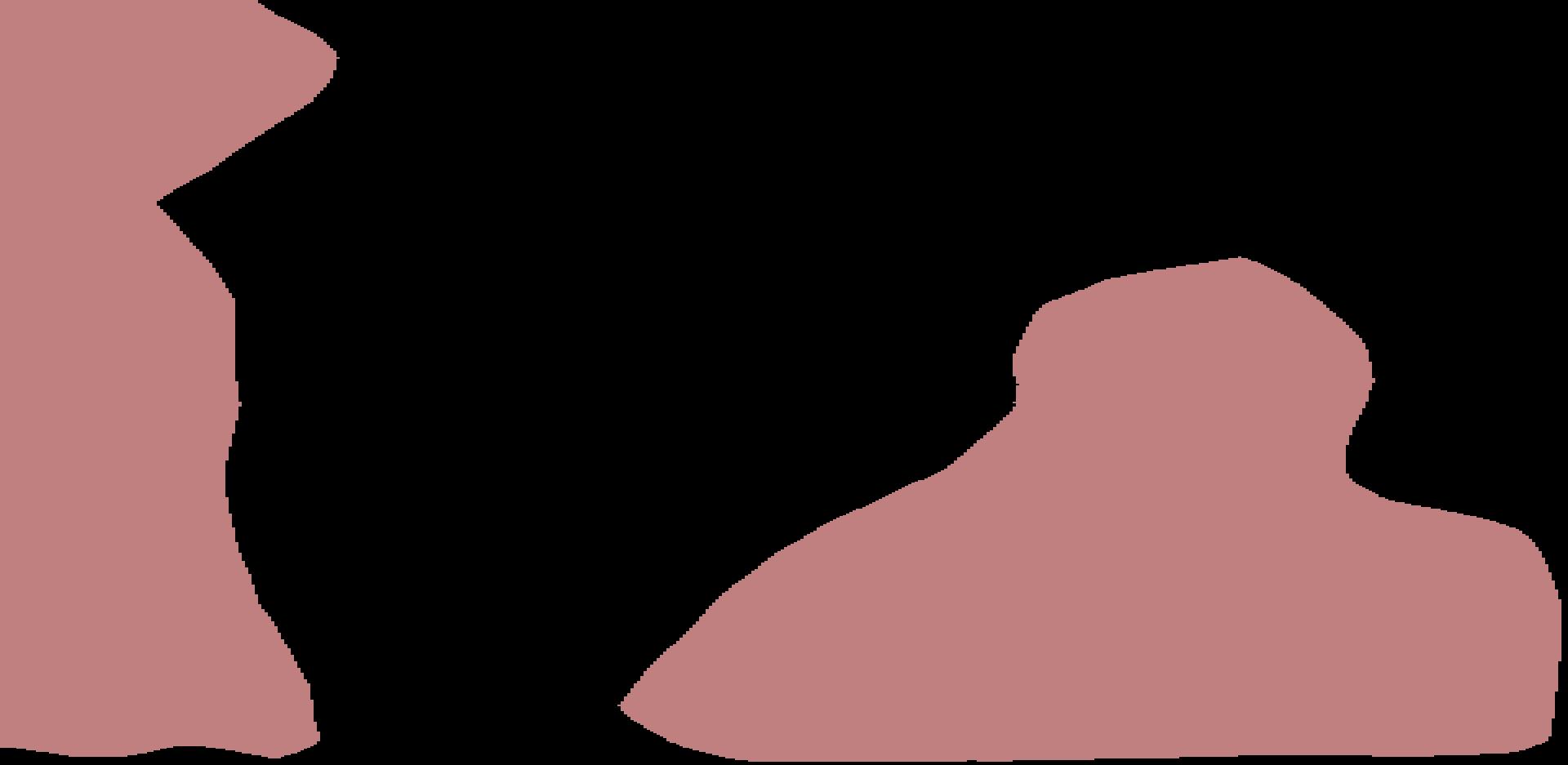}\\
        \includegraphics[width=0.993\textwidth, height=0.4in]{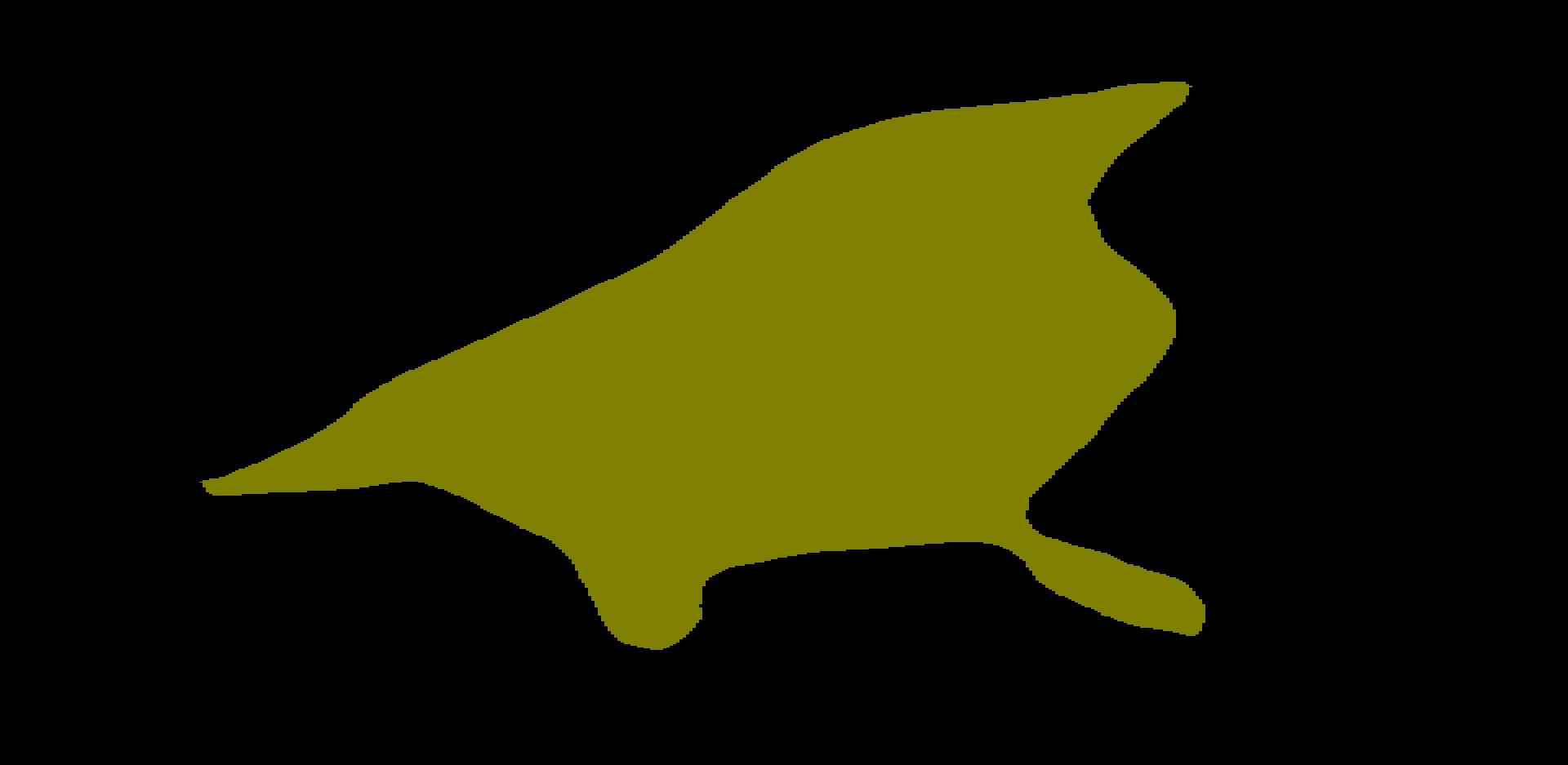}\\
        \includegraphics[width=0.993\textwidth, height=0.4in]{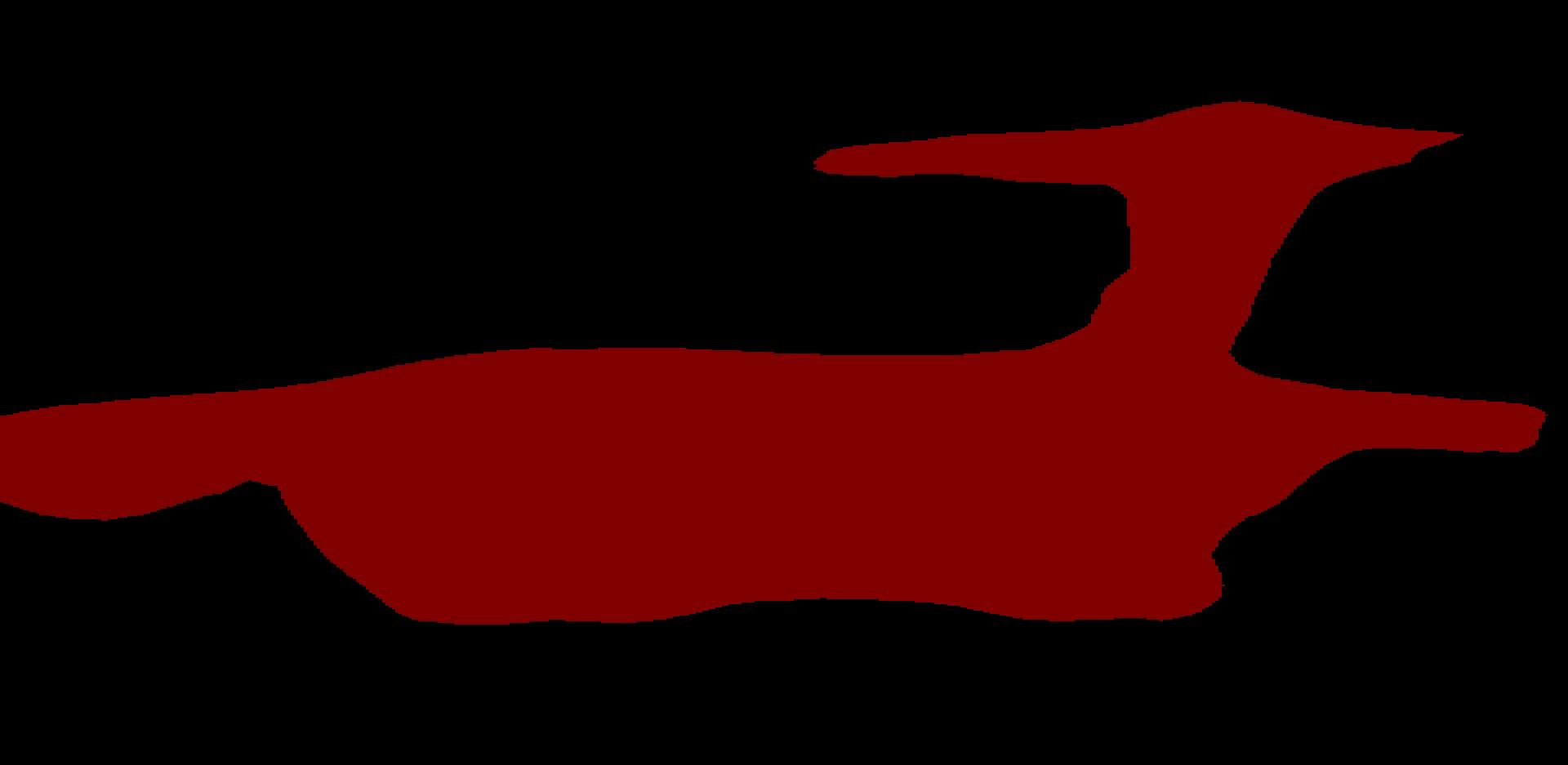}\\
        \footnotesize (c) Encnet

    \end{minipage}%
    \begin{minipage}{0.24\linewidth}
        \centering
        \includegraphics[width=0.993\textwidth,height=0.4in]{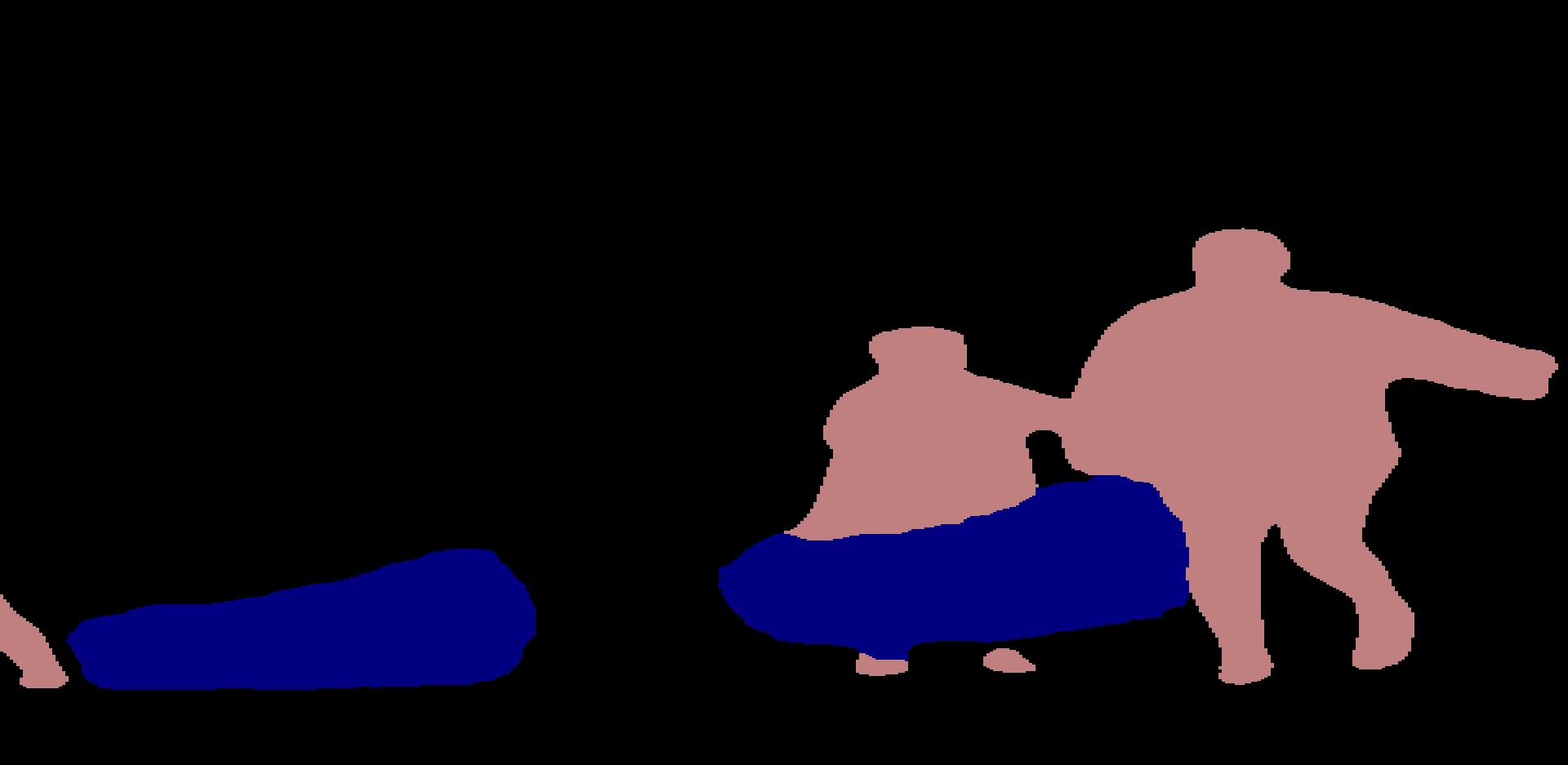}\\
        \includegraphics[width=0.993\textwidth, height=0.4in]{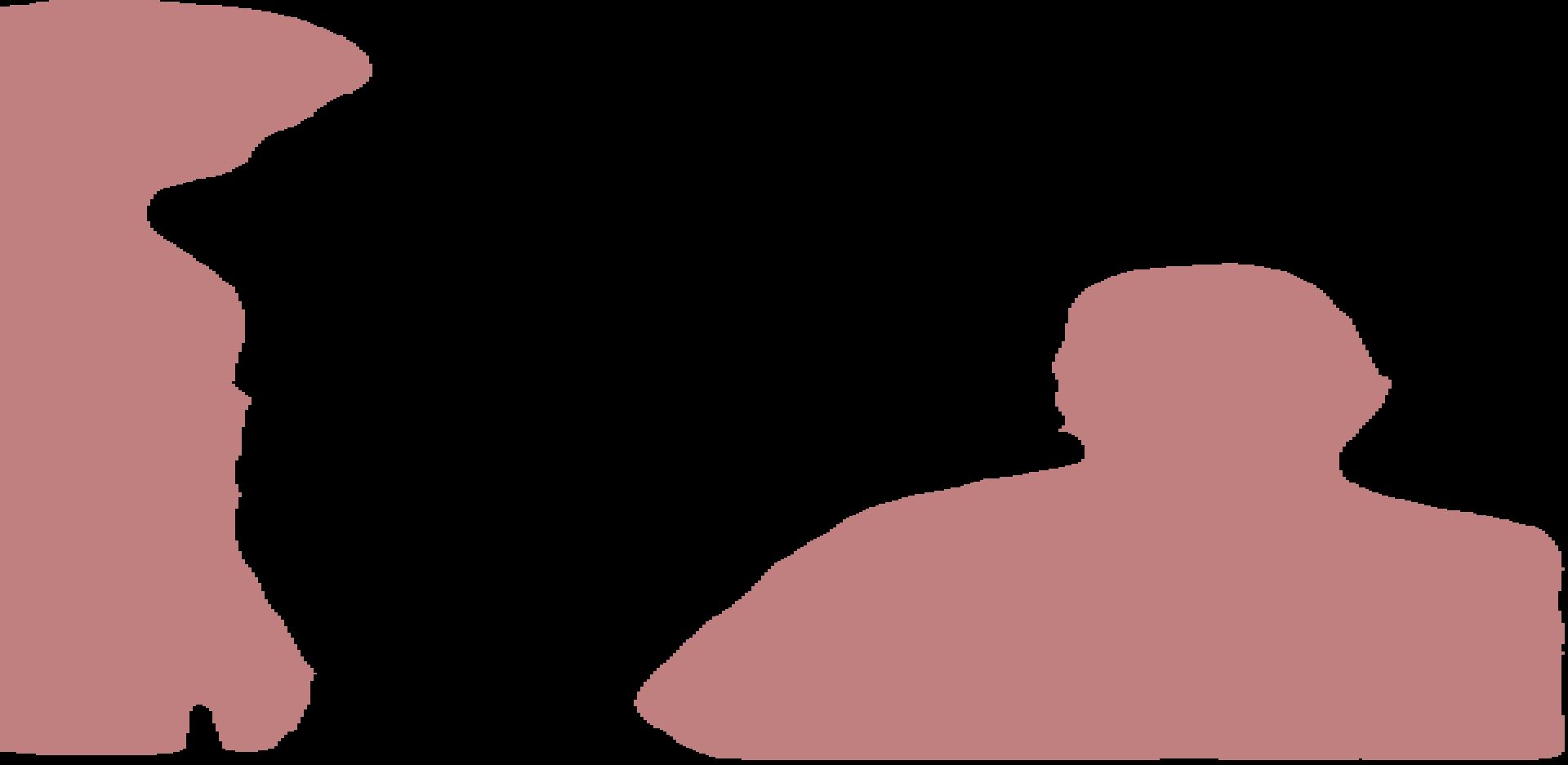}\\
        \includegraphics[width=0.993\textwidth, height=0.4in]{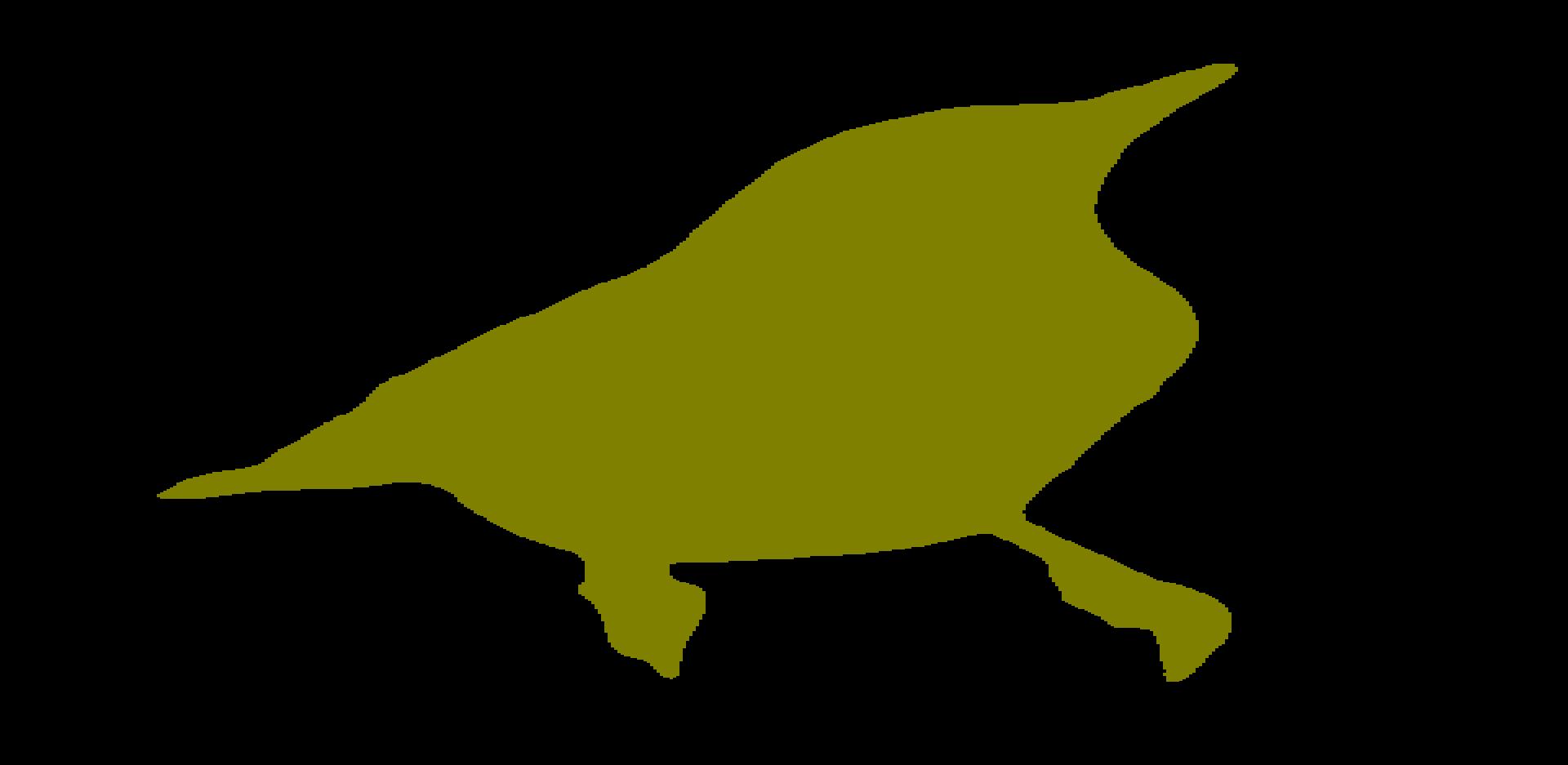}\\
        \includegraphics[width=0.993\textwidth, height=0.4in]{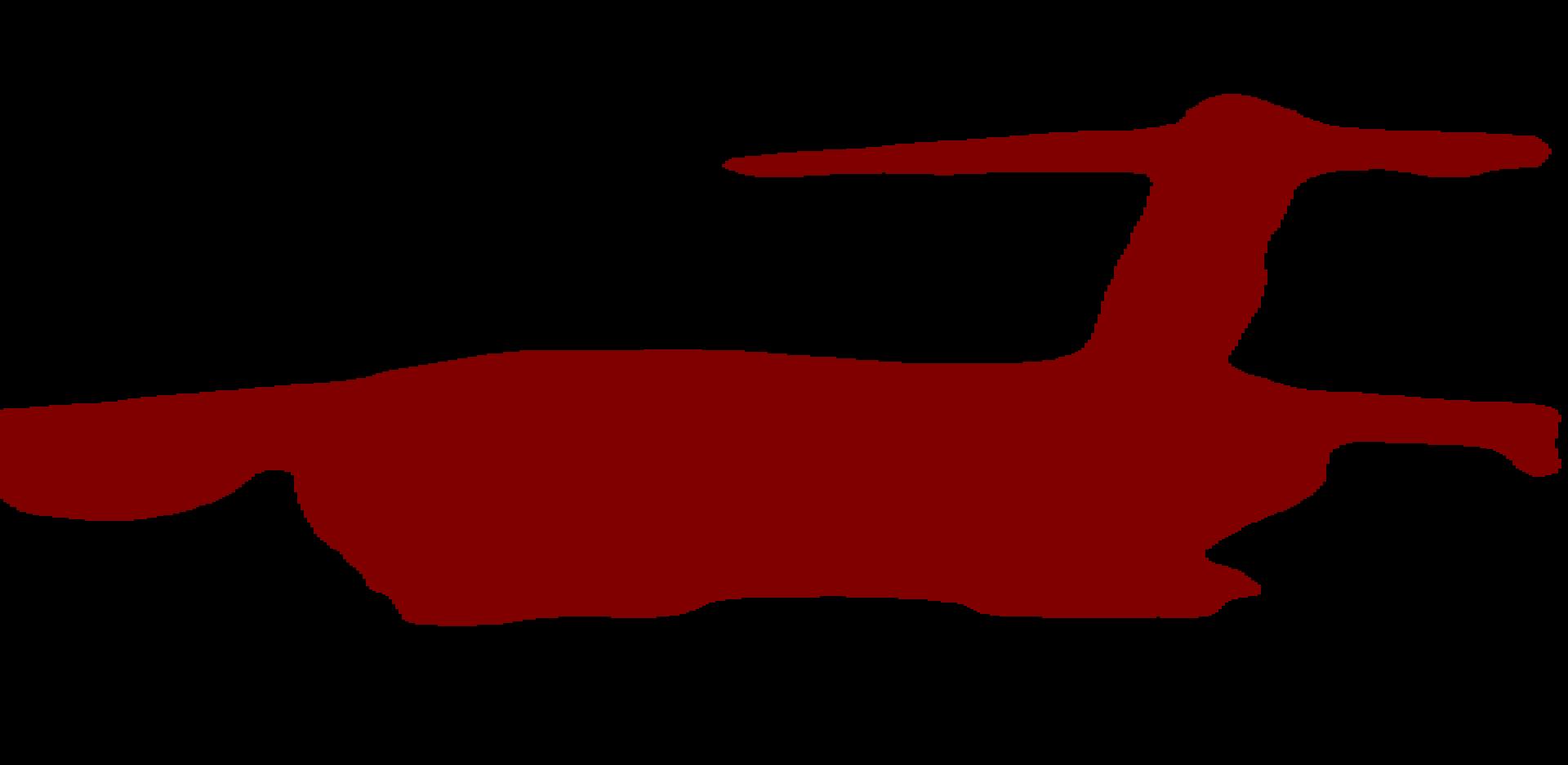}\\
        \footnotesize (d) \method

    \end{minipage}%
\centering
\caption{Qualitative results on the Pascal VOC2012 dataset.}
\label{fig:vis_voc}
\end{figure}

\begin{table}[]
    \centering
    \caption{Semantic Segmentation: PASCAL VOC 2012 validation set. All the methods are tested with multi-scale inputs. $*$ means adopting COCO-pretrained weights. 
    }
    \resizebox{0.4\textwidth}{!}{
    \begin{tabular}{llc}
    \toprule[1.2pt]
    Method & Backbone & mIoU\% \\
    \midrule
    DeepLabV3~\cite{chen2017rethinking} & D-ResNet-101    & 75.7 \\
    Dynamic~\cite{li2020learning} & Layer33   & 79.0 \\
    Res2Net~\cite{gao2019res2net} & Res2Net-101    & 80.2 \\
    DANet~\cite{fu2019dual} & ResNet-101    & 80.4 \\ 
    Auto-Deeplab~\cite{liu2019auto}$^*$ & ResNet-101   & 82.0 \\
    EncNet~\cite{zhang2018context} & D-ResNet-101    & 85.9 \\
    SANet~\cite{zhong2020squeeze}$^*$ & ResNet-101   & 86.1 \\ 
    \midrule
    \method{}  & D-ResNet-101   & {\bf 89.8} \\ 
    \bottomrule[1.2pt]
    \end{tabular}}
    \label{tab:overall_voc12}
\end{table}

\begin{figure*}[t]
\centering
    \begin{minipage}{0.19\linewidth}
        \centering
        \includegraphics[width=0.993\textwidth,height=0.7in]{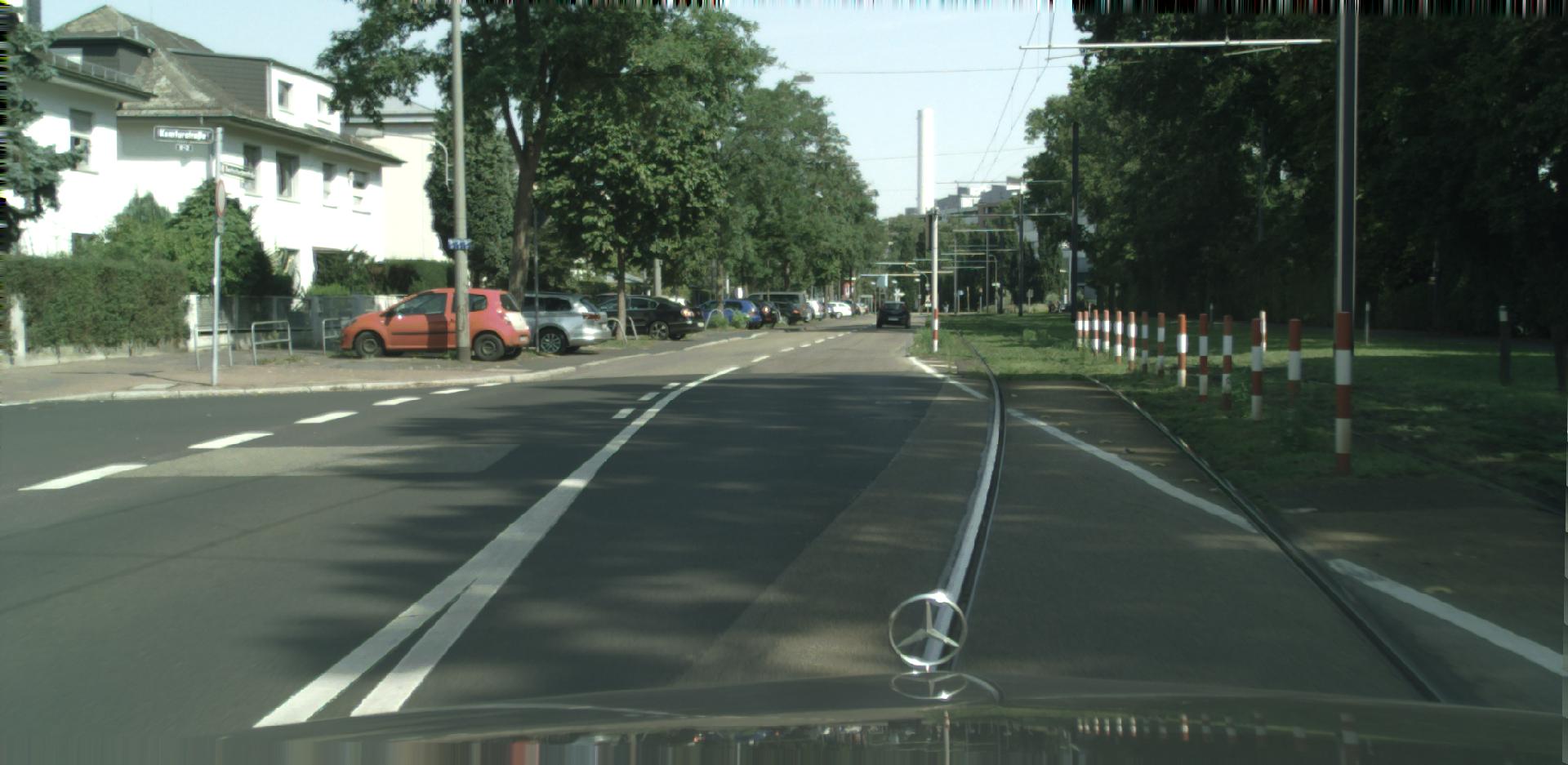}\\
        \includegraphics[width=0.993\textwidth, height=0.7in]{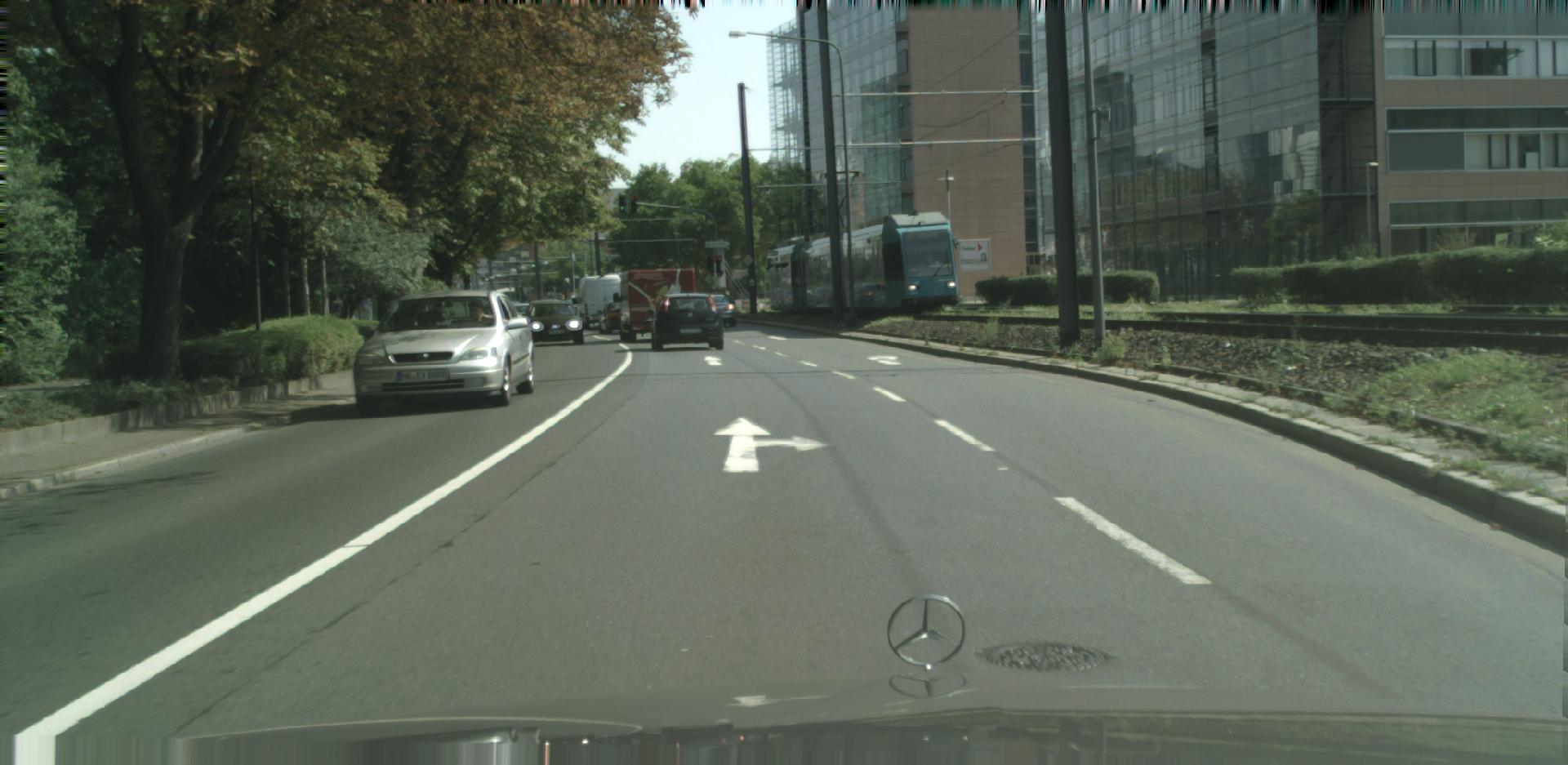}\\
        \includegraphics[width=0.993\textwidth, height=0.7in]{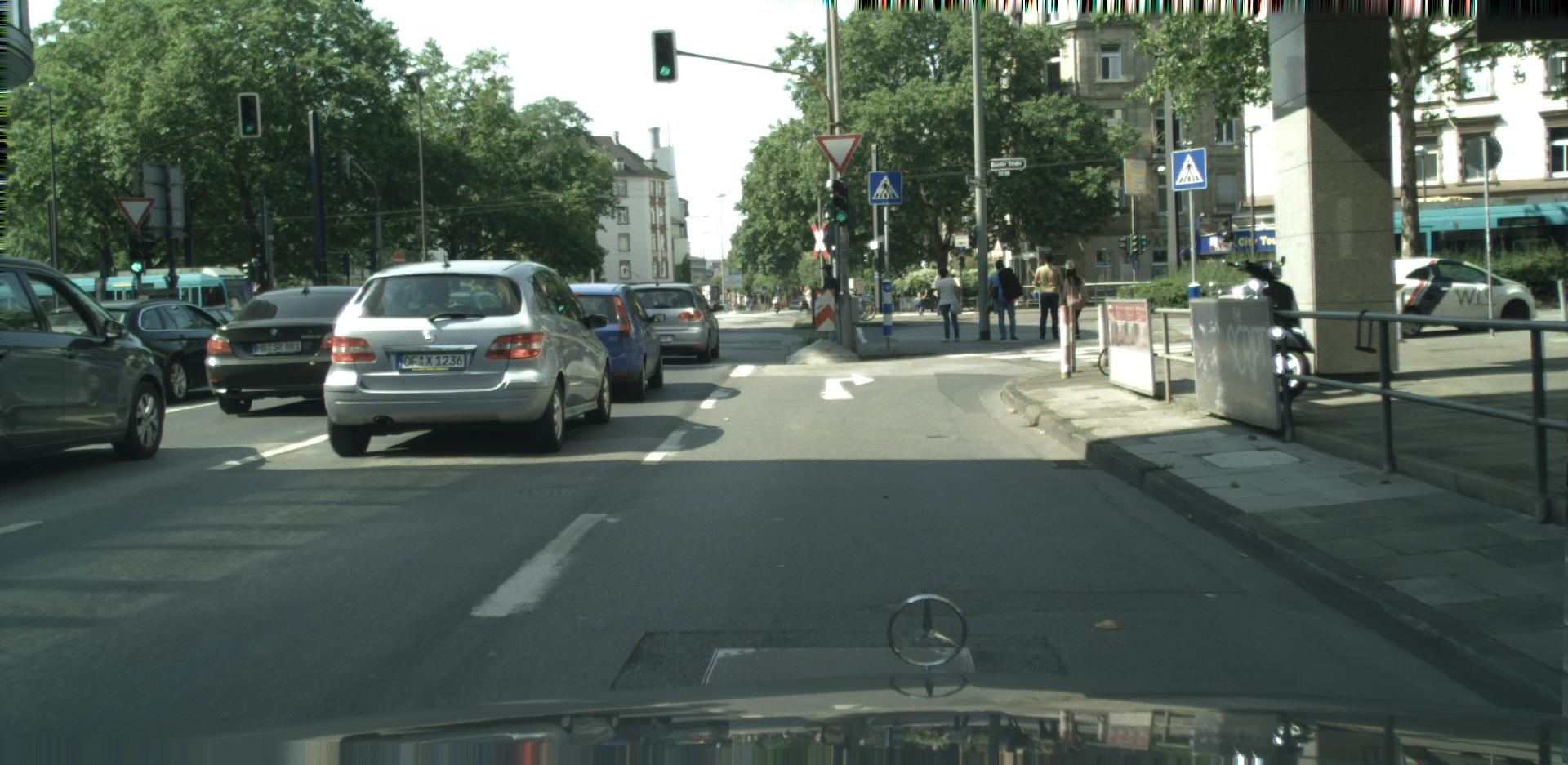}\\
        \includegraphics[width=0.993\textwidth, height=0.7in]{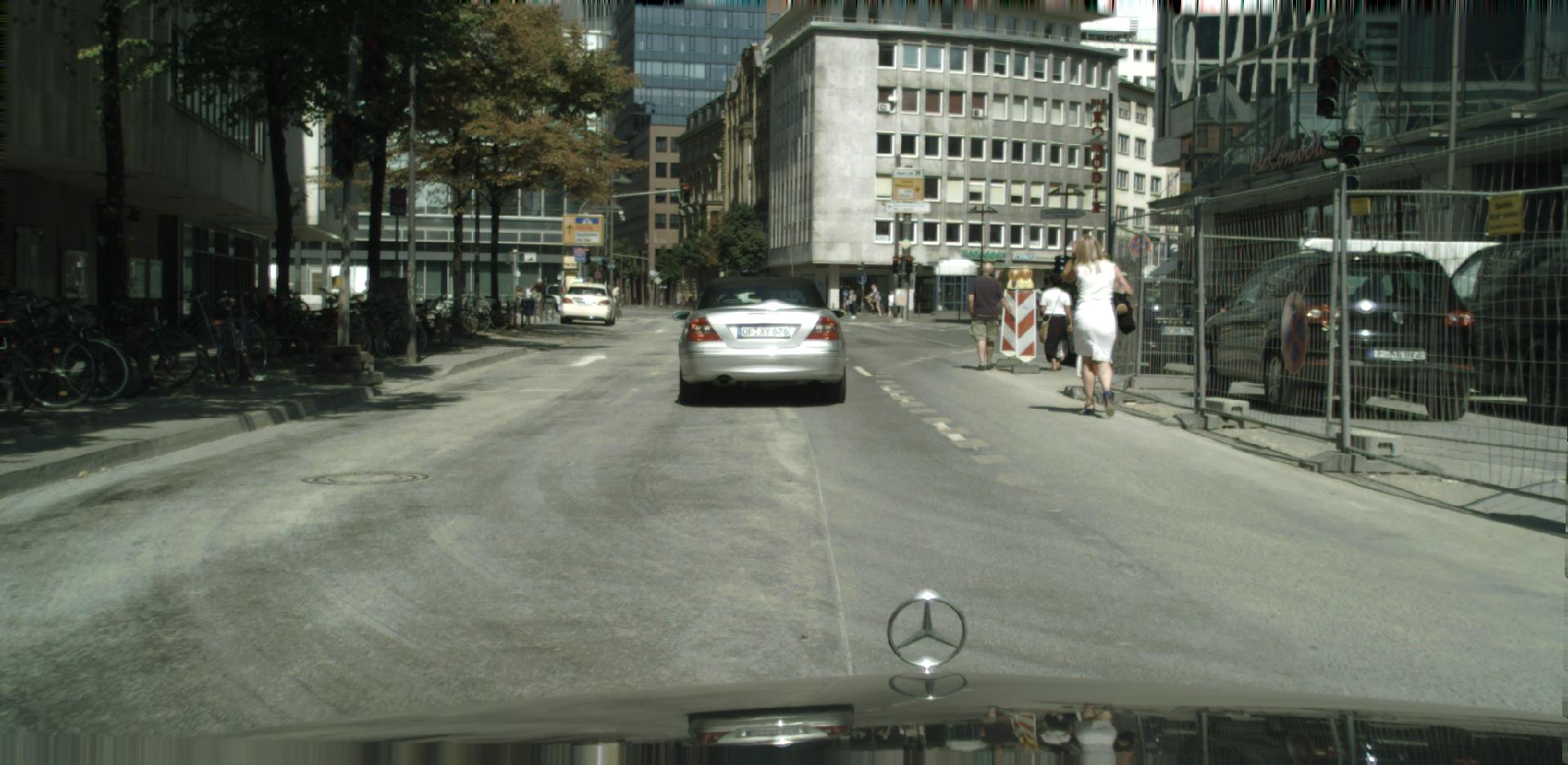}\\
        \footnotesize (a) Image
        
    \end{minipage}%
    \begin{minipage}{0.19\linewidth}
        \centering
        \includegraphics[width=0.993\textwidth,height=0.7in]{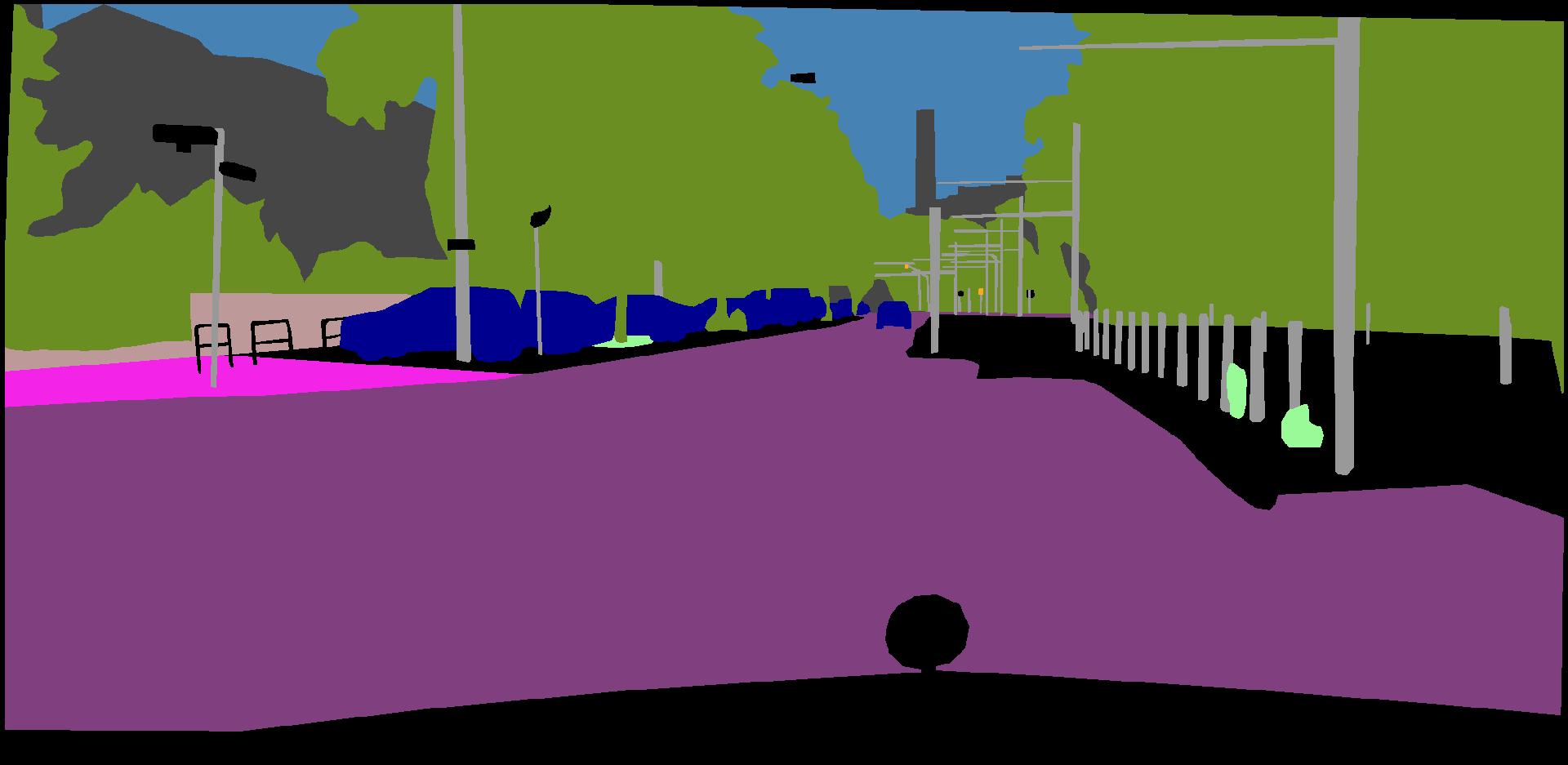}\\
        \includegraphics[width=0.993\textwidth, height=0.7in]{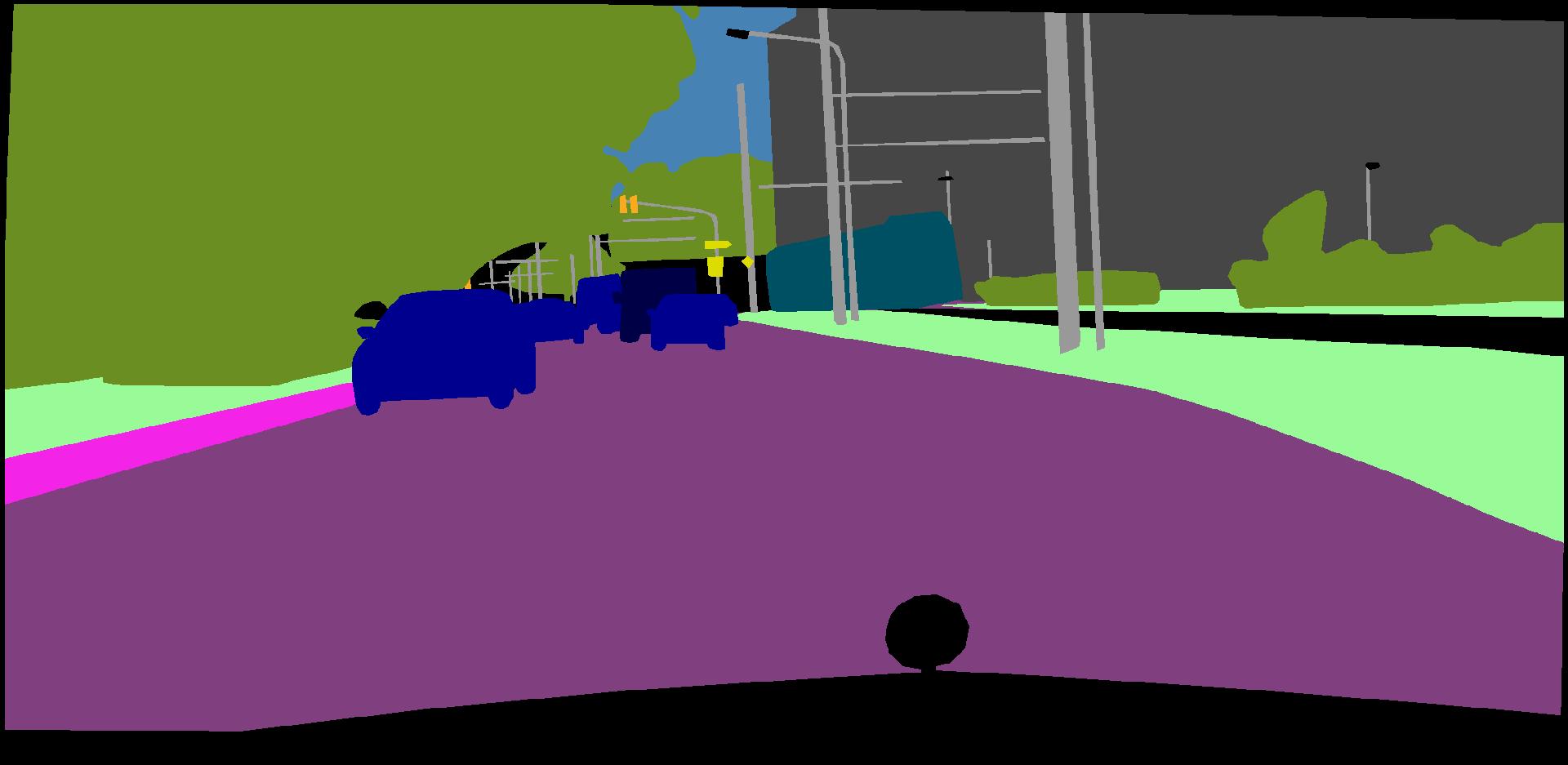}\\
        \includegraphics[width=0.993\textwidth, height=0.7in]{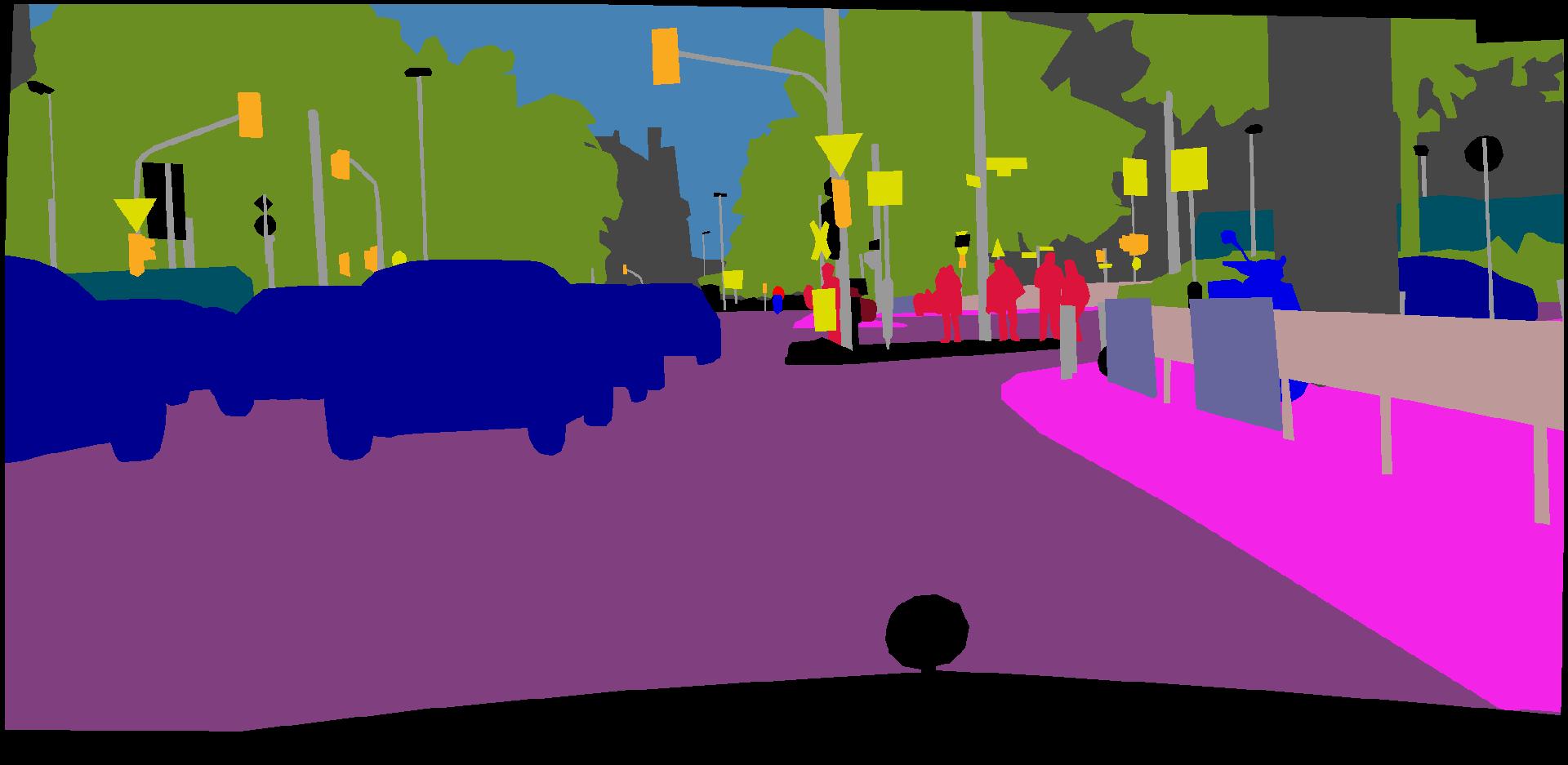}\\
        \includegraphics[width=0.993\textwidth, height=0.7in]{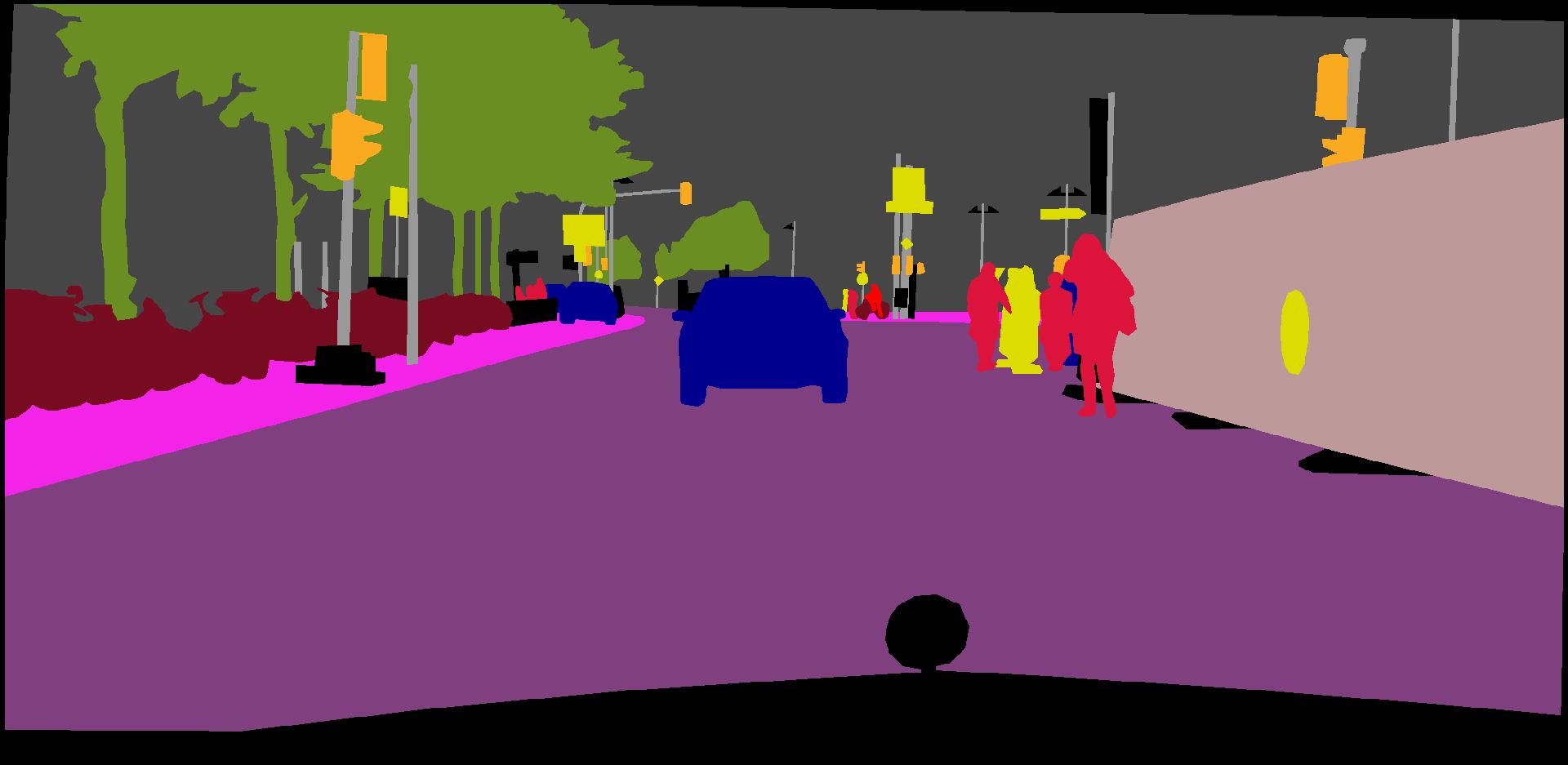}\\
        \footnotesize (b) GT

    \end{minipage}%
    \begin{minipage}{0.19\linewidth}
        \centering
        \includegraphics[width=0.993\textwidth,height=0.7in]{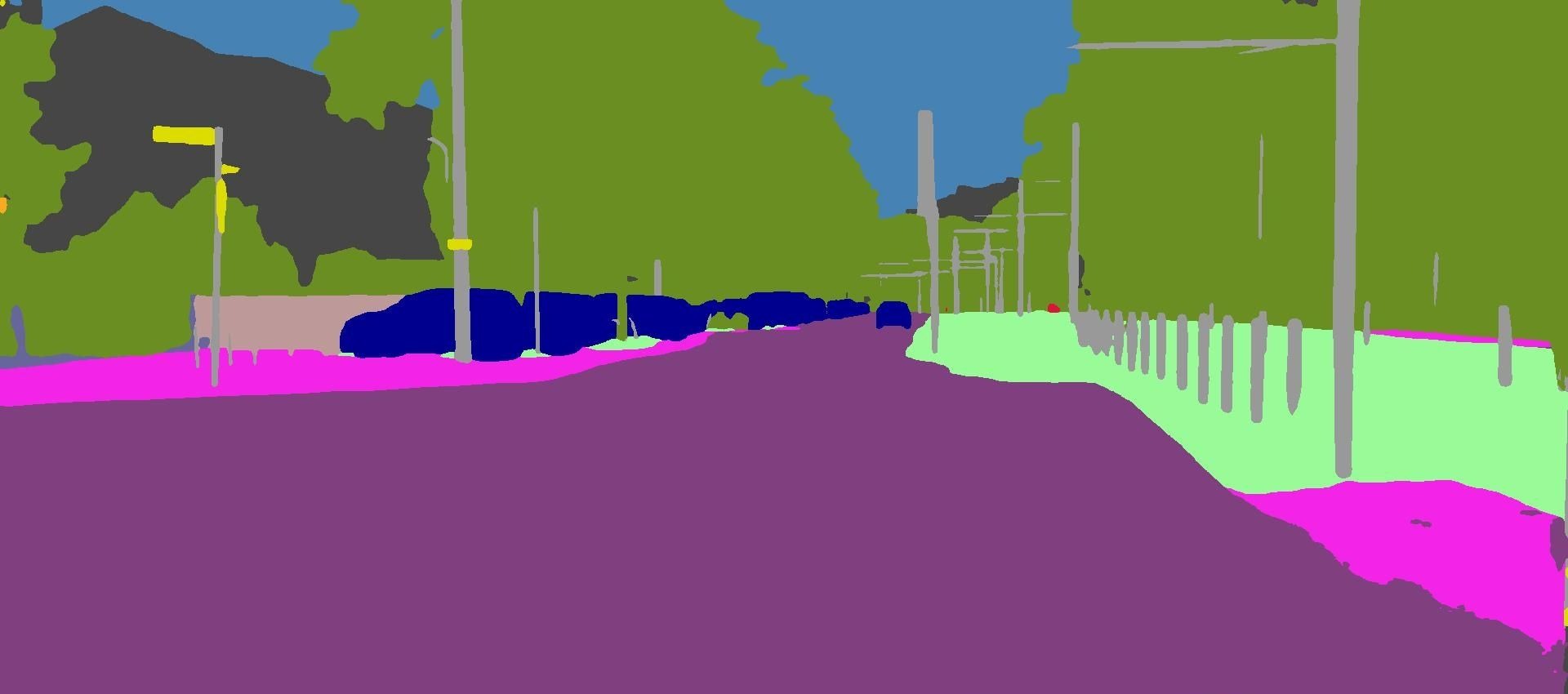}\\
        \includegraphics[width=0.993\textwidth, height=0.7in]{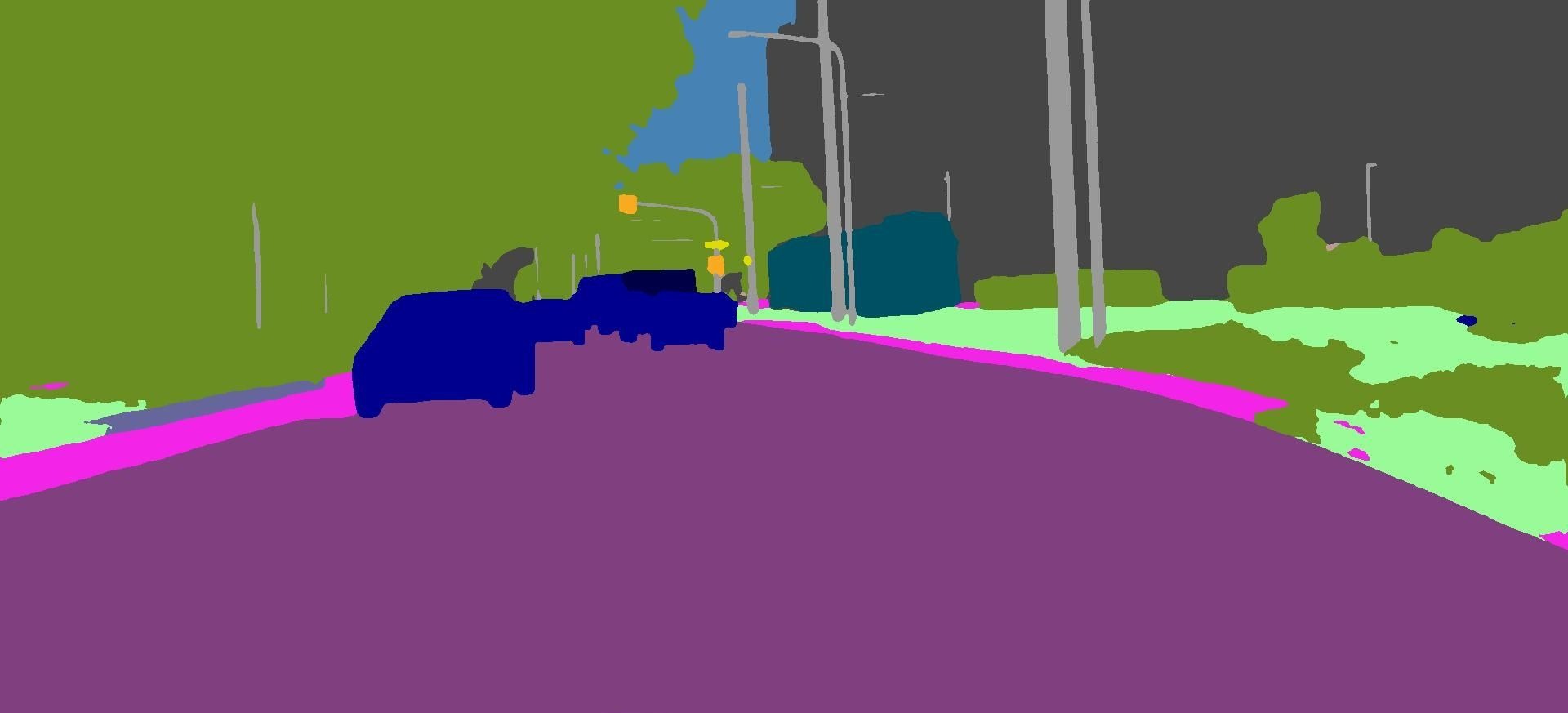}\\
        \includegraphics[width=0.993\textwidth, height=0.7in]{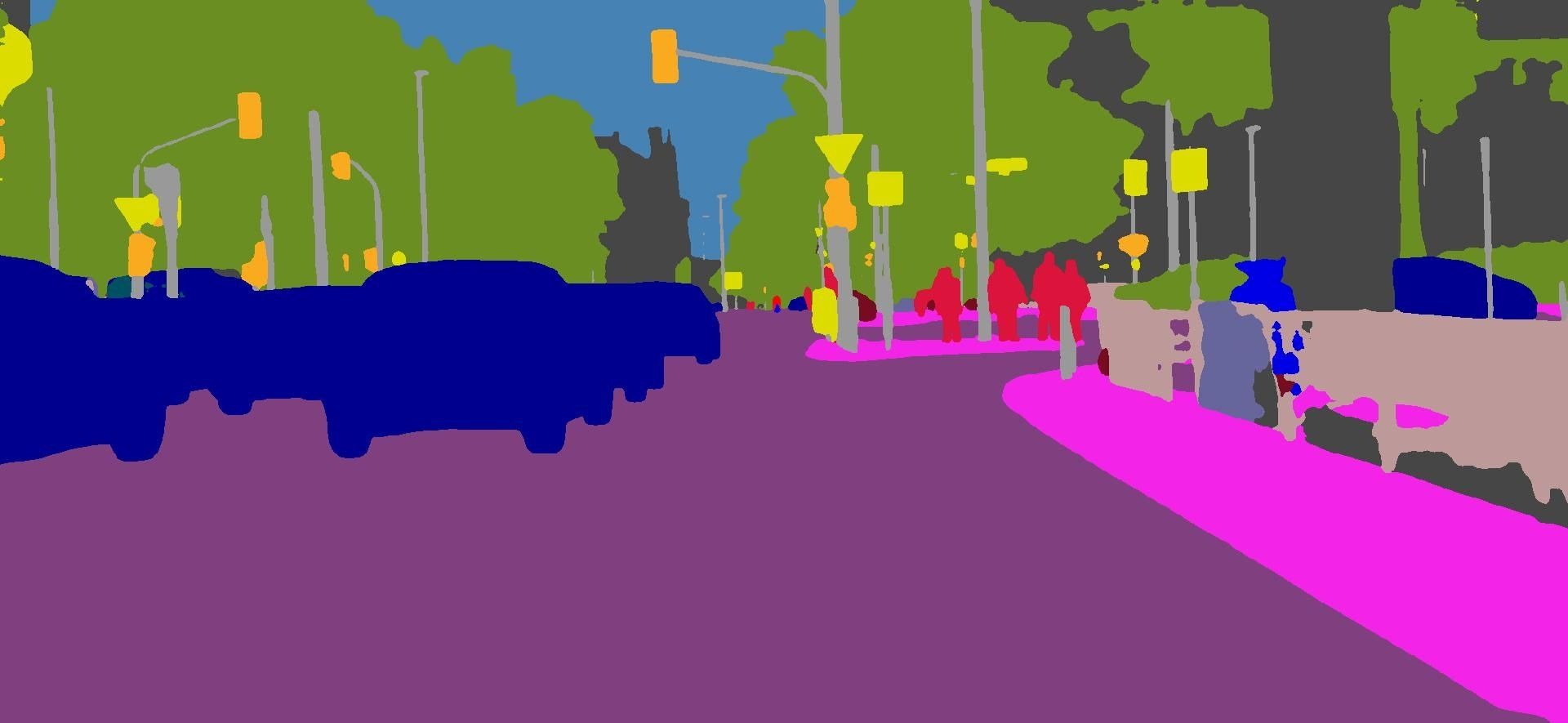}\\
        \includegraphics[width=0.993\textwidth, height=0.7in]{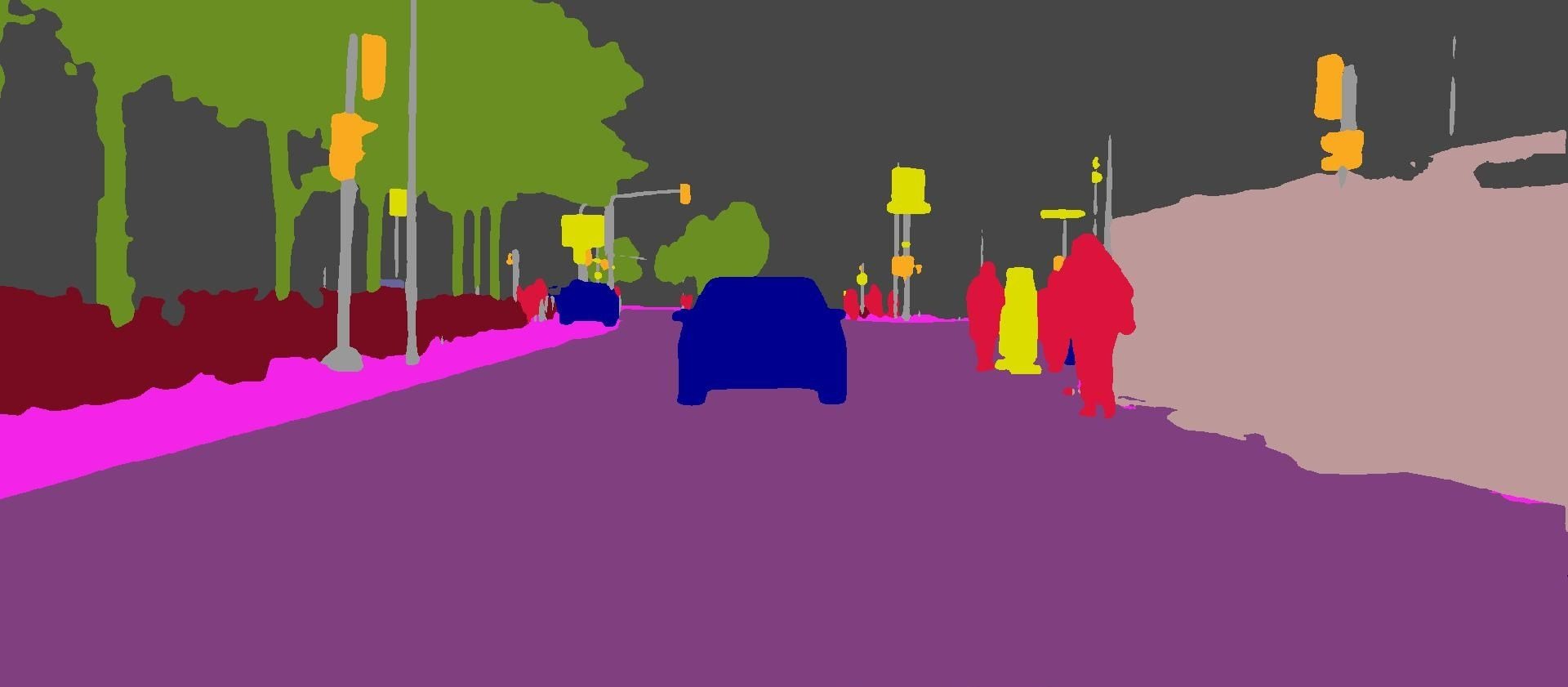}\\
        \footnotesize (c) DANet

    \end{minipage}%
    \begin{minipage}{0.19\linewidth}
        \centering
        \includegraphics[width=0.993\textwidth,height=0.7in]{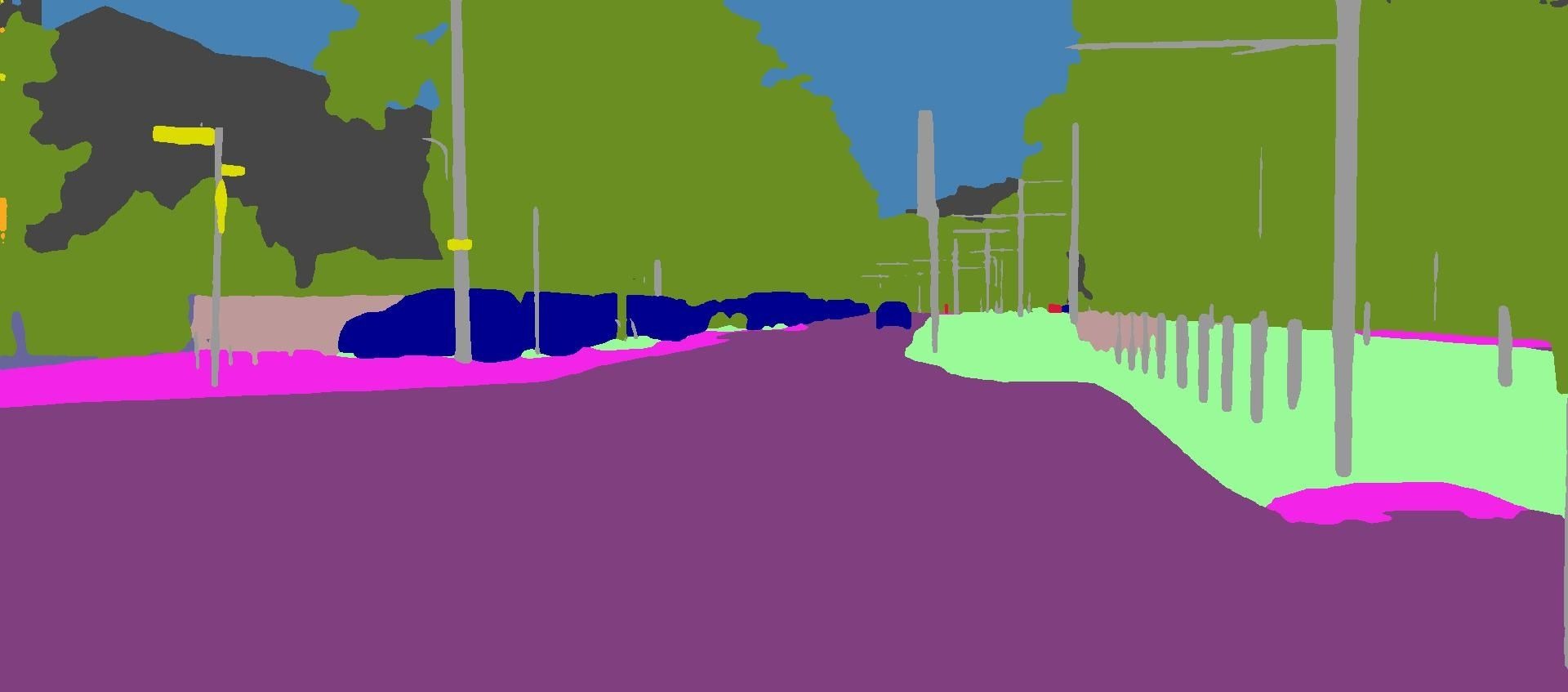}\\
        \includegraphics[width=0.993\textwidth, height=0.7in]{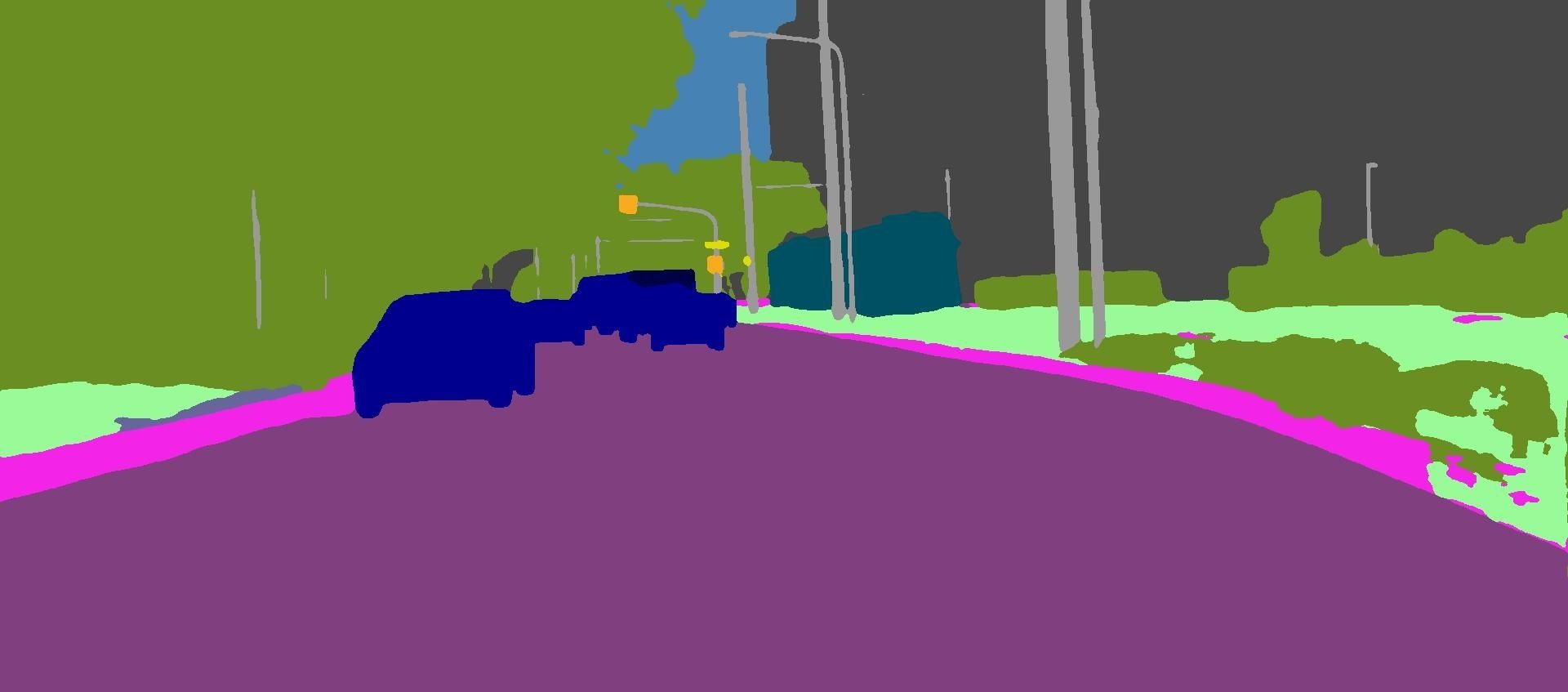}\\
        \includegraphics[width=0.993\textwidth, height=0.7in]{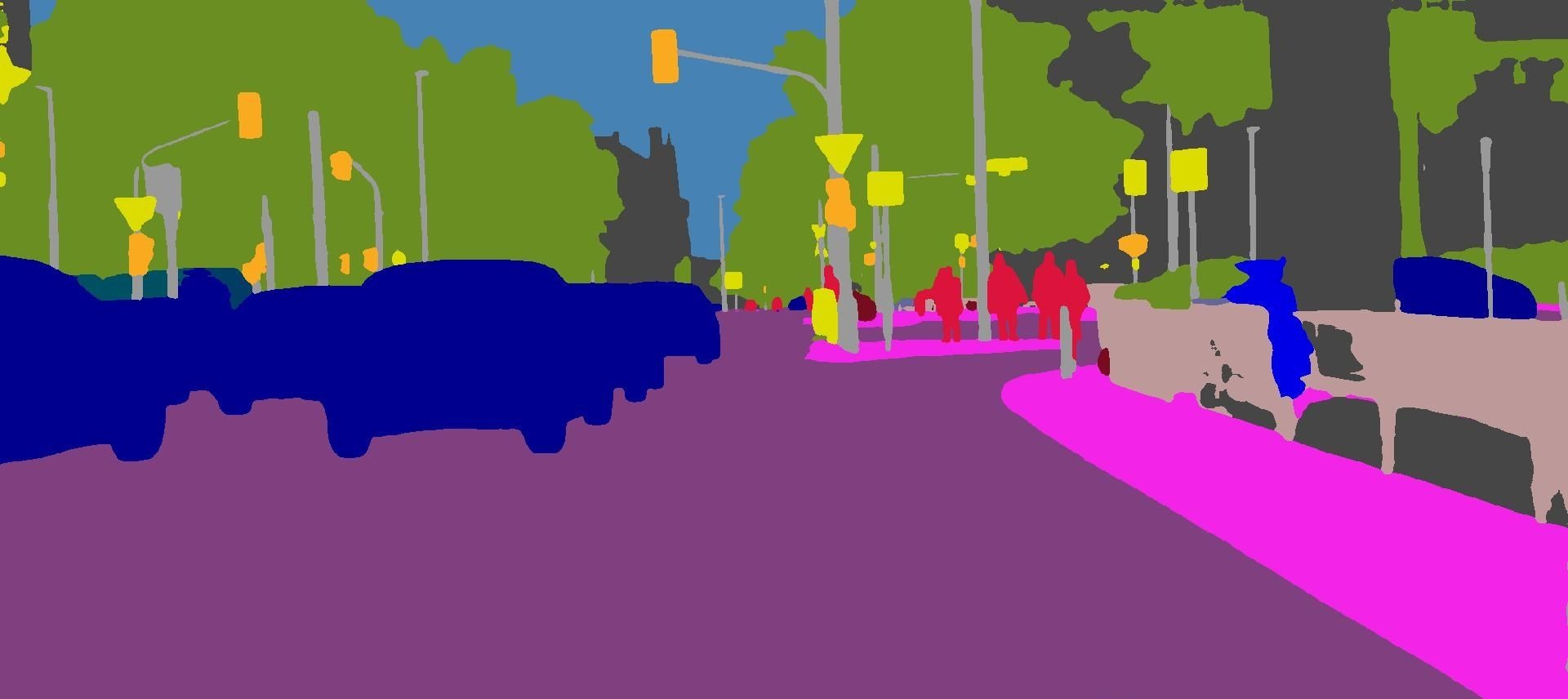}\\
        \includegraphics[width=0.993\textwidth, height=0.7in]{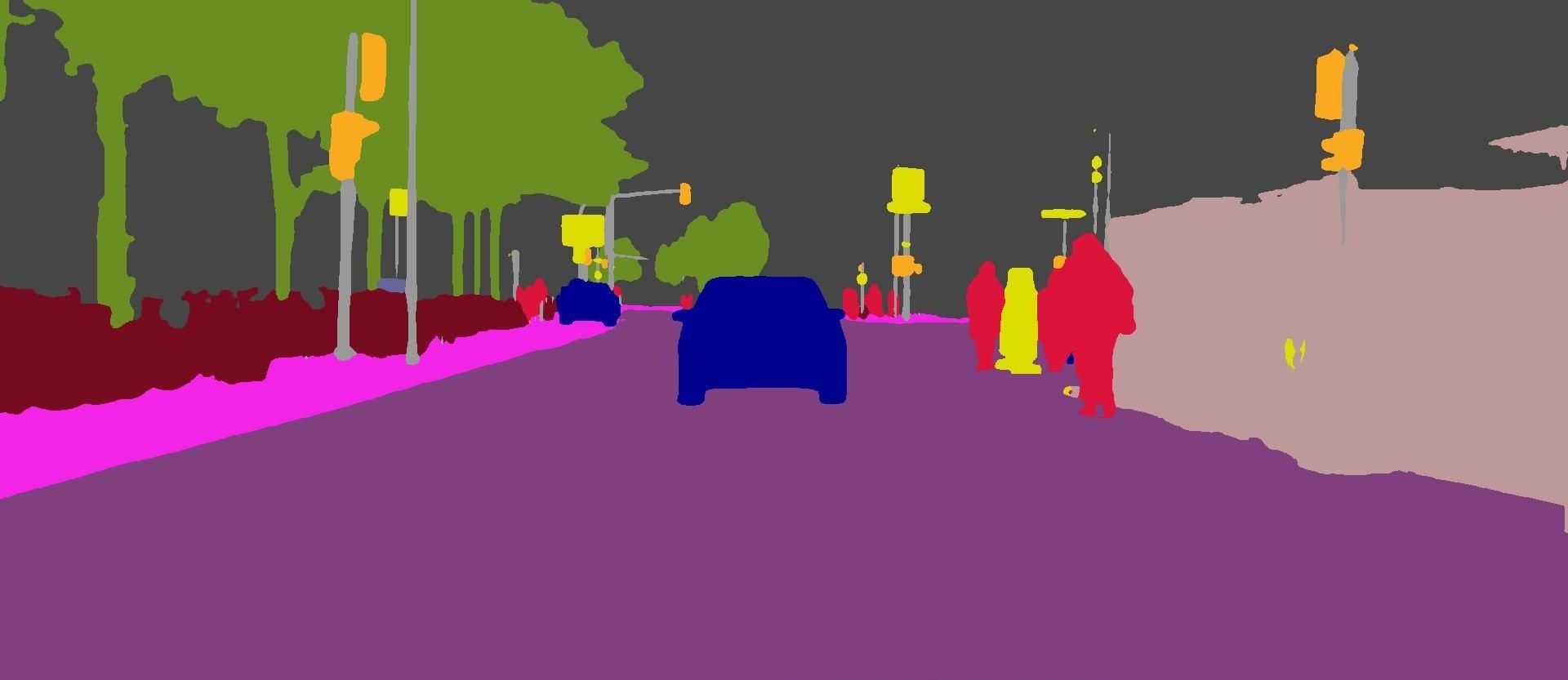}\\
        \footnotesize (d) HRNet

    \end{minipage}%
    \begin{minipage}{0.19\linewidth}
        \centering
        \includegraphics[width=0.993\textwidth,height=0.7in]{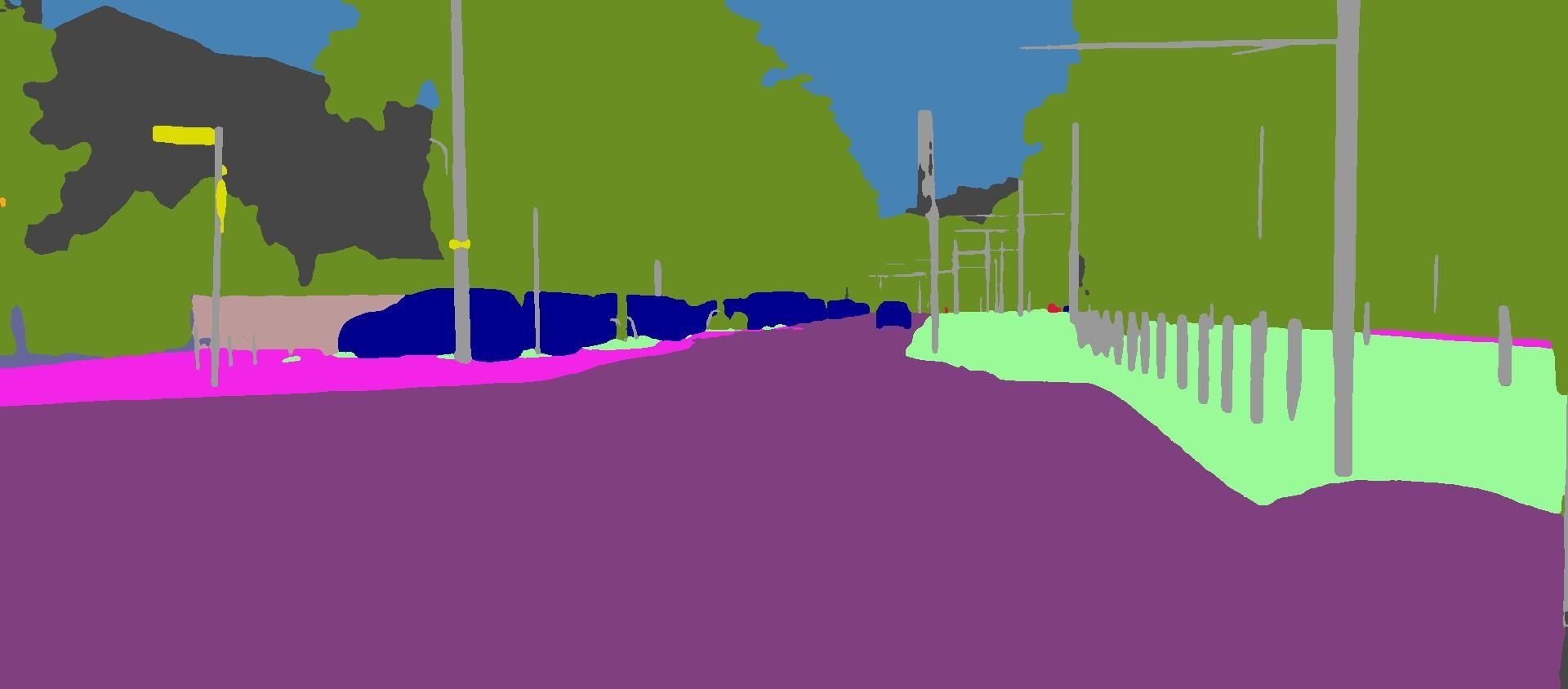}\\
        \includegraphics[width=0.993\textwidth, height=0.7in]{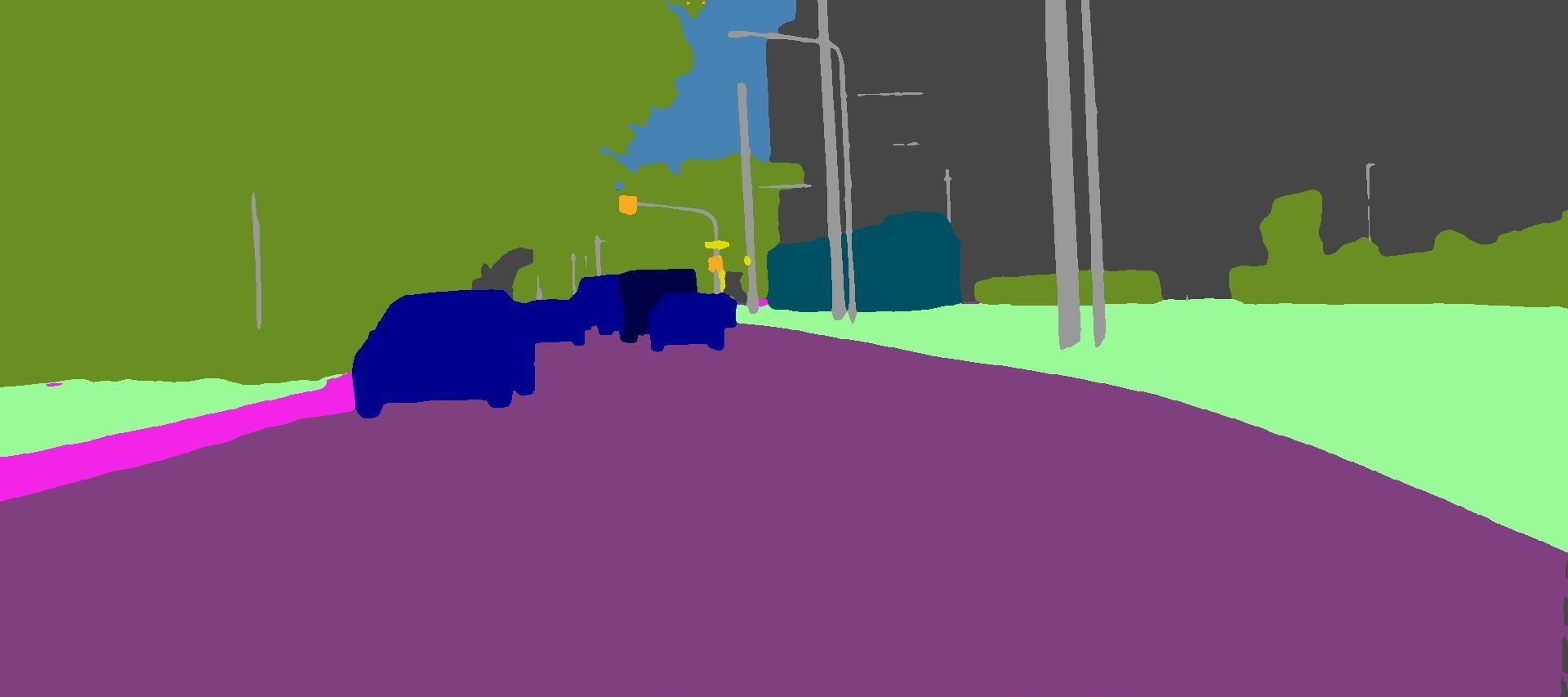}\\
        \includegraphics[width=0.993\textwidth, height=0.7in]{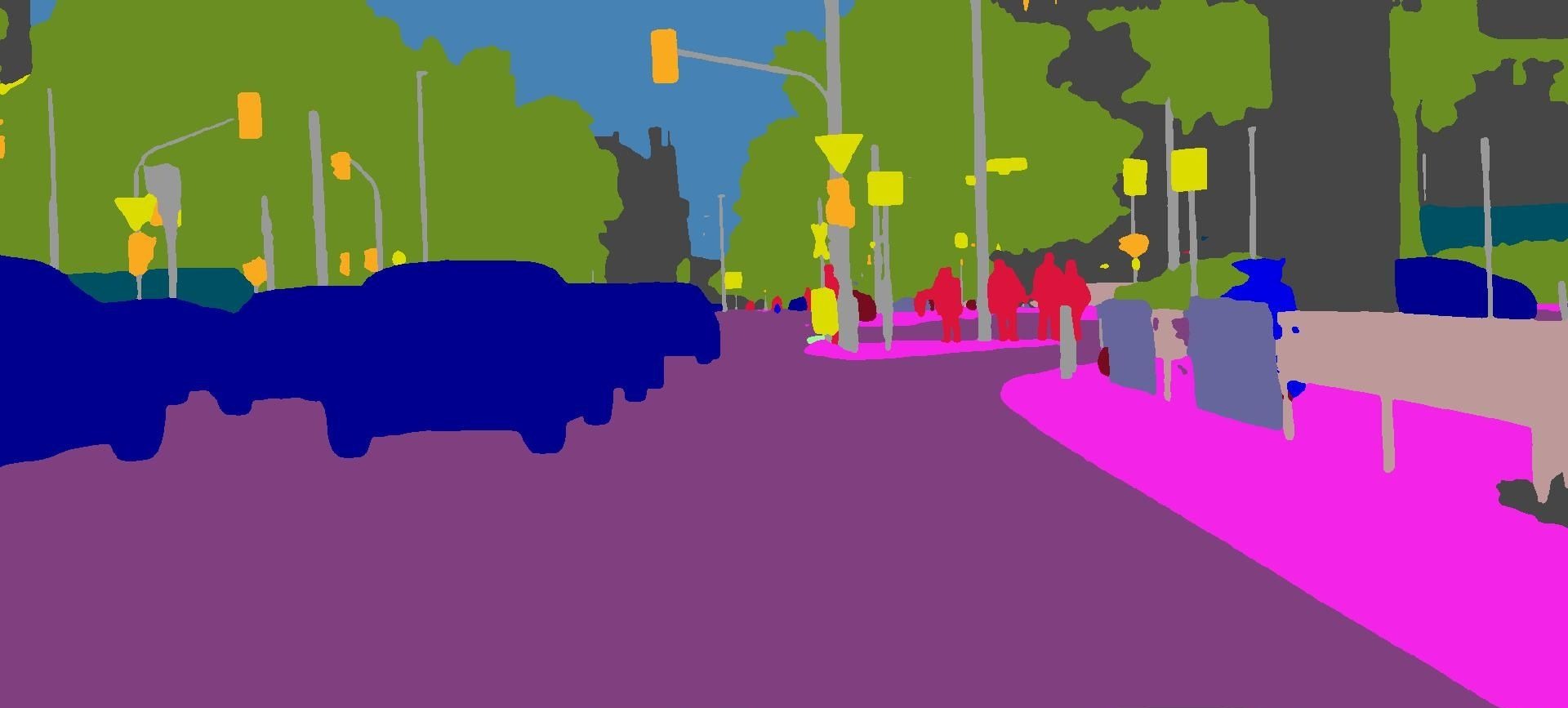}\\
        \includegraphics[width=0.993\textwidth, height=0.7in]{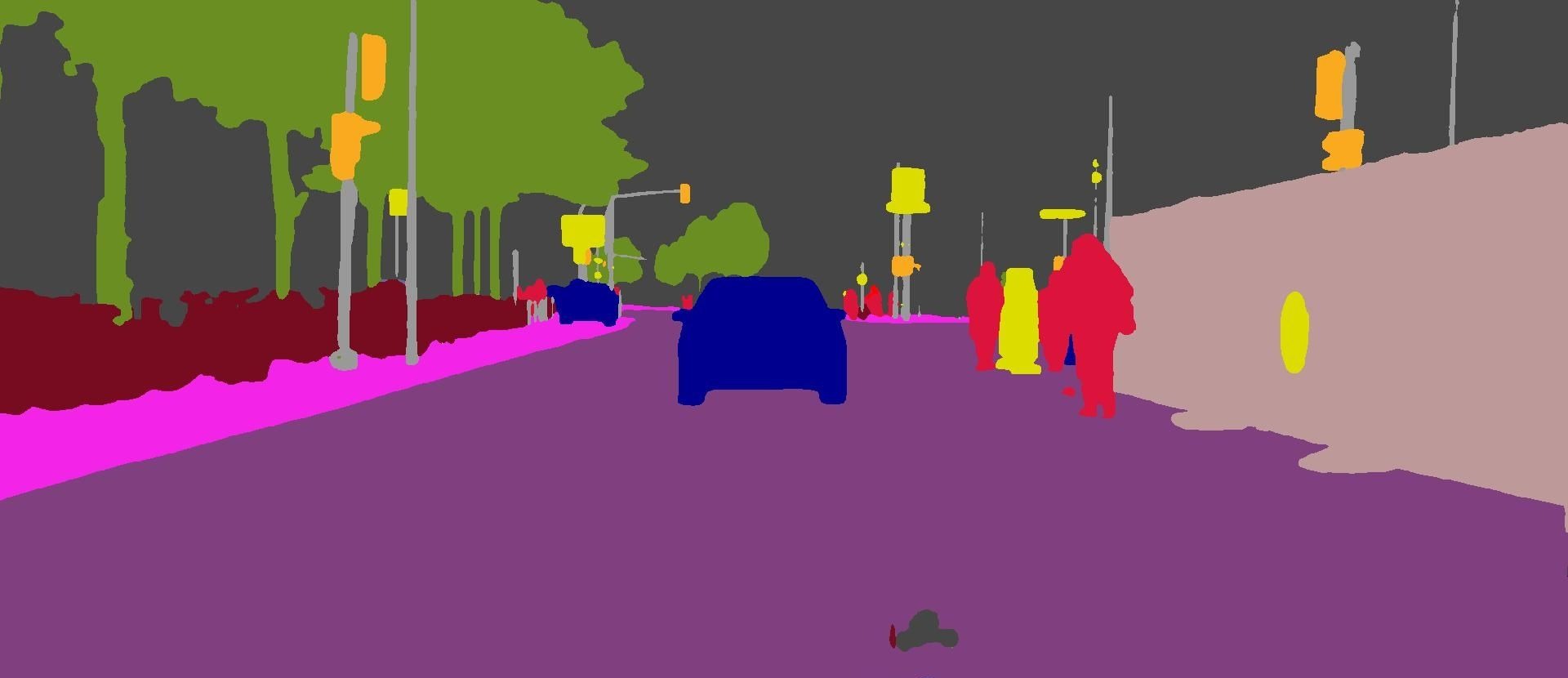}\\
        \footnotesize (e) \method

    \end{minipage}%
\centering
\caption{Qualitative semantic segmentation results on Cityscapes dataset.}
c
\label{fig:vis_citys}
\end{figure*}

\begin{table}[t]
\centering
\caption{Semantic Segmentation: Cityscapes validation and test set (trained on the standard training set).D-ResNet-101 is short for Dilated-ResNet-101.}
\label{tab:overall_citys}

    \resizebox{0.45\textwidth}{!}{
    \begin{tabular}{lcccc}
    \toprule[1.2pt]
    Method & Backbone  & Test set & mIoU \\
    \midrule
    DeepLabv3~\cite{chen2017rethinking}& D-ResNet-101 & Val & 78.5 \\  
    PSPNet~\cite{zhao2016pyramid} &D-ResNet-101 & Val & 79.7 \\  
    Dynamic~\cite{li2020learning} & Layer33-PSP  & Val & 79.7 \\  
    SpyGR~\cite{li2020spatial} & ResNet-101  & Val & 80.5 \\
    HRNet~\cite{wang2020deep} & HRNetV2-W48  & Val & 81.1 \\
    CCNet~\cite{huang2019ccnet} & D-ResNet-101  & Val & 81.3 \\
    DANet~\cite{fu2019dual} & D-ResNet-101  & Val & 81.5 \\
    Panoptic-DeepLab~\cite{cheng2020panoptic} & D-ResNet-101 & Val & 81.5 \\
    {OCR~\cite{YuanCW19}} & HRNetV2-W48 & Val & 81.6\\
    {CDGCNet~\cite{hu2020class}} & D-ResNet-101 & Val & 81.9\\ 
   {Wang~\etal~\cite{wang2021exploring}} & HRNetV2-W48 & Val & 82.2\\
    \midrule
    {\method{} } & D-ResNet-101  & Val & {82.0} \\
    \method{}  & HRNetV2-W48  & Val & \textbf{82.3} \\
    \midrule
    \midrule
    PSANet~\cite{zhao2018psanet} & D-ResNet-101  & Test & 78.6 \\
    PAN~\cite{li2018pyramid} & D-ResNet-101  & Test & 78.6 \\
    AAF~\cite{ke2018adaptive} & D-ResNet-101  & Test & 79.1 \\
    HRNet~\cite{wang2020deep} & HRNetV2-W48  & Test & 80.4 \\
    Dynamic~\cite{li2020learning} & Layer33-PSP  & Test & 80.7 \\
    {Wang~\etal~\cite{wang2021exploring} }& HRNetV2-W48 & Test & \textbf{81.4}\\
    \midrule
    {\method{} } & D-ResNet-101 & Test & {81.2}\\
    \method{}  & HRNetV2-W48 & Test & \textbf{81.4}\\
    \bottomrule[1.2pt]
    \end{tabular}}
\end{table}

\begin{table}[]
\centering
\caption{Semantic segmentation: Cityscapes test set (learned on the train+val set, multi-scale and flipping). D-ResNet-101 is short for Dilated-ResNet-101. \vspace{2mm}}
\label{tab:overview_city}
\resizebox{0.45\textwidth}{!}{%
\begin{tabular}{llcccc}
\toprule
Method & Backbone & mIoU & iIoU cla & IoU cat. & iIoU cat. \\
\midrule
DeepLab~\cite{chen2016deeplab} & D-ResNet-101 & 70.4 & 42.6 & 86.4 & 67.7 \\
PADNet~\cite{xu2018pad} & D-ResNet-101 & 80.3 & 58.8 & 90.8 & 78.5 \\
Dynamic~\cite{li2020learning}& Layer33-PSP & 80.7 & - & - & - \\
SVCNet~\cite{ding2019semantic} & ResNet-101 & 81.0 & - & - & - \\
ANN~\cite{zhu2019asymmetric} & D-ResNet-101 & 81.3 & - & - & - \\
CCNet~\cite{huang2019ccnet} & D-ResNet-101 & 81.4 & - & - & - \\
DANet~\cite{fu2019dual} & D-ResNet-101 & 81.5 & - & - & - \\
DGMN~\cite{zhang2020dynamic} & D-ResNet-101 & 81.6 & - & - & - \\
SpyGR~\cite{li2020spatial} & ResNet-101 & 81.6 & - & - & - \\
HRNet~\cite{wang2020deep} & HRNetV2-W48 & 81.6 & 61.8 & \textbf{92.1} & \textbf{82.2} \\
ACFNet~\cite{zhang2019acfnet} & ResNet-101 & 81.8 & - & - & - \\
DGCNet~\cite{zhang2019dual} & ResNet-101 & 82.0 & - & - & - \\
HANet~\cite{choi2020cars} & ResNext-101 & 82.1 & - & - & - \\
\midrule
{VISTA-Net} & D-ResNet-101 & {81.7} & {62.0} & 91.6 & 81.5 \\
VISTA-Net & HRNetV2-W48 & \textbf{82.2} & \textbf{62.7} & 91.9 & 82.1 \\
\bottomrule
\end{tabular}%
}
\label{tab:overall_citys_test}
\end{table}

\begin{table}[!t]
\caption{Surface normal prediction: ScanNet dataset.}
\centering
\label{tab:overall_sn}
\resizebox{0.45\textwidth}{!}{%
\begin{tabular}{lccccc}
\toprule
\multirow{2}{*}{Methods} & \multicolumn{2}{c}{Error metric} & \multicolumn{3}{c}{Accuracy metric} \\
 & mean & median & 11.25 & 22.5 & 30 \\ 
\midrule
Skip-Net\cite{bansal2016marr} & 26.2 & 20.6 & 28.8 & 54.3 & 67.0\\
Zhang\etal~\cite{zhang2017physically} & 23.3 & 16.0 & 40.4 & 63.1 & 71.9\\
GeoNet~\cite{qi2018geonet} & 19.8 & 11.3 & 49.7 & 70.4 & 77.7 \\
FrameNet~\cite{huang2019framenet} & 15.3 & 8.1 & 60.6 & 78.6 & 84.7 \\
\midrule
VISTA-Net ) &  \textbf{15.1} & \textbf{7.5} & \textbf{63.8} & \textbf{80.0} & \textbf{85.2} \\ 
\bottomrule
\end{tabular}}
\end{table}

\noindent \textbf{Semantic Segmentation.} 
We first compare VISTA-Net with the most recent methods on the Pascal-Context dataset, including \cite{zhang2018context,fu2019dual,zhu2019asymmetric,ding2019semantic,zhang2019co,wang2020deep,he2019adaptive,zhong2020squeeze,li2020spatial,xu2020probabilistic}.
VISTA-Net, as shown in Table~\ref{tab:overall-pcontext}, is 0.3 points better according to the mIoU metric than the best available method, \ie~PGA-net. Importantly, VISTA-Net 
outperforms EncNet~\cite{zhang2018context}, which uses only channel-wise attention, as well as DANet~\cite{fu2019dual}, which considers separate spatial and channel attention models. Meanwhile, the visualisation result is shown in Fig.~\ref{fig:vis_pcontext}.

We also compare our method with state of the art methods on PASCAL VOC2012, including~\cite{chen2017rethinking,li2020learning,gao2019res2net,fu2019dual,liu2019auto,zhang2018context,zhong2020squeeze,huynh2020guiding}.
Unsurprisingly, our method not only outperforms single channel attention methods like EncNet~\cite{zhang2018context} and single spatial attention methods like SANet~\cite{zhong2020squeeze} but also DANet~\cite{fu2019dual} which considers separate spatial and channel attention models.
VISTA-Net, as shown in Table~\ref{tab:overall_voc12}, is 3.7 points better according to the mIoU metric than the best available method, \ie~SANet. 
This clearly confirms the advantage of our probabilistic formulation and the importance of handling the spatial and channel wise attention in a structured manner within an unified probabilistic framework.

Meanwhile, the visualisation result is shown in Fig.~\ref{fig:vis_voc}. 
In Table~\ref{tab:overall_citys} we report the results of our method on Cityscape val/test dataset. For fair comparison, all methods do not use multi-scale and flipping.
{According to Table~\ref{tab:overall_citys}, \method~outperforms the competitors of 0.4\% and  0.7\% mIoU in validation and testing set. According to the visualization results (Fig.~\ref{fig:vis_citys}), our method captures more information about details with respect to DANet(dual attention) and HRNet (complex multi-scale).} For example in the fourth row, our method successfully predicts the yellow warning symbol while DANet and HRNet both miss it. 
Moreover, we  also compare our method with state of the art attention based methods learned  on  the  train+val set with multi-scale and flipping in Table~\ref{tab:overall_citys_test}, including~\cite{xu2018pad,li2020learning,ding2019semantic,zhu2019asymmetric,huang2019ccnet,fu2019dual,zhang2020dynamic,li2020spatial,wang2020deep,zhang2019acfnet,zhang2019dual,choi2020cars}. We outperform not only DANet and HRNet but also the state-of-the-art approach HANet.

\begin{figure}[h]
\centering
    \begin{minipage}{0.15\linewidth}
        \centering
        \includegraphics[width=0.993\textwidth,height=0.4in]{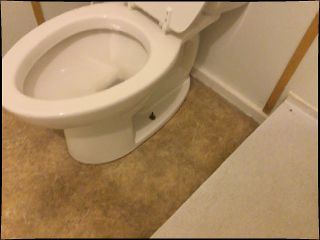}\\
        \includegraphics[width=0.993\textwidth,height=0.4in]{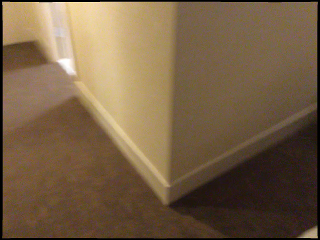}\\ 
        \includegraphics[width=0.993\textwidth,height=0.4in]{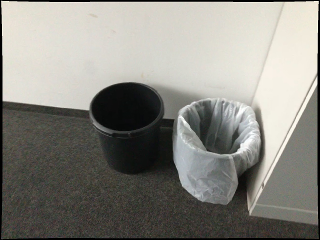}\\   
        \includegraphics[width=0.993\textwidth,height=0.4in]{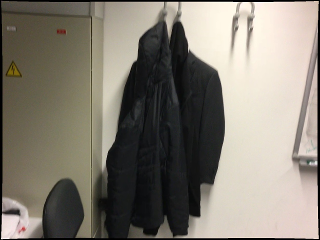}\\ 
        \footnotesize (a) Image

    \end{minipage}%
    \begin{minipage}{0.15\linewidth}
        \centering
        \includegraphics[width=0.993\textwidth,height=0.4in]{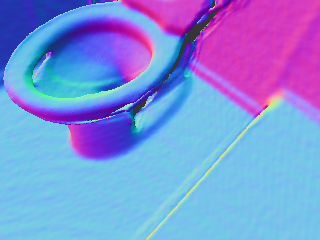}\\
        \includegraphics[width=0.993\textwidth,height=0.4in]{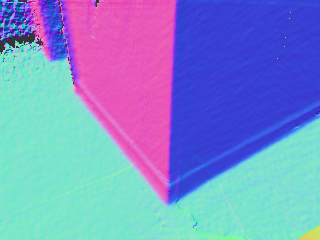}\\ 
        \includegraphics[width=0.993\textwidth,height=0.4in]{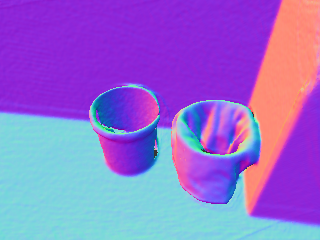}\\   
        \includegraphics[width=0.993\textwidth,height=0.4in]{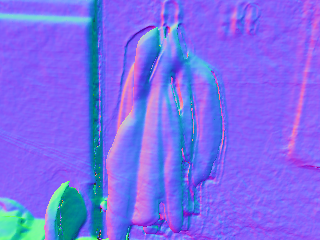}\\ 
       \footnotesize (b) GT

    \end{minipage}%
    \begin{minipage}{0.15\linewidth}
        \centering
        \includegraphics[width=0.993\textwidth,height=0.4in]{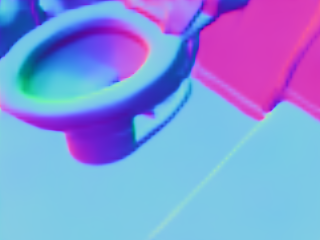}\\
        \includegraphics[width=0.993\textwidth,height=0.4in]{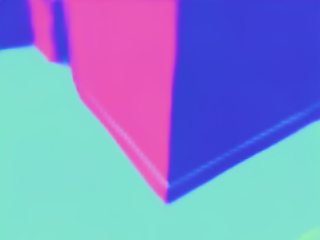}\\ 
        \includegraphics[width=0.993\textwidth,height=0.4in]{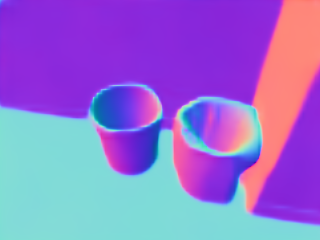}\\   
        \includegraphics[width=0.993\textwidth,height=0.4in]{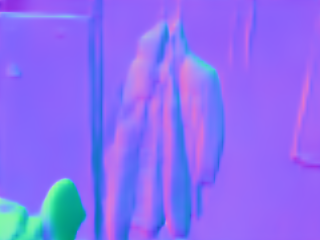}\\ 
        \footnotesize (c) Pred.

    \end{minipage}%
    \begin{minipage}{0.15\linewidth}
        \centering
        \includegraphics[width=0.993\textwidth,height=0.4in]{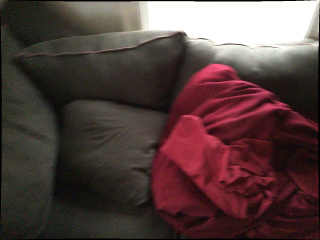}\\
        \includegraphics[width=0.993\textwidth,height=0.4in]{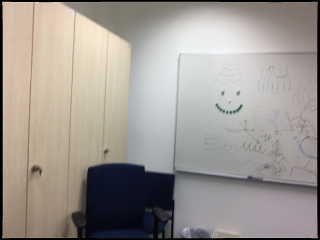}\\ 
        \includegraphics[width=0.993\textwidth,height=0.4in]{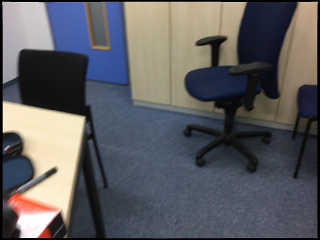}\\   
        \includegraphics[width=0.993\textwidth,height=0.4in]{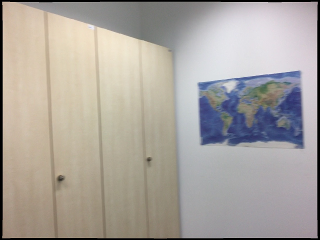}\\ 
        \footnotesize (d) Image

    \end{minipage}%
    \begin{minipage}{0.15\linewidth}
        \centering
        \includegraphics[width=0.993\textwidth,height=0.4in]{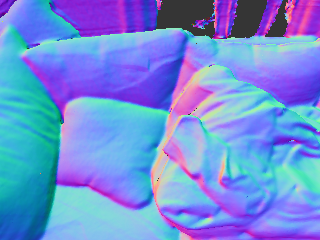}\\
        \includegraphics[width=0.993\textwidth,height=0.4in]{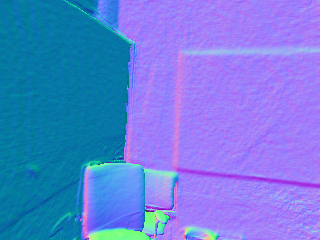}\\ 
        \includegraphics[width=0.993\textwidth,height=0.4in]{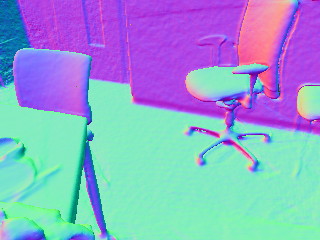}\\   
        \includegraphics[width=0.993\textwidth,height=0.4in]{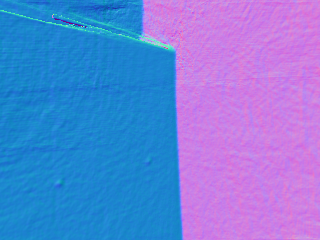}\\ 
        \footnotesize (e) GT

    \end{minipage}%
    \begin{minipage}{0.15\linewidth}
        \centering
        \includegraphics[width=0.993\textwidth,height=0.4in]{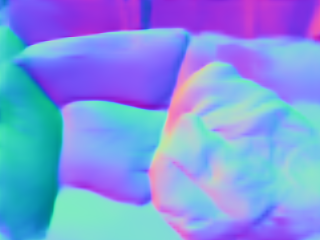}\\
        \includegraphics[width=0.993\textwidth,height=0.4in]{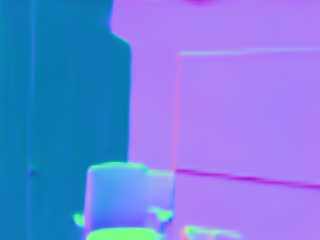}\\ 
        \includegraphics[width=0.993\textwidth,height=0.4in]{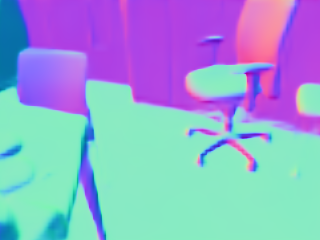}\\   
        \includegraphics[width=0.993\textwidth,height=0.4in]{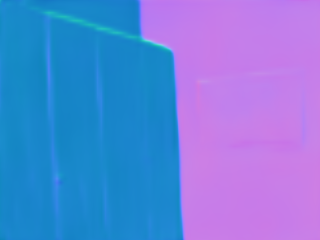}\\ 
        \footnotesize (f) Pred.

    \end{minipage}%
    \caption{Qualitative examples on ScanNet dataset.}
    \label{fig:vis_sn}
\end{figure}

\noindent\textbf{Surface Normal Estimation.} We compare VISTA-Net with the state-of-the-art RGB-based methods, including Eigen~\etal~\cite{eigen2015predicting}, GeoNe~\cite{qi2018geonet} and FrameNet~\cite{huang2019framenet}. We adapt the publicly available training code and keep their fine-tune and pre-train model. The results shown in Table~\ref{tab:overall_sn}. Our method outperforms the state-of-the-art on both the error metric and accuracy metric. The qualitative results are shown in Fig.~\ref{fig:vis_sn}.
{They confirms the benefit of the proposed approach in joint structured spatial-channel attention estimation for wide range of deep representation learning tasks.}

\begin{figure*}[h]
\centering
    \begin{minipage}{0.15\linewidth}
        \centering
        \includegraphics[width=0.993\textwidth,height=0.4in]{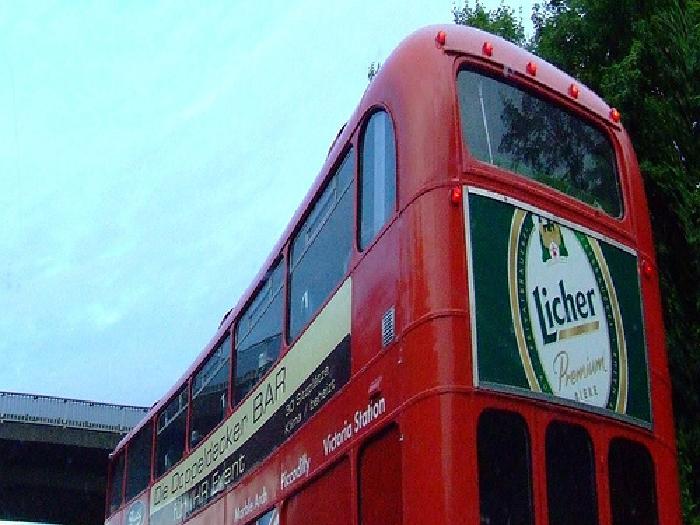}\\
        \includegraphics[width=0.993\textwidth,height=0.4in]{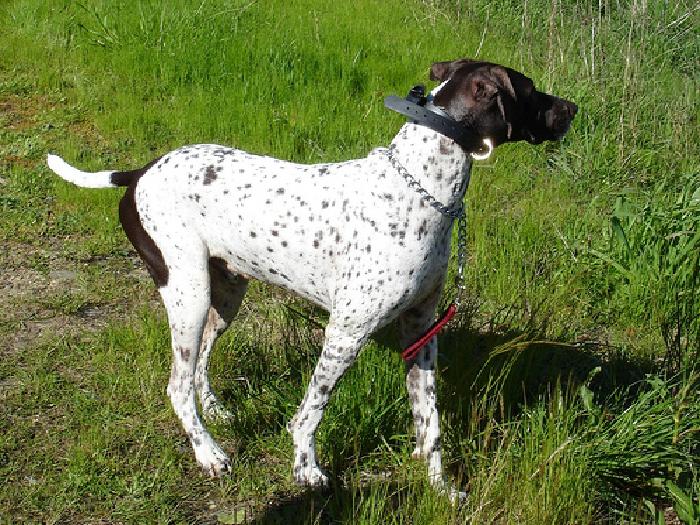}\\ 
        \includegraphics[width=0.993\textwidth,height=0.4in]{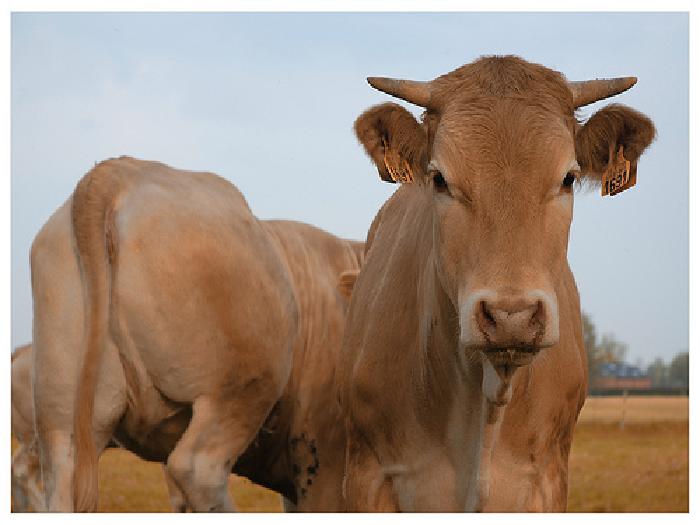}\\ 
        \includegraphics[width=0.993\textwidth,height=0.4in]{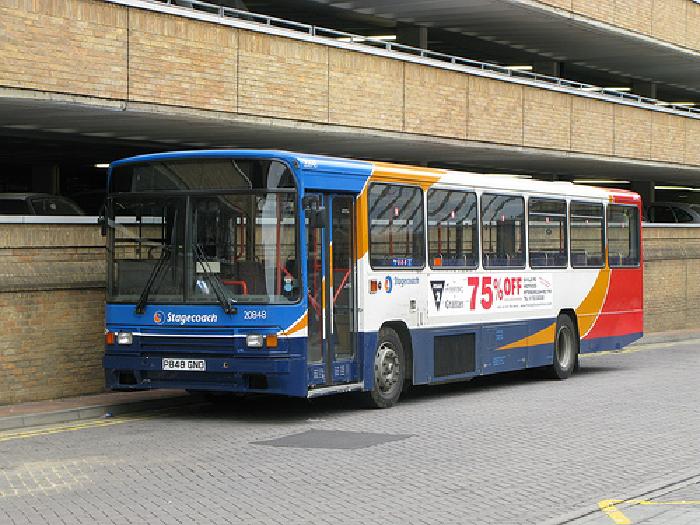}\\ 
        \footnotesize (a) Image
    \end{minipage}%
    \begin{minipage}{0.15\linewidth}
        \centering
        \includegraphics[width=0.993\textwidth,height=0.4in]{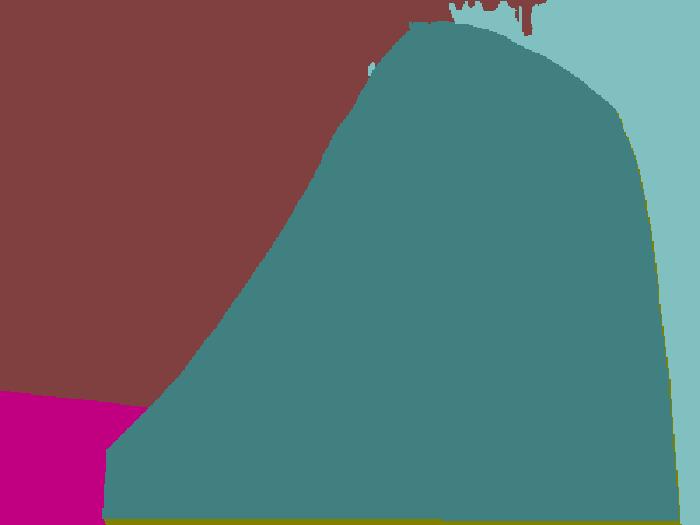}\\
        \includegraphics[width=0.993\textwidth,height=0.4in]{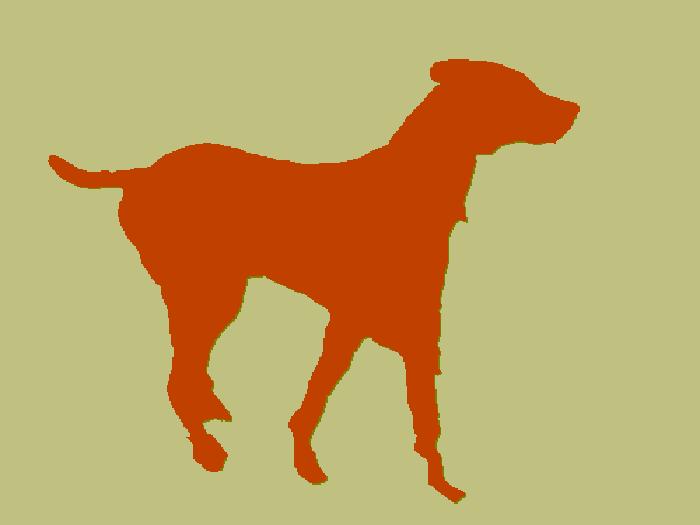}\\ 
        \includegraphics[width=0.993\textwidth,height=0.4in]{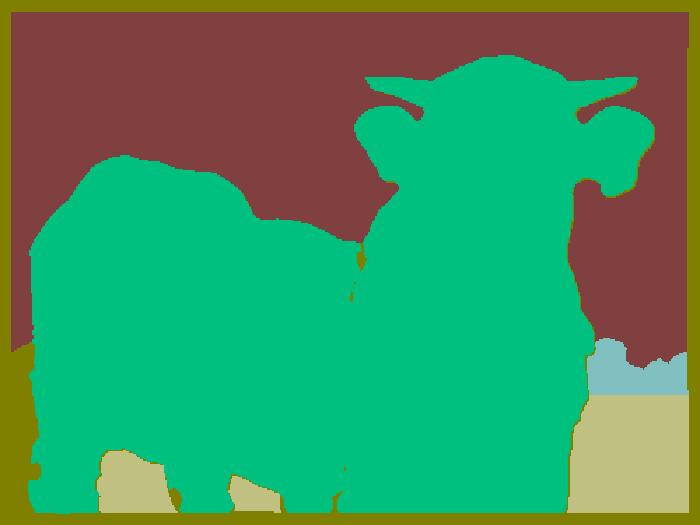}\\ 
        \includegraphics[width=0.993\textwidth,height=0.4in]{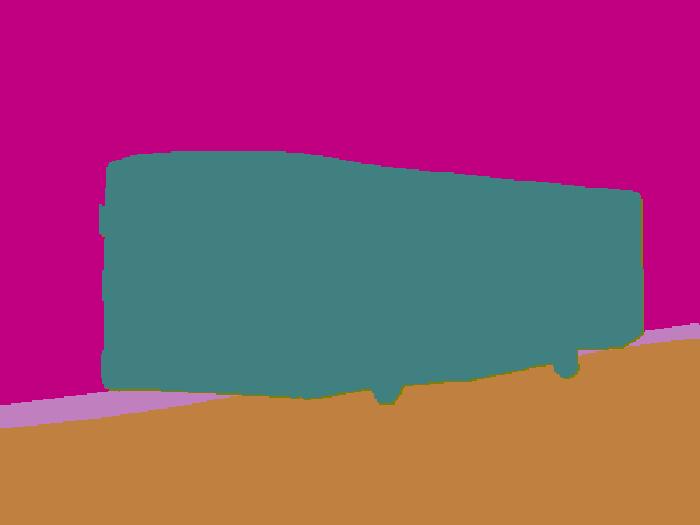}\\
        \footnotesize (b) GT
    \end{minipage}%
    \begin{minipage}{0.15\linewidth}
        \centering
        \includegraphics[width=0.993\textwidth,height=0.4in]{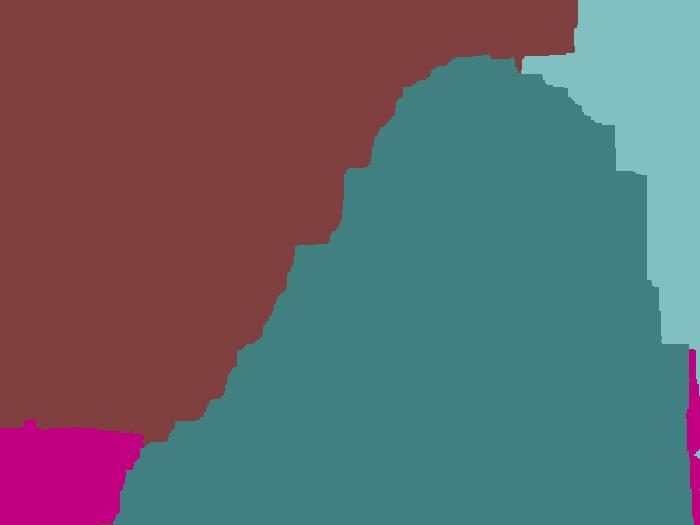}\\
        \includegraphics[width=0.993\textwidth,height=0.4in]{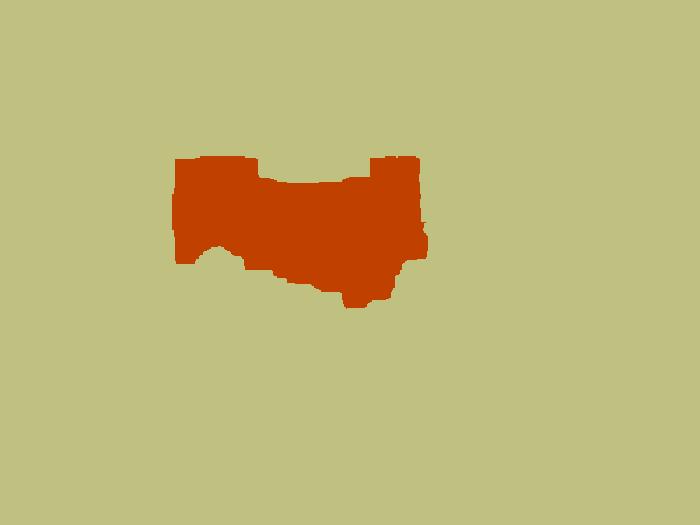}\\ 
        \includegraphics[width=0.993\textwidth,height=0.4in]{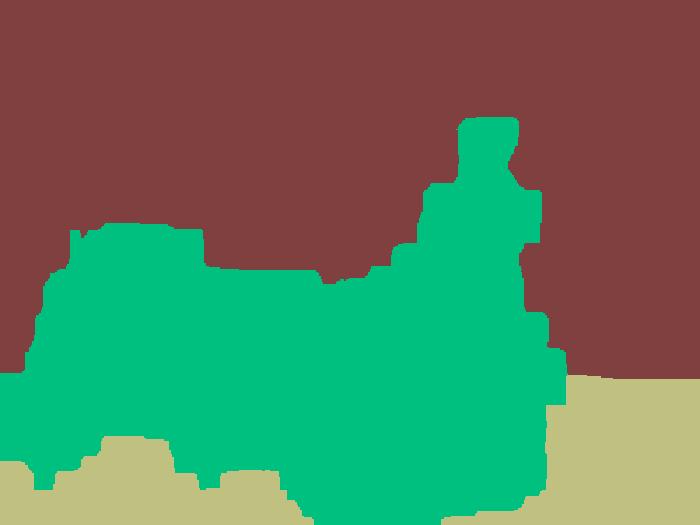}\\ 
        \includegraphics[width=0.993\textwidth,height=0.4in]{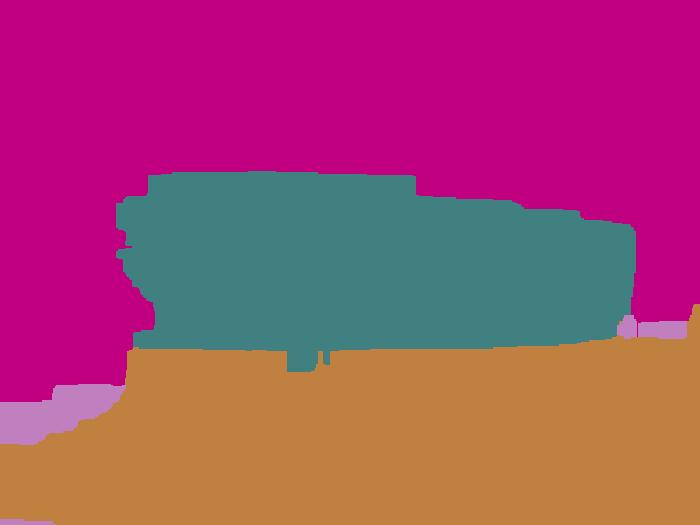}\\   
        \footnotesize (c) W/o $Att$
    \end{minipage}%
    \begin{minipage}{0.15\linewidth}
        \centering
        \includegraphics[width=0.993\textwidth,height=0.4in]{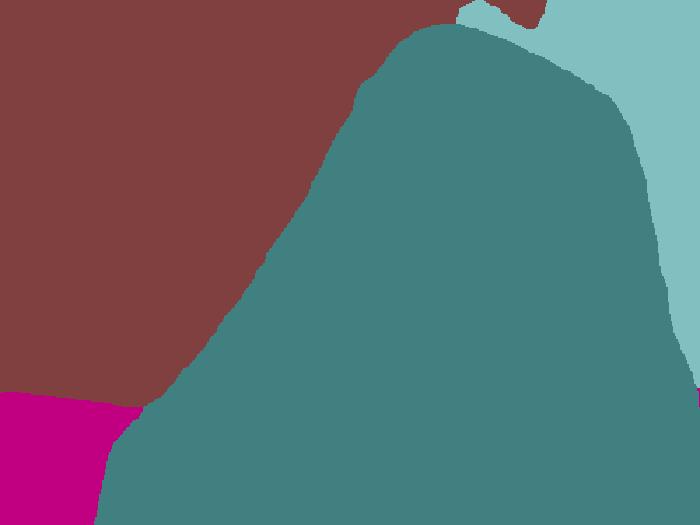}\\
        \includegraphics[width=0.993\textwidth,height=0.4in]{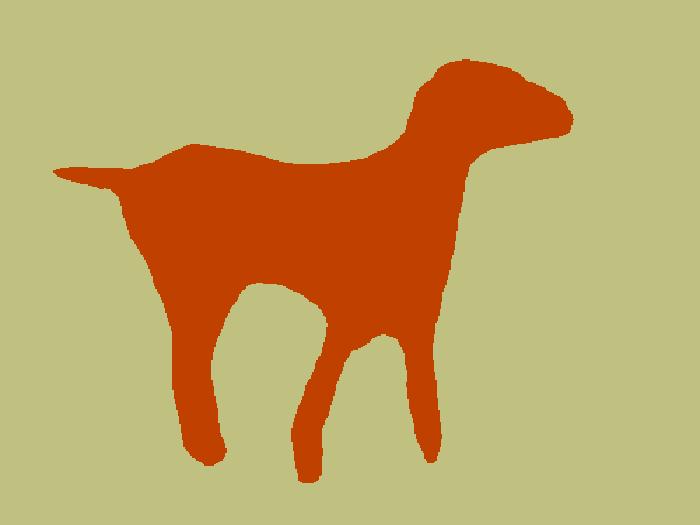}\\ 
        \includegraphics[width=0.993\textwidth,height=0.4in]{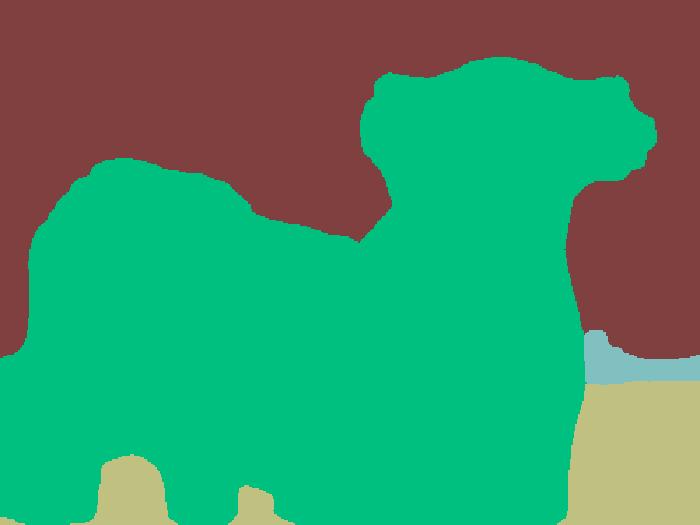}\\ 
        \includegraphics[width=0.993\textwidth,height=0.4in]{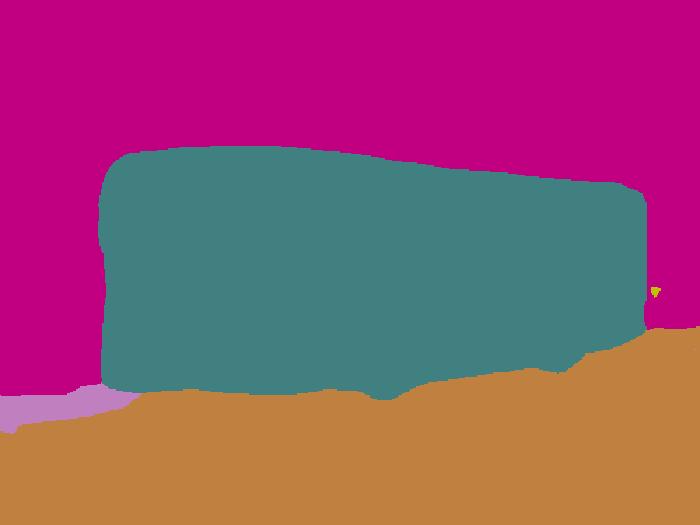}\\    
        \footnotesize (d) W/ $Att_{sp}$

    \end{minipage}%
    \begin{minipage}{0.15\linewidth}
        \centering
        \includegraphics[width=0.993\textwidth,height=0.4in]{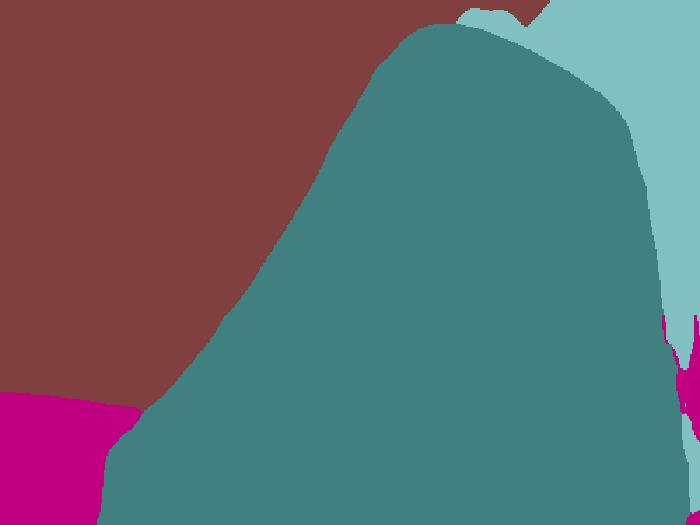}\\
        \includegraphics[width=0.993\textwidth,height=0.4in]{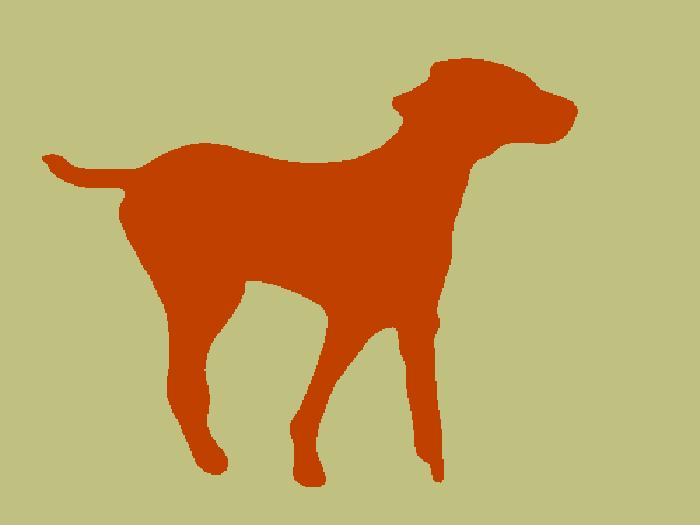}\\ 
        \includegraphics[width=0.993\textwidth,height=0.4in]{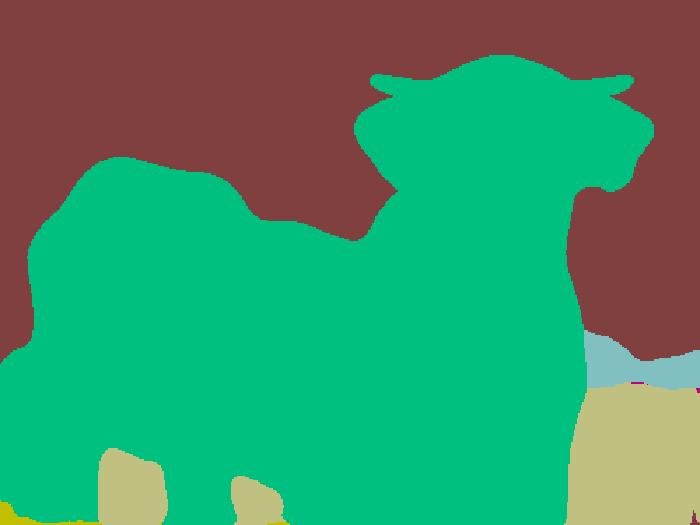}\\ 
        \includegraphics[width=0.993\textwidth,height=0.4in]{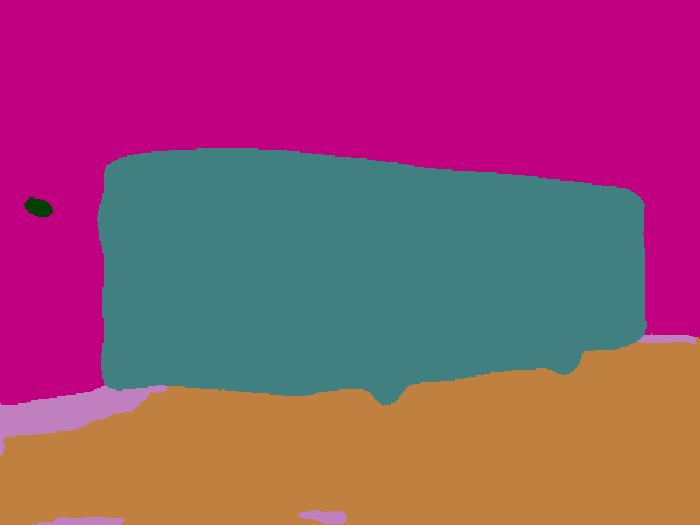}\\   
        \footnotesize (e) W/ $Att_{ch}$

    \end{minipage}%
    \begin{minipage}{0.15\linewidth}
        \centering
        \includegraphics[width=0.993\textwidth,height=0.4in]{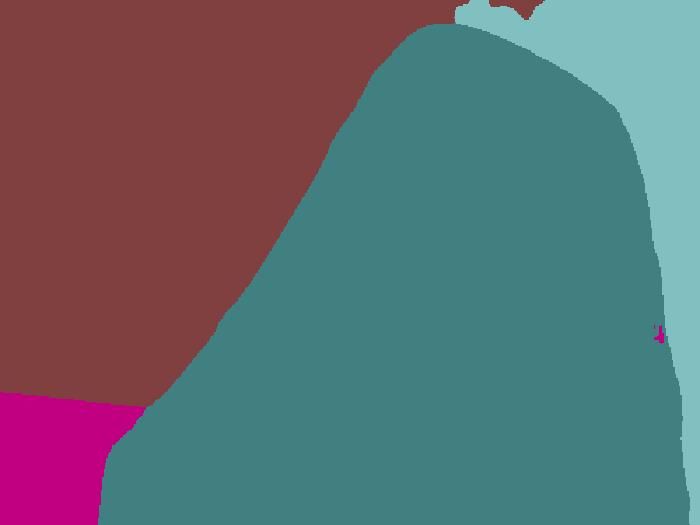}\\
        \includegraphics[width=0.993\textwidth,height=0.4in]{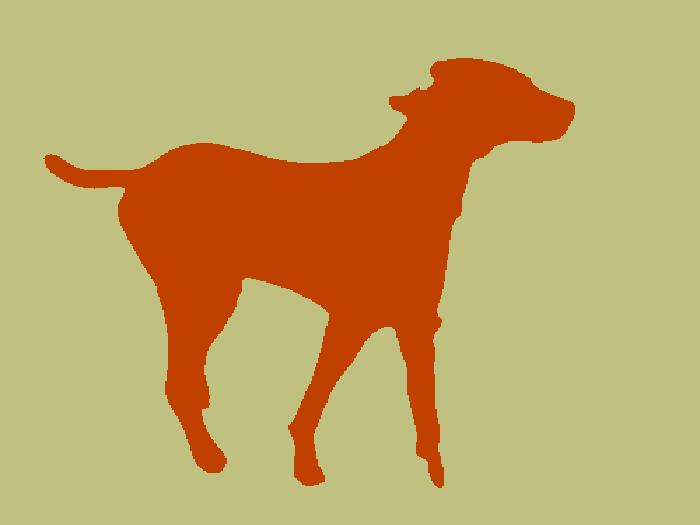}\\ 
        \includegraphics[width=0.993\textwidth,height=0.4in]{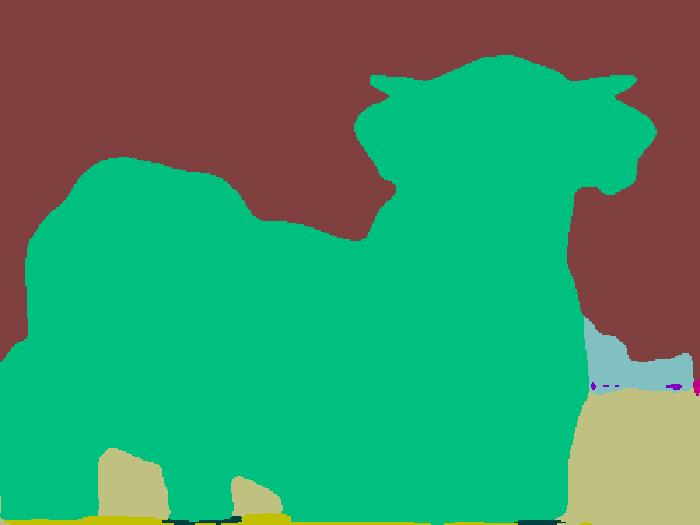}\\ 
        \includegraphics[width=0.993\textwidth,height=0.4in]{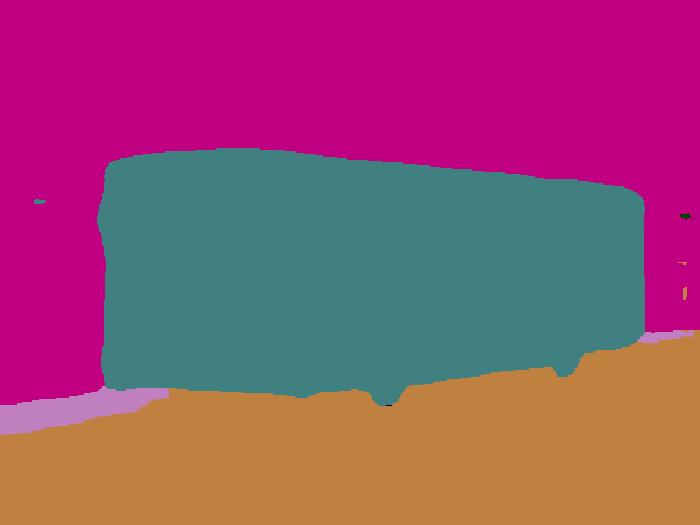}\\ 
        \footnotesize (f) Full

    \end{minipage}%
    \caption{Comparison of different variations of VISTA-Net on Pascal-context dataset. }
    \label{fig:pcontext_ab}
\end{figure*}

\begin{table}[]
        \centering 
        \caption{Ablation study on the Pascal-context dataset: performance of VISTA-Net  for different attention mechanisms and scales.}
        \resizebox{0.43\textwidth}{!}{%
            \begin{tabular}{lcccc}
            \toprule[1.2pt]
            Scales & Structured Attention   & Probabilistic & mIoU & PixAcc \\ \midrule
             DANet \cite{fu2019dual} & Separate Attention & No & 52.6 & - \\ \midrule
            \multirow{5}{*}{Single scale} & No structure & Yes & 51.7 & 78.9 \\ 
            & Spatial & Yes & 53.0 & 79.7\\
            & Channel & Yes & 53.1 & 79.8\\
            & Low-rank tensor & No & 53.2 & 79.9 \\
            & High-rank tensor & Yes & 53.9 & 80.3 \\
             \midrule
            \multirow{5}{*}{Multiple scale} & No structure & Yes & 52.8 & 79.5 \\
            & Spatial & Yes & 54.8 & 80.8 \\
            & Channel & Yes & 54.6 & 80.6 \\
            & Low-rank tensor & No & 54.7 & 80.7 \\
            & High-rank tensor & Yes & \textbf{56.1} & \textbf{81.7}\\
            \bottomrule[1.2pt] 
        \end{tabular}}
    \label{tab:ab_pcontext}
\end{table}
\begin{table}[]
\centering
\caption{ Computational cost analysis for different values of $T$ on Pascal-context dataset. }
\label{tab:pcontext efficiency}
\resizebox{0.4\textwidth}{!}{%
\begin{tabular}{cccccc}
\toprule
\multicolumn{1}{l}{Rank} & \multicolumn{1}{l}{IoU} & \multicolumn{1}{l}{pixAcc} & \multicolumn{1}{l}{\# param.} &
\multicolumn{1}{c}{{GFLOPs}} & \multicolumn{1}{c}{FPS}  \\
\midrule
0 & 54.2 & 80.4 & \textbf{45.80M} & {\textbf{738.8}} & \textbf{1.106$\pm $0.046}  \\
1 & 55.4 & 81.1 & 49.85M & {804.2} & $1.075\pm 0.018 $ \\
3 & 55.6 & 81.2 & 52.68M & {849.8} & $1.011\pm 0.019$  \\
5 & 55.5 & 80.9 & 54.89M & {885.4} & $1.068\pm 0.026$  \\
7 & 55.9 & 81.2 & 56.79M & {916.1} & $0.957\pm 0.021$  \\
9 & \textbf{56.1} & \textbf{81.7} & 58.85M & {949.3} & $0.868\pm 0.014$ \\
\bottomrule
\end{tabular}%
}
\end{table}

\par
\begin{figure}
    \centering
    \includegraphics[width=0.35\textwidth]{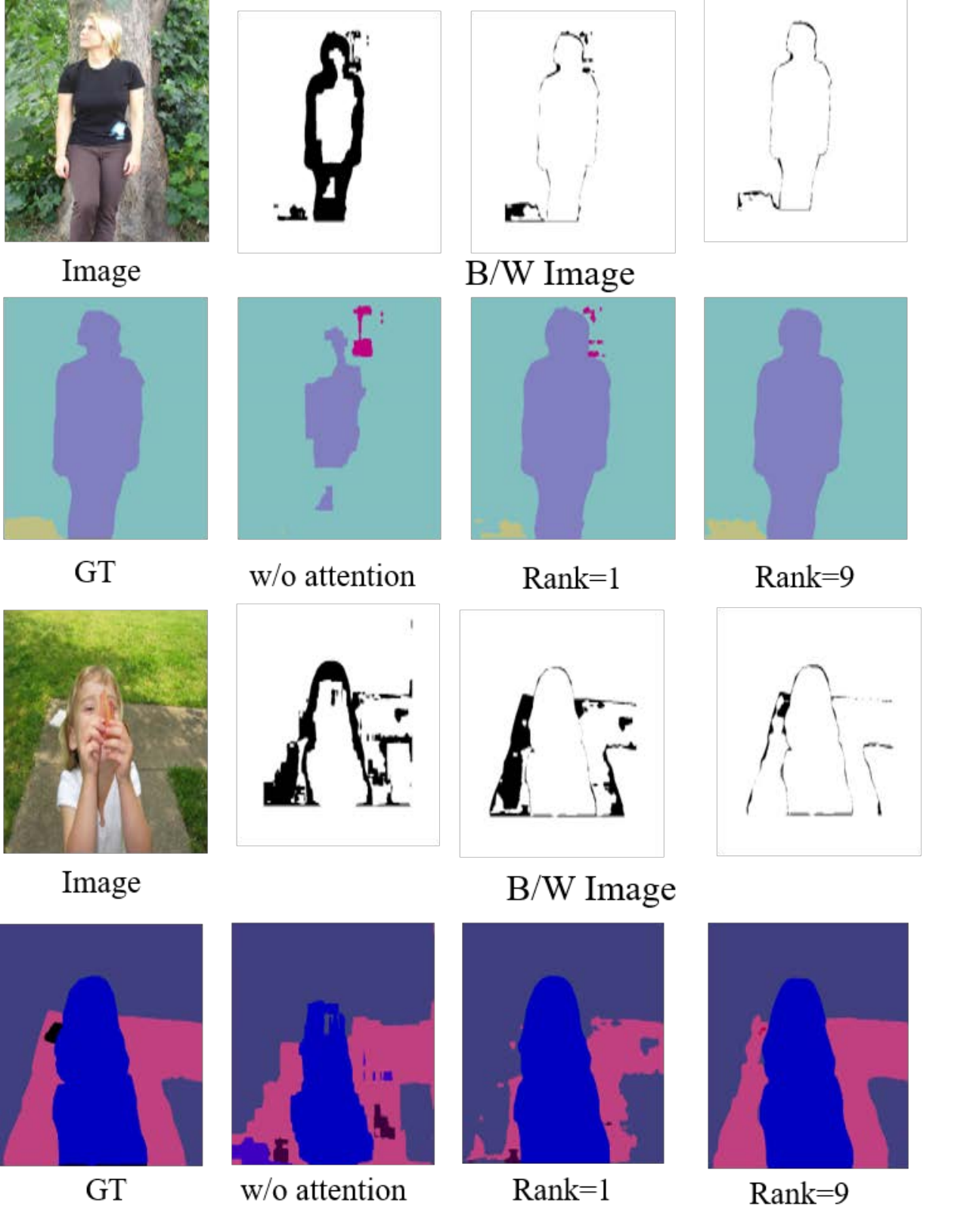}
    \caption{Pascal-context dataset. Comparison of different variations of VISTA-Net.}
    \label{fig:pcontext_sup}
\end{figure}

\noindent\textbf{Ablation Study.}
We also perform an ablation study on the Pascal-context dataset to further demonstrate each proposed component's impact. {Table.~\ref{tab:ab_pcontext} shows that the performance of VISTA-Net degrades not only when the model does not employ the structured attention mechanism but also when only channel-wise (Fig.\ref{fig:teaser}.(a)) or spatial-wise attention (Fig.\ref{fig:teaser}.(b)) is used. Moreover, we can also see the advantage of using the proposed probabilistic formulation for joint modeling both spatial- and channel-wise attention in a principled manner. For the sake of completeness, we also report the results of DANet, which corresponds to separate spatial and channel-wise attention (Fig.\ref{fig:teaser}.(c))}
Meanwhile, Fig.~\ref{fig:pcontext_ab} depicts segmentation maps obtained on the Pascal-Context dataset using different versions of our method. In particular, we visualize (c) VISTA-Net w/o attention, (d) VISTA-Net w/o Spatial Attention, (e) VISTA-Net w/o Channel Attention, and (f) VISTA-Net (full model). From left to right, the results become more similar to the ground truth, indicating our proposed attention model's clear advantage.
Interestingly, the performance achieved in each of the variants (spatial, channel) is similar. This leads us to believe that the proposed method's competitive advantage is combining structured attention with a probabilistic formulation. Notably, the feature refinement through message passing seems to be the most crucial contribution to improving performance.  In Table.~\ref{tab:pcontext efficiency} we show the results of our experiments in order to analyze the computational cost of our method. In particular, we perform an analysis on the Pascal-context dataset and at varying $T$. In the Table~\ref{tab:pcontext efficiency}, FPS means Frames Per Second. We run 10 times experiments and provide mean and variance of FPS. Meanwhile, we use input feature map of size $480\times 480$ to evaluate their complexity during inference.
{As expected, we notice an increased computational burden while augmenting the rank. This, however, has been rewarded with a noticeable increase in performance of the model. Table~\ref{tab:pcontext efficiency} shows the efficiency of the model at the increasing of the rank also in terms of floating-point operations per second (FLOPs).}
Finally, in Fig.~\ref{fig:pcontext_sup} we propose a few qualitative results compared with their B/W images on the Pascal-context dataset. It is shown the importance of the attention model, and the result obtained increasing the iterations. In the odd rows are shown the misclassified pixels (in black). The image shows clearly how the proposed iterative approach based on message passing is beneficial for the final prediction.

\begin{table}[]
\centering
\caption{Complexity comparison on the Cityscapes validation dataset.}
\label{tab:complexity}
\resizebox{1\linewidth}{!}{
\begin{tabular}{lcccc}
\toprule
Method & backbone & parameters & {GFLOPs}& mIoU  \\
\midrule
DeepLabv3~\cite{chen2017rethinking} & D-ResNet-101 & \textbf{58.0M}      & {1778.7} &78.5 \\
PSPNet~\cite{zhao2016pyramid} & D-ResNet-101 & 65.9M      & {2017.6}& 79.7 \\
DANet~\cite{fu2019dual} & D-ResNet-101 & 70.1M      & {3938.2}& 81.5 \\
CCNet~\cite{huang2019ccnet} & D-ResNet-101 & 78.9M      & {2852.5}& 81.3 \\
HRNet~\cite{wang2020deep} & HRNetV2-W48  & 65.9M      & {\textbf{696.2}}& 81.1  \\
OCR~\cite{YuanCW19} & HRNetV2-W48  & 70.3M      & {1206.3 }& 81.8 \\
VISTA-Net & HRNetV2-W48  & 70.4M & {1320.7} & \textbf{82.3}\\
\bottomrule
\end{tabular}}
\end{table}

\par\noindent\textbf{Complexity.}
We compare the efficiency of our \method{} with the efficiencies of the multi-scale context schemes and the relational context schemes. 
{Table.~\ref{tab:complexity} provides the comparison with several representative methods on the Cityscapes validation set (single scale and no flipping) in terms of parameter and computational complexity. The GFLOPs is calculated on the input size $1024\times2048$. According to results in Table.~\ref{tab:complexity}, our structure attention's computation complexity (measured by the number of FLOPs) is quite close to OCR~\cite{YuanCW19}. Compared with OCR~\cite{YuanCW19}, though computation complexity has a slight increase, this increase is acceptable compared with the obvious performance improvement.
In general, \method{} can be considered a good trade-off between performance, parameters and GFLOPs.}

\subsection{{Qualitative Results of Learned Attentions}}

\begin{figure}[h]
\centering
\subfigure{
    \begin{minipage}{0.24\linewidth}
        \centering
        \includegraphics[width=0.993\textwidth,height=0.7in]{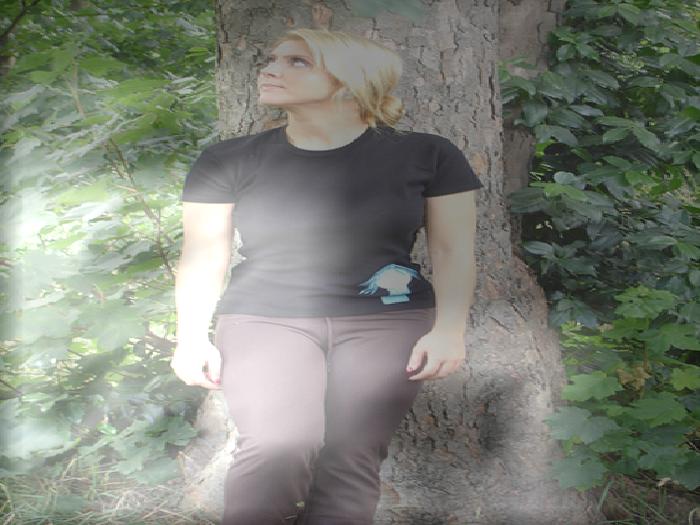}\\
        \includegraphics[width=0.993\textwidth,height=0.7in]{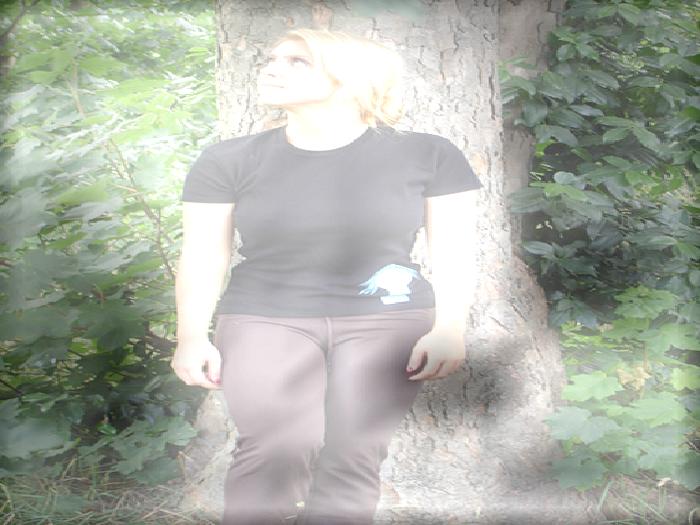}\\  
    \end{minipage}%
}%
\subfigure{
    \begin{minipage}{0.24\linewidth}
        \centering
        \includegraphics[width=0.993\textwidth,height=0.7in]{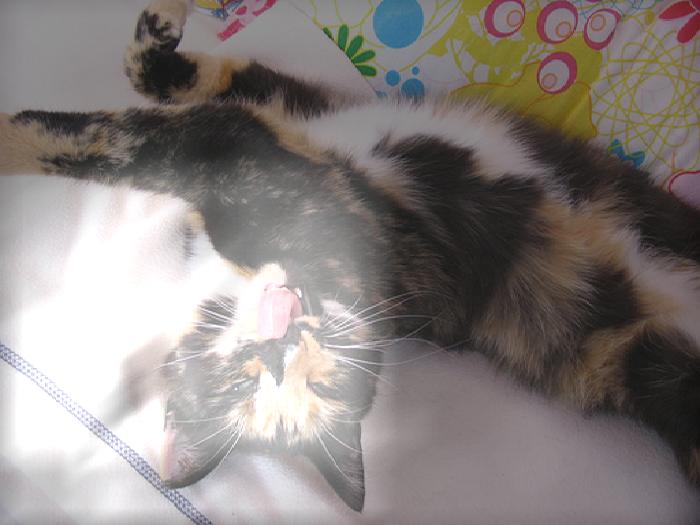}\\
        \includegraphics[width=0.993\textwidth,height=0.7in]{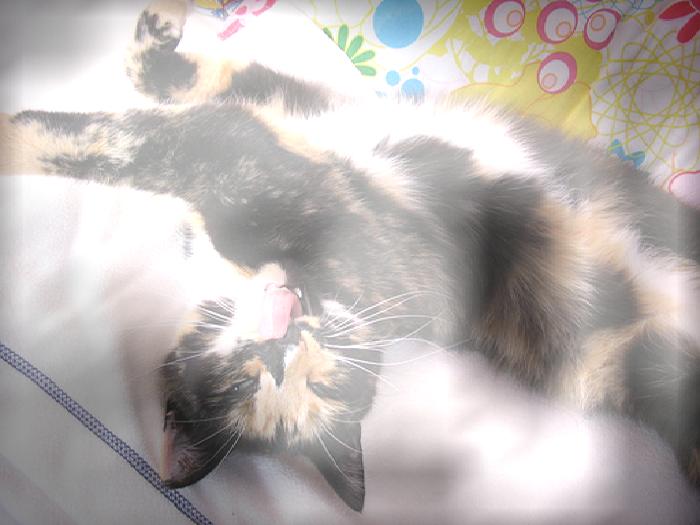}\\  
    \end{minipage}%
}%
\subfigure{
    \begin{minipage}{0.24\linewidth}
        \centering
        \includegraphics[width=0.993\textwidth,height=0.7in]{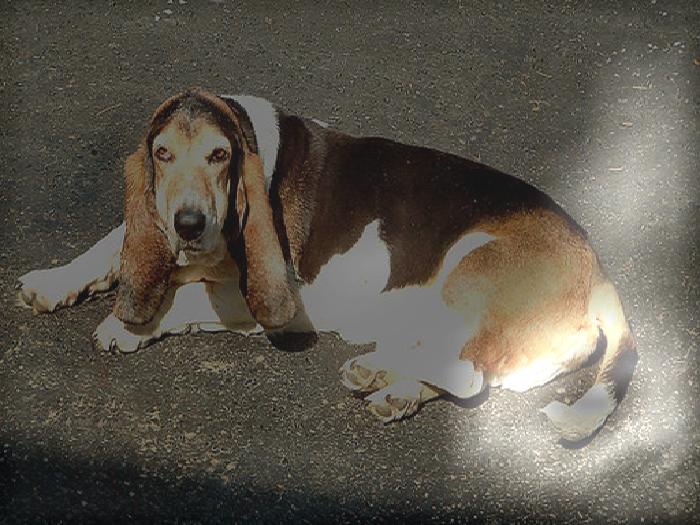}\\
        \includegraphics[width=0.993\textwidth,height=0.7in]{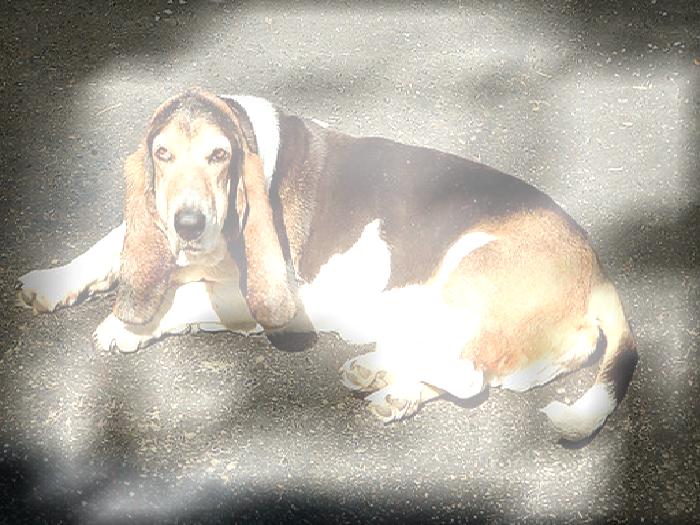}\\
    \end{minipage}%
}%
\subfigure{
    \begin{minipage}{0.24\linewidth}
        \centering
        \includegraphics[width=0.993\textwidth,height=0.7in]{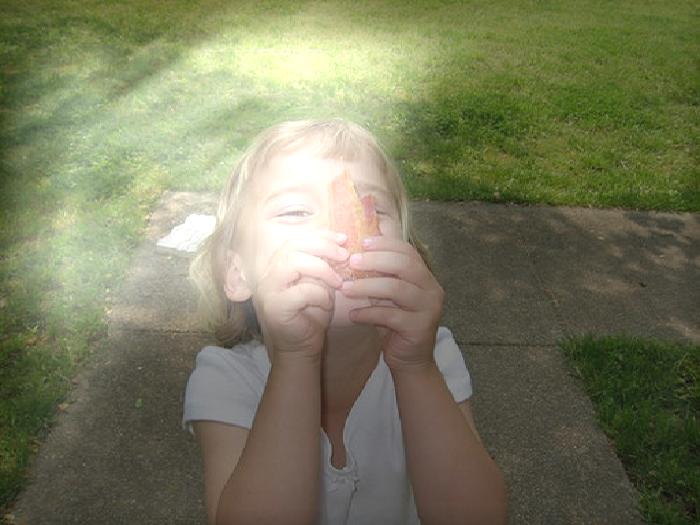}\\
        \includegraphics[width=0.993\textwidth,height=0.7in]{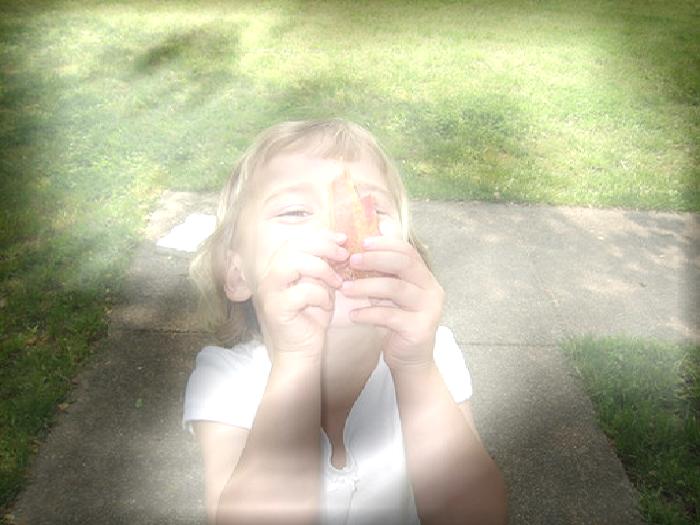}\\
    \end{minipage}%
}%
    \caption{Attention map visualisation on Pascal context dataset. First row: structured attention tensor $\vect{a}$, see~(\ref{eq:structured-attention}). Second row, spatial attention maps $\vect{m}$.}
    \label{fig:attention}
\end{figure}

\begin{figure}[!t]
\centering
\subfigure{
    \begin{minipage}{0.19\linewidth}
        \centering
        \includegraphics[width=0.993\textwidth,height=0.5in]{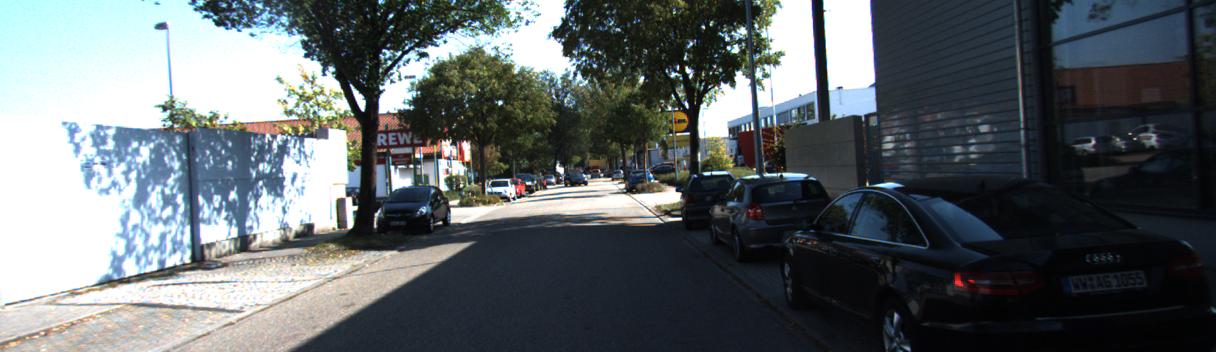}\\  
        \includegraphics[width=0.993\textwidth,height=0.5in]{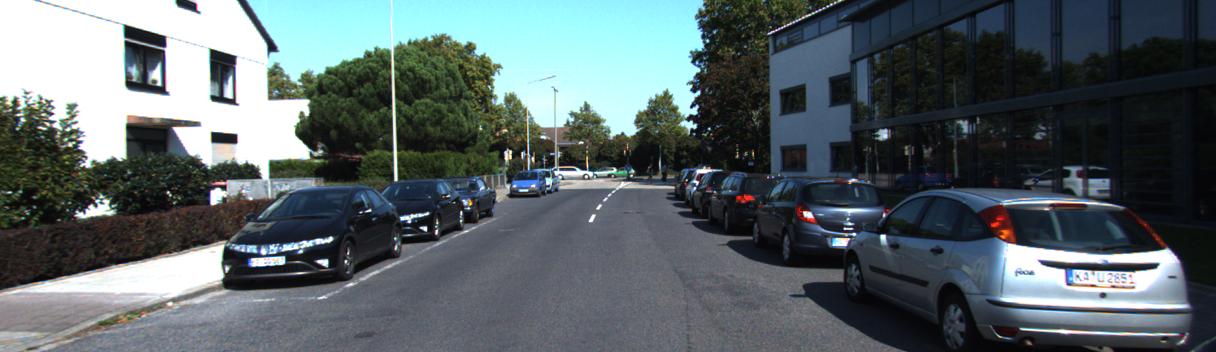}\\  
        \includegraphics[width=0.993\textwidth,height=0.5in]{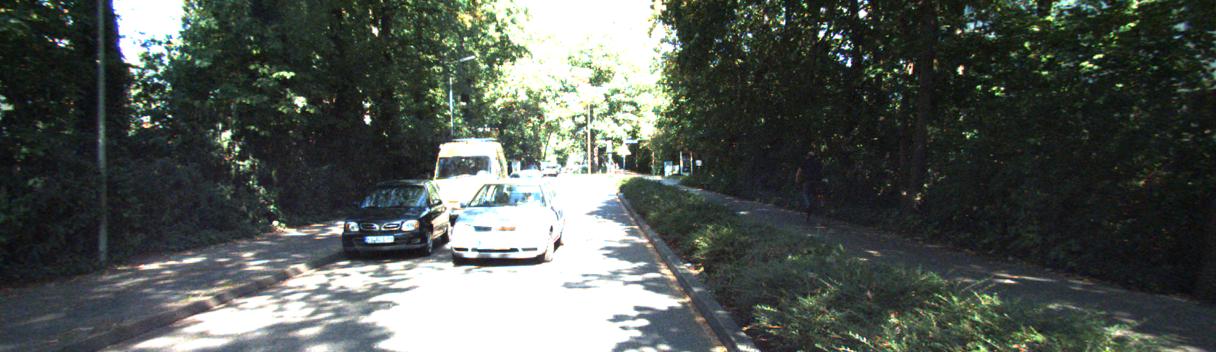}\\  
        \includegraphics[width=0.993\textwidth,height=0.5in]{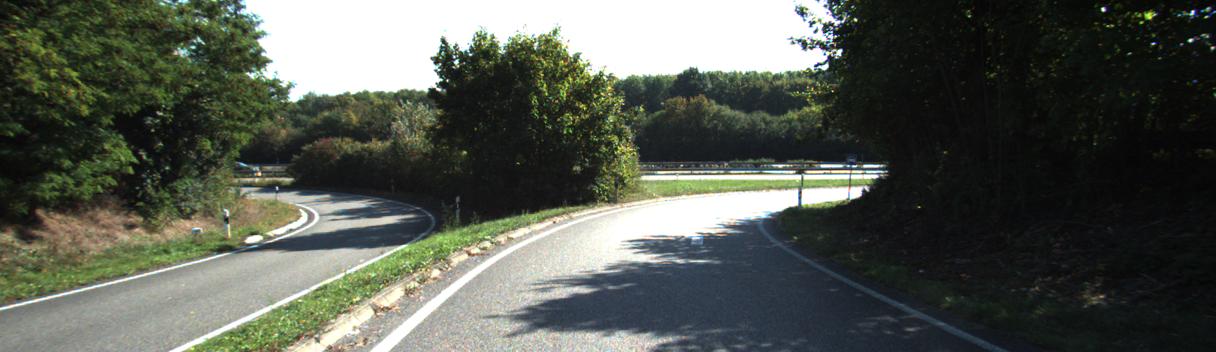}\\  
        \includegraphics[width=0.993\textwidth,height=0.5in]{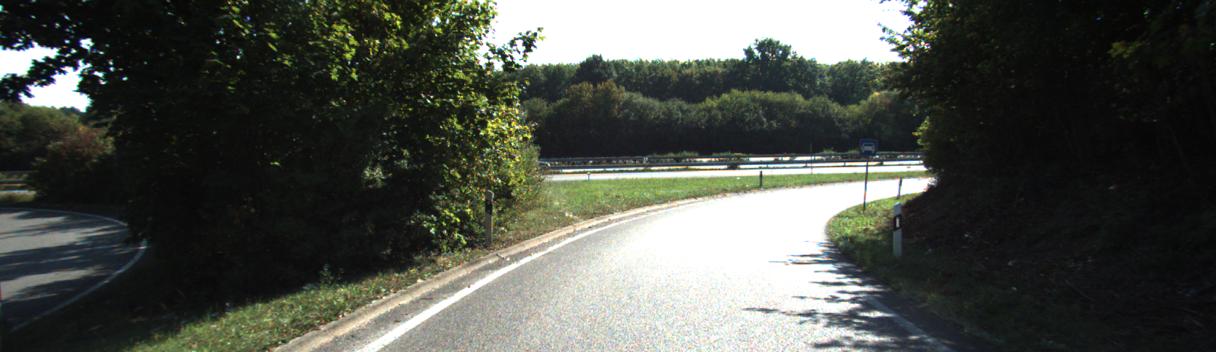}\\  
        \includegraphics[width=0.993\textwidth,height=0.5in]{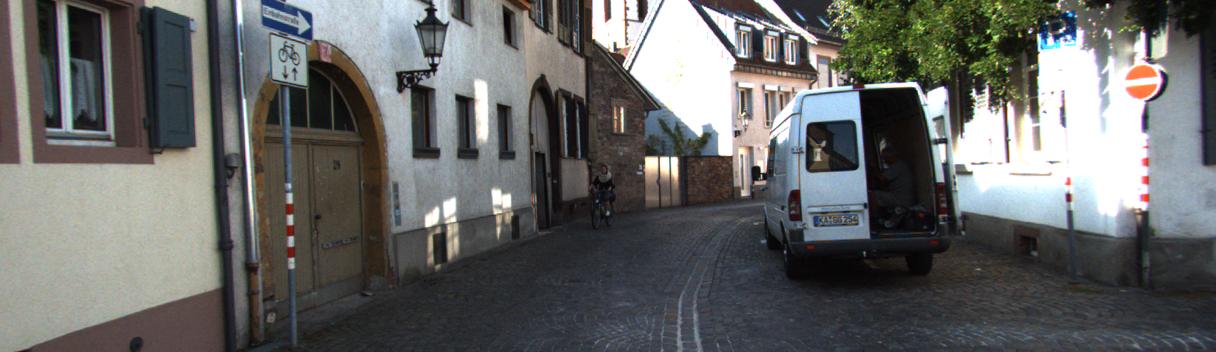}\\  
    \end{minipage}%
}%
\subfigure{
    \begin{minipage}{0.19\linewidth}
        \centering
        \includegraphics[width=0.993\textwidth,height=0.5in]{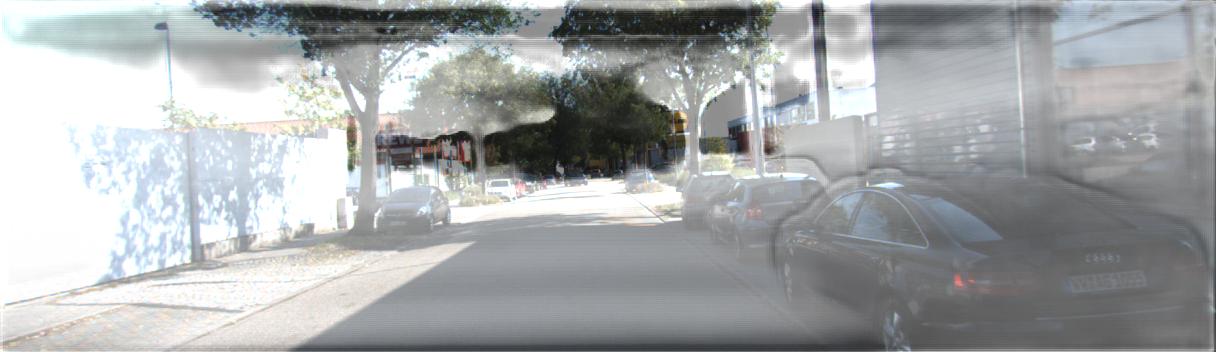}\\  
        \includegraphics[width=0.993\textwidth,height=0.5in]{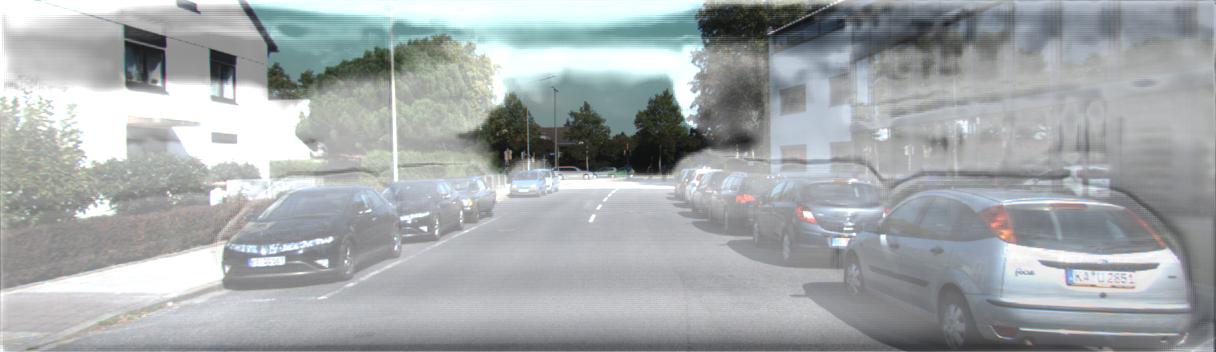}\\  
        \includegraphics[width=0.993\textwidth,height=0.5in]{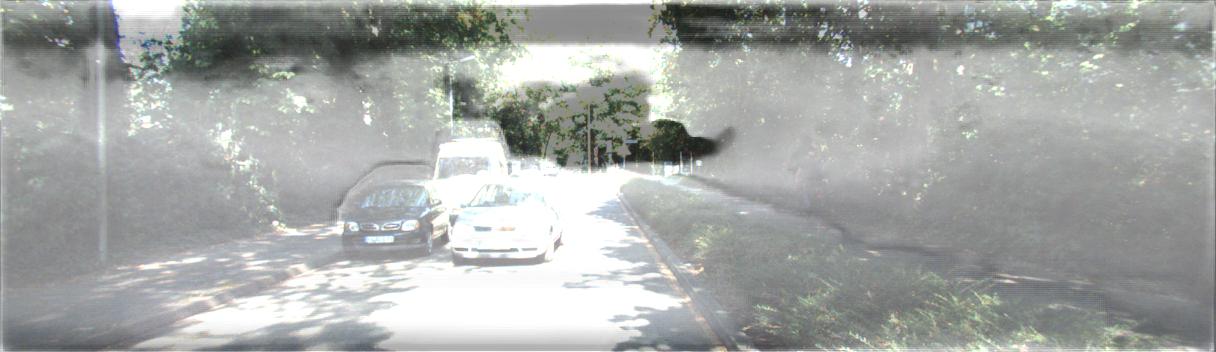}\\  
        \includegraphics[width=0.993\textwidth,height=0.5in]{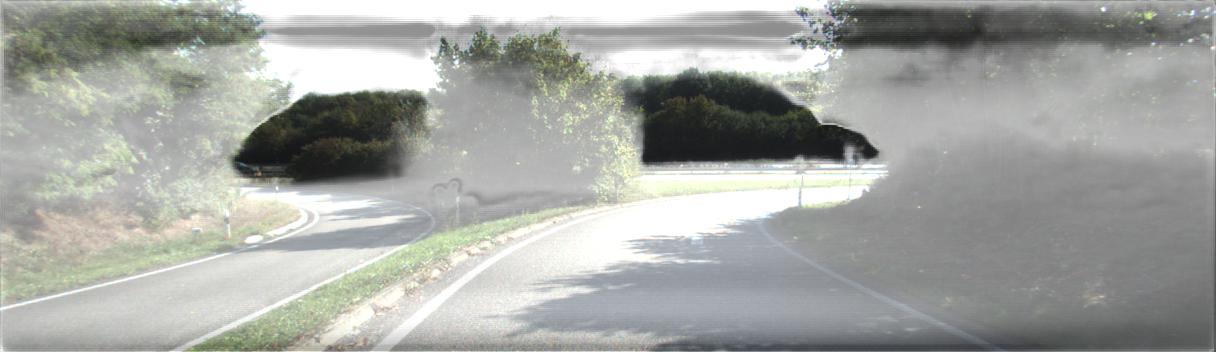}\\  
        \includegraphics[width=0.993\textwidth,height=0.5in]{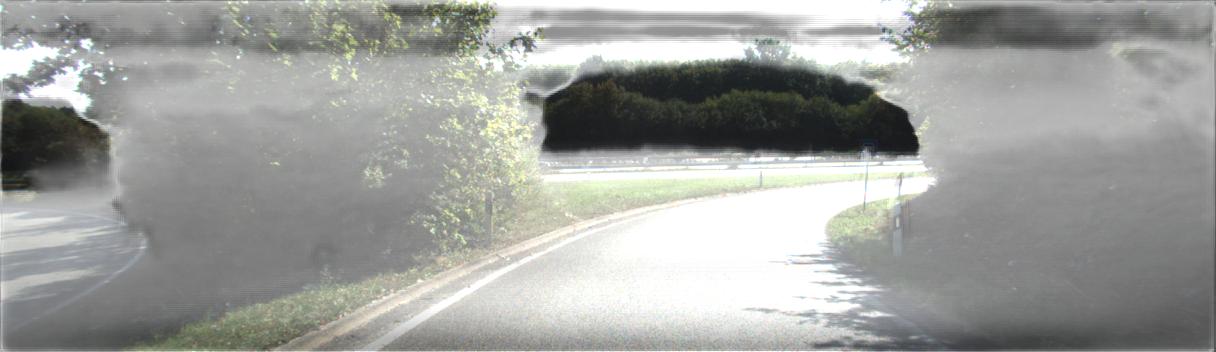}\\  
        \includegraphics[width=0.993\textwidth,height=0.5in]{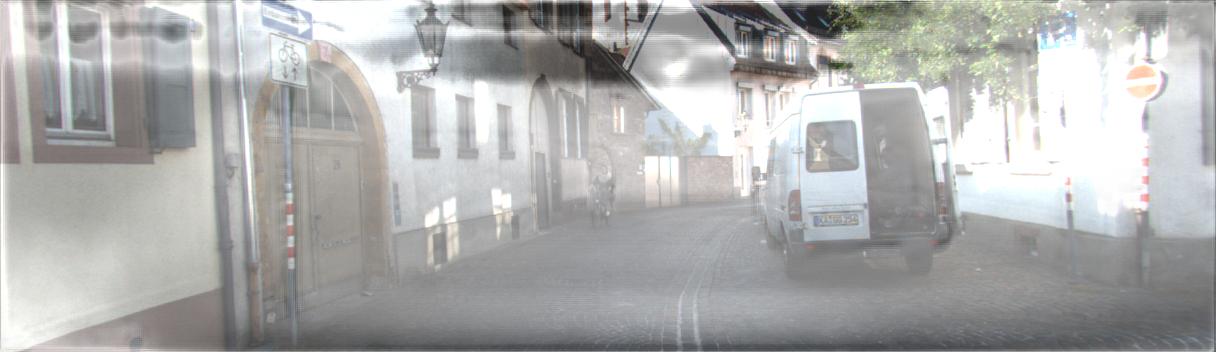}\\ 
    \end{minipage}%
}%
\subfigure{
    \begin{minipage}{0.19\linewidth}
        \centering
        \includegraphics[width=0.993\textwidth,height=0.5in]{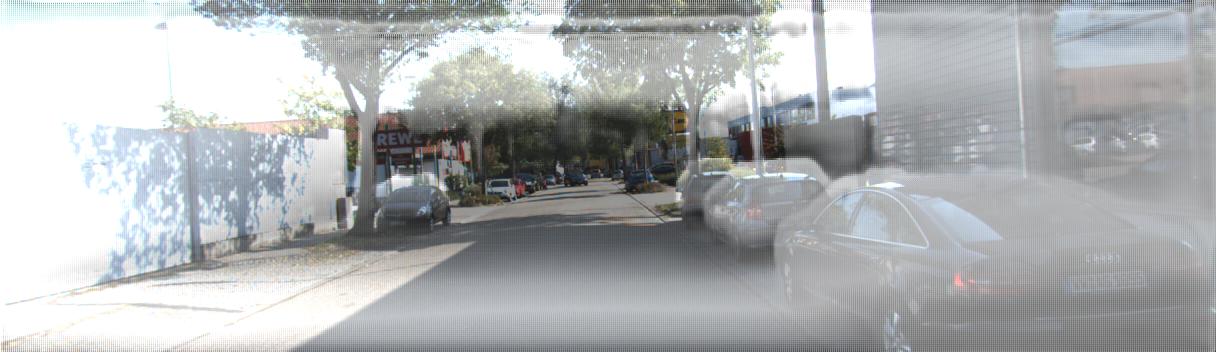}\\  
        \includegraphics[width=0.993\textwidth,height=0.5in]{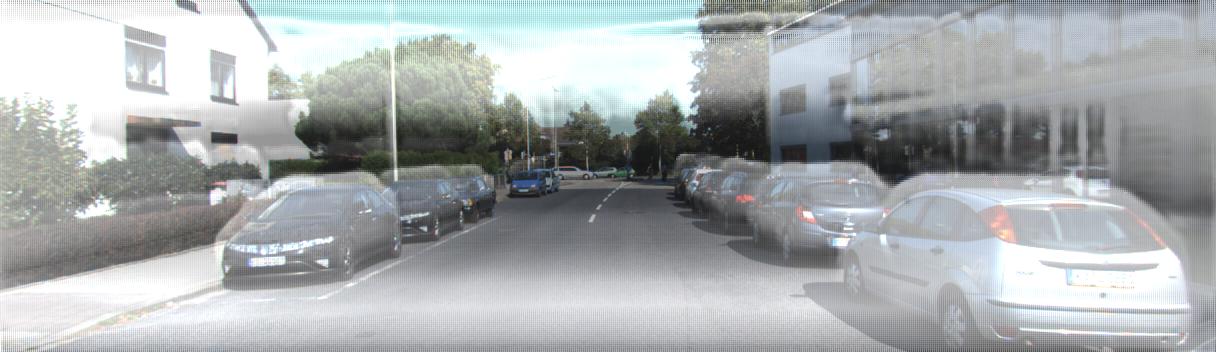}\\  
        \includegraphics[width=0.993\textwidth,height=0.5in]{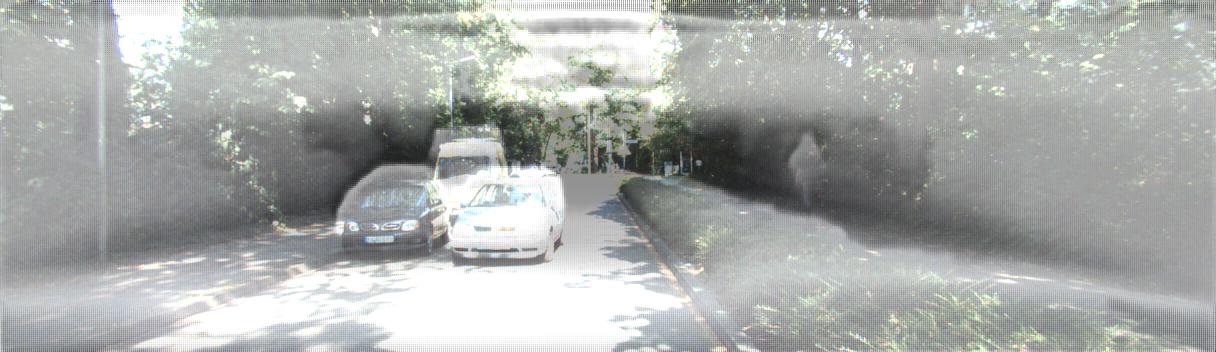}\\  
        \includegraphics[width=0.993\textwidth,height=0.5in]{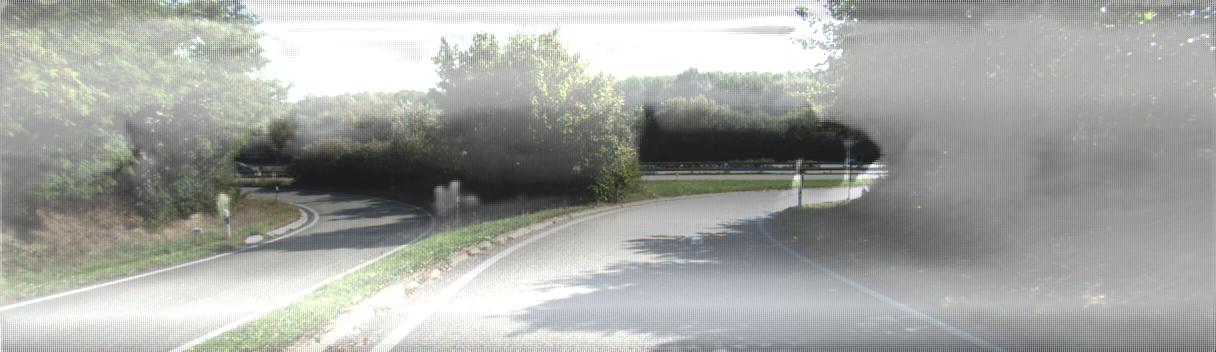}\\  
        \includegraphics[width=0.993\textwidth,height=0.5in]{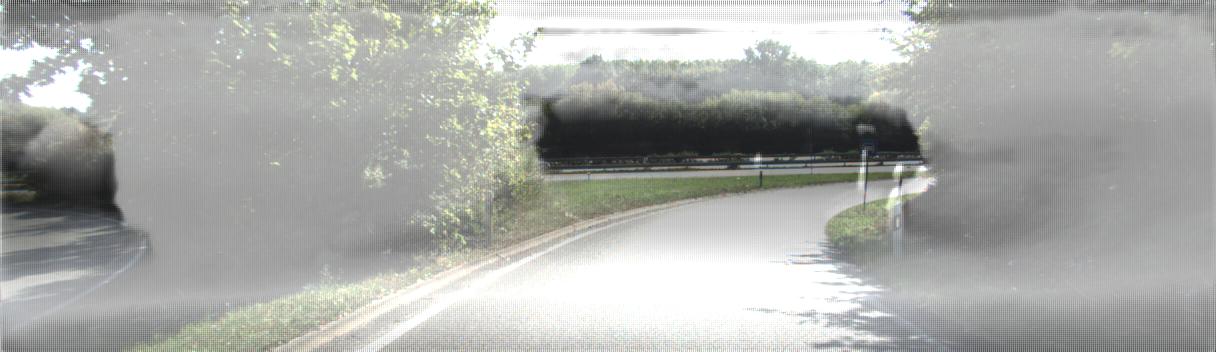}\\  
        \includegraphics[width=0.993\textwidth,height=0.5in]{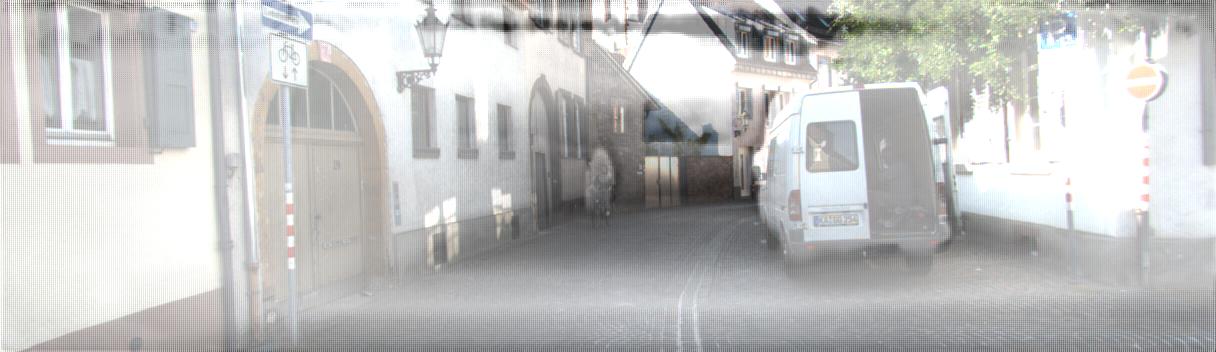}\\ 
    \end{minipage}%
}%
\subfigure{
    \begin{minipage}{0.19\linewidth}
        \centering
        \includegraphics[width=0.993\textwidth,height=0.5in]{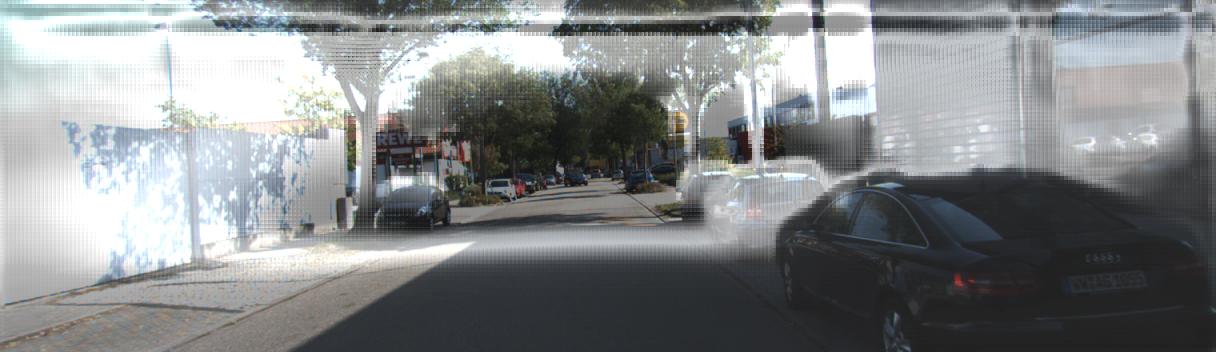}\\  
        \includegraphics[width=0.993\textwidth,height=0.5in]{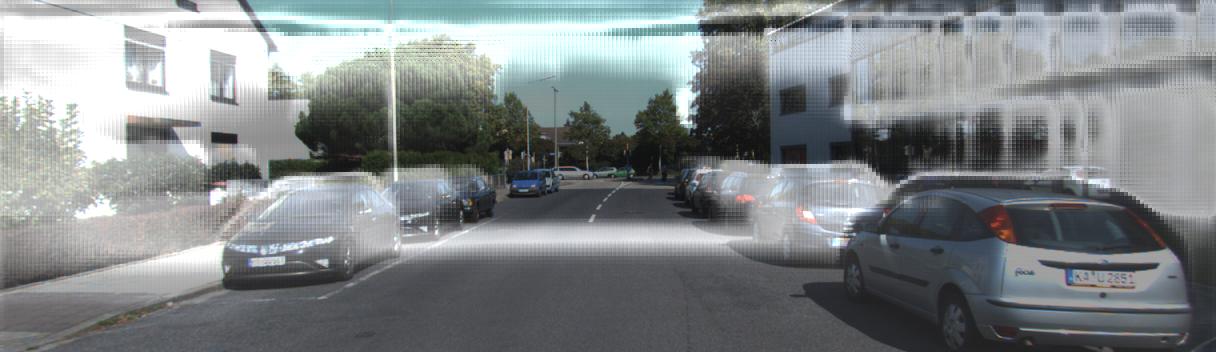}\\  
        \includegraphics[width=0.993\textwidth,height=0.5in]{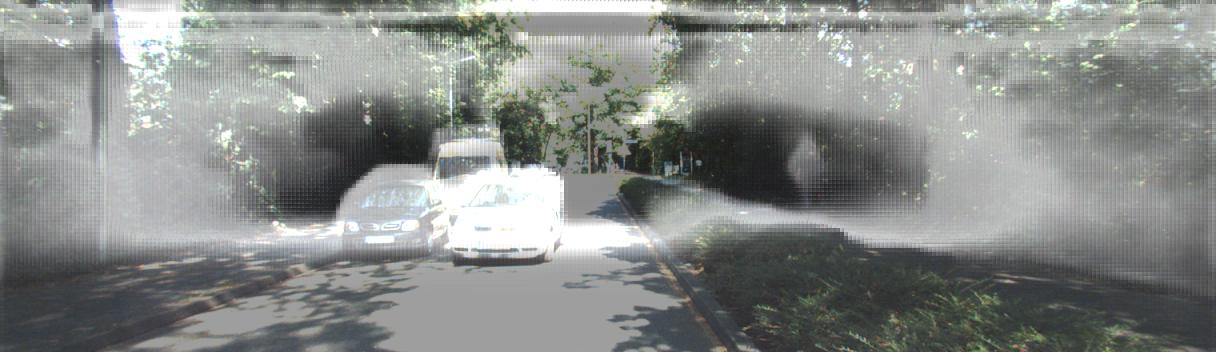}\\  
        \includegraphics[width=0.993\textwidth,height=0.5in]{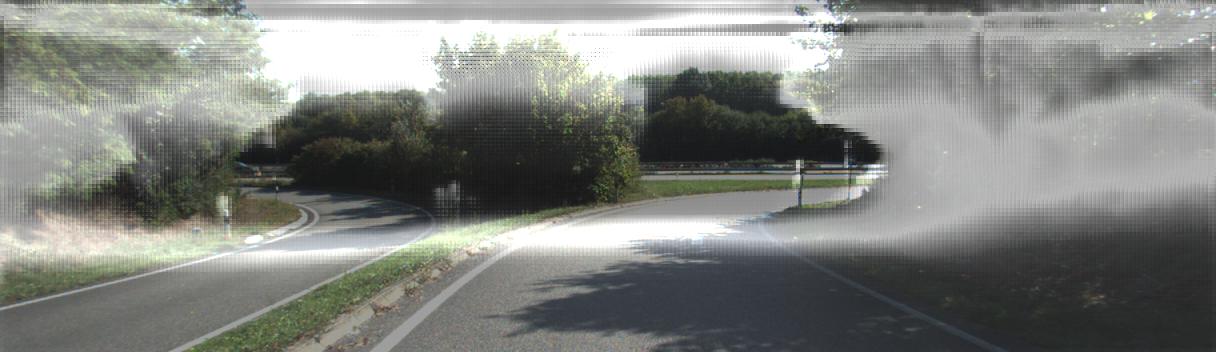}\\  
        \includegraphics[width=0.993\textwidth,height=0.5in]{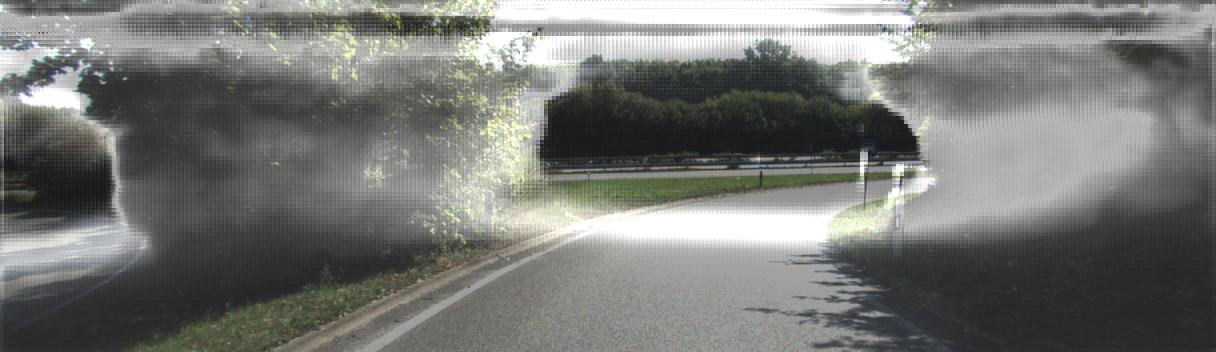}\\  
        \includegraphics[width=0.993\textwidth,height=0.5in]{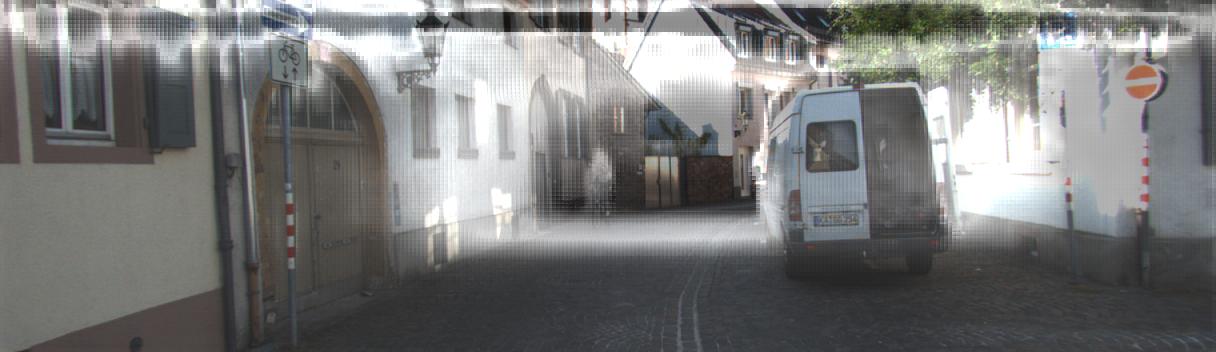}\\ 
    \end{minipage}%
}%
\subfigure{
    \begin{minipage}{0.19\linewidth}
        \centering
        \includegraphics[width=0.993\textwidth,height=0.5in]{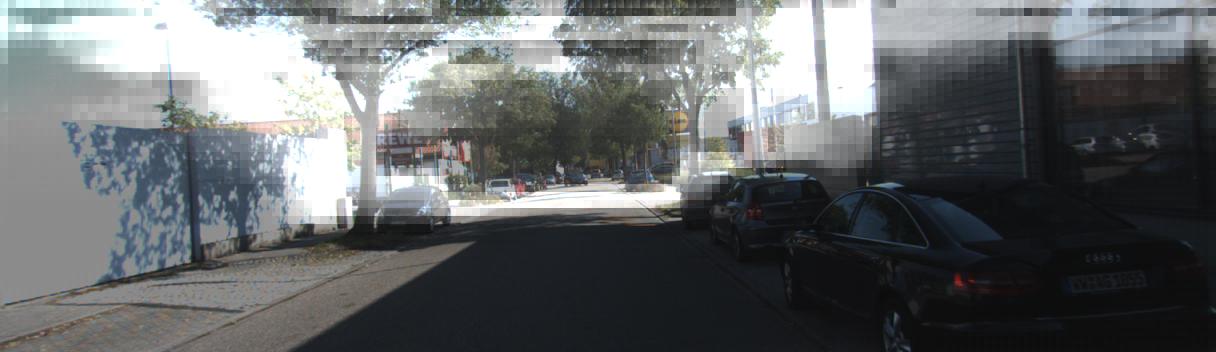}\\  
        \includegraphics[width=0.993\textwidth,height=0.5in]{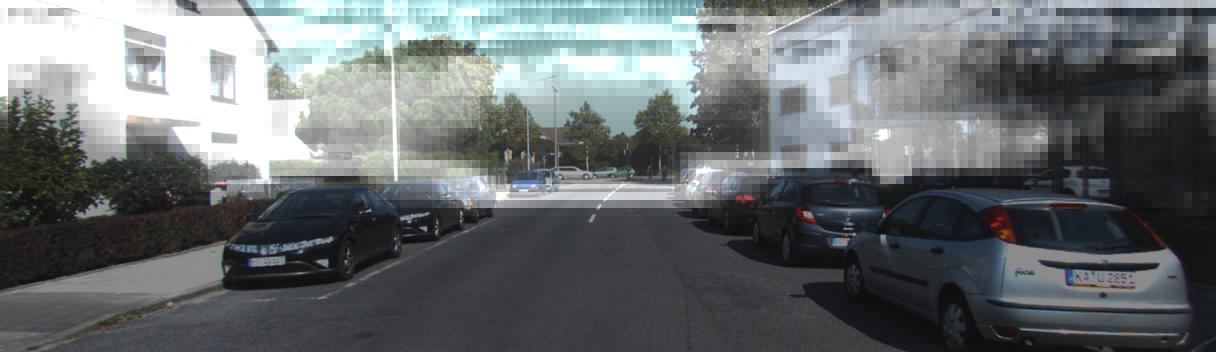}\\  
        \includegraphics[width=0.993\textwidth,height=0.5in]{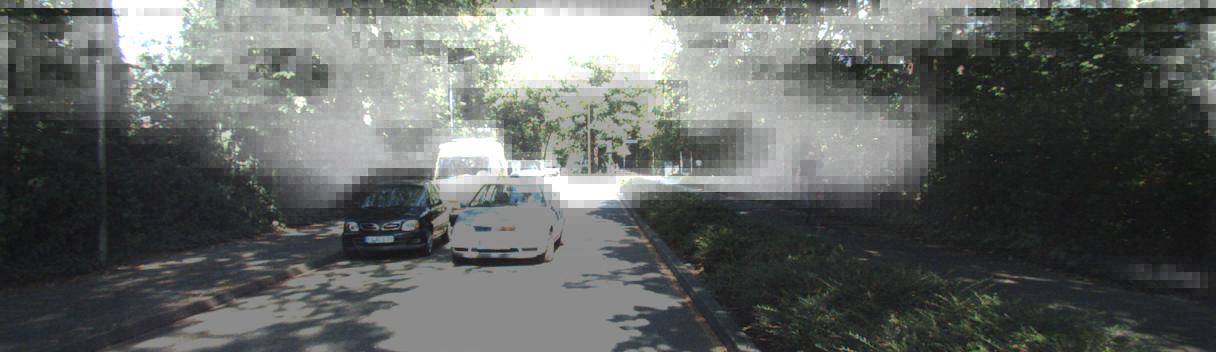}\\  
        \includegraphics[width=0.993\textwidth,height=0.5in]{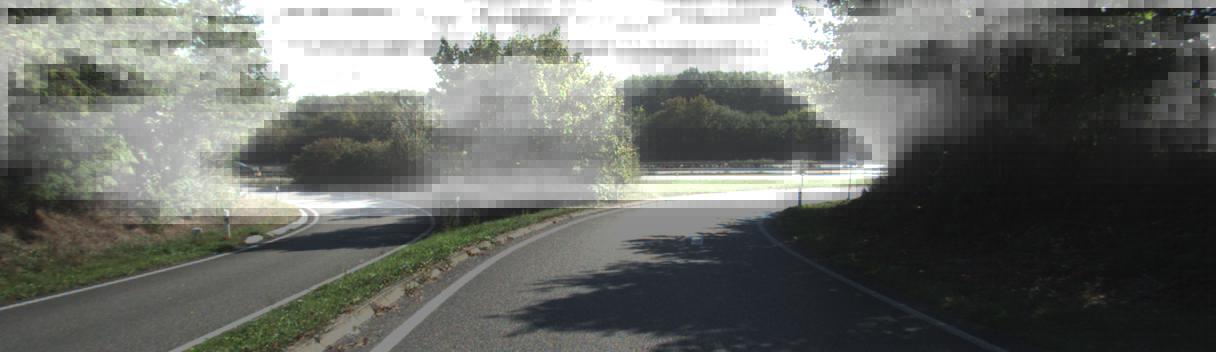}\\  
        \includegraphics[width=0.993\textwidth,height=0.5in]{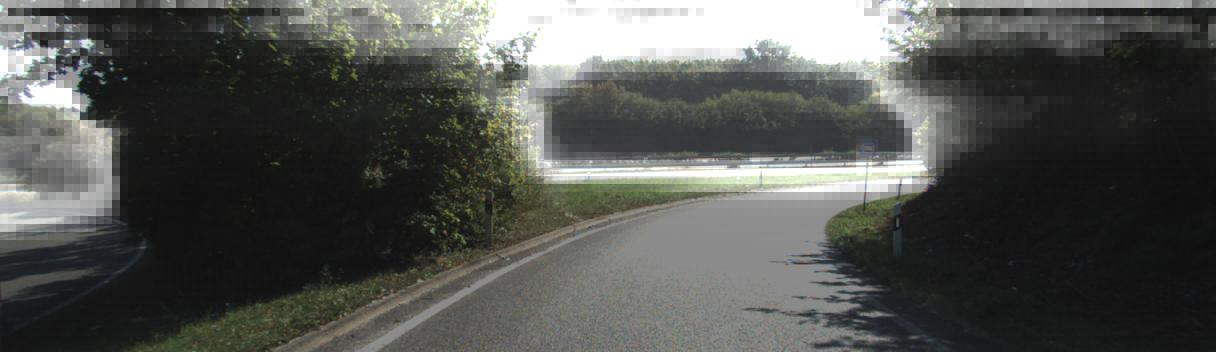}\\  
        \includegraphics[width=0.993\textwidth,height=0.5in]{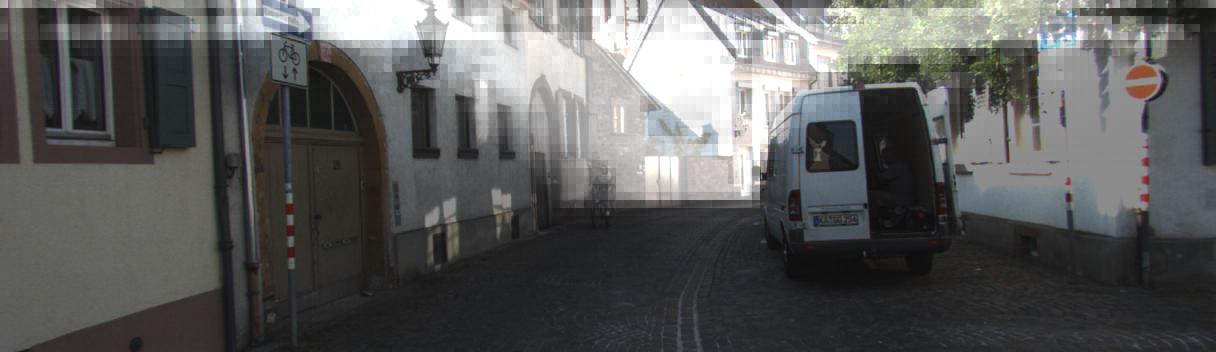}\\ 
    \end{minipage}%
}%
\centering
\caption{Qualitative structured attention examples of monocular depth prediction on the KITTI raw dataset. First column is original image and next four columns is structured attention, defined by~(\ref{eq:structured-attention}). }
\vspace{-0.2cm}
\label{fig:vis_sa_kitti}
\end{figure}

{Fig.~\ref{fig:attention} shows different visualizations regarding the learned structured attention and spatial attention on an image from the Pascal-Context dataset. The first row shows the overall structured attention tensor $\vect{a}$ as defined in~(\ref{eq:structured-attention}). The second row shows the spatial attention map of the structured tensor $\vect{m}$. While the latter seem to be spread all along the dog's body with different shapes, we observe that by optimally combining the $\vect{m}^t$ and the $\vect{v}^t$, different slices of the final structured attention tensor are able to focus on different important parts of the dog: the head, the body and the tail, thus allowing to take much more accurate pixel-level predictions for segmentation. The same phenomenon can also be found in pictures with different subjects (e.g. humans and cats). Although the visualisation result of spatial attention is noisy, corresponding structured attention would focus on useful parts as a result of the interactions between channel-wise and spatial-wise attention. }
We also provide the computed attention maps on some sample images in KITTI dataset in Fig.~\ref{fig:vis_sa_kitti}.
As expected, the final structured attention tensors manage to capture important information among different depths. 
For example, in the fourth row, structured attention focus on 
farthest frost, middle jungle, and close road.
Fig.~\ref{fig:vis_sa_kitti} proves that different structured attention tensor $\vect{a}$ can capture {distinct and representative semantic information} due to {the} combination of {both} channel- and spatial-wise attention.

\begin{figure}[!h]
\centering
    \begin{minipage}{0.24\linewidth}
        \centering
        \includegraphics[width=0.993\textwidth,height=0.5in]{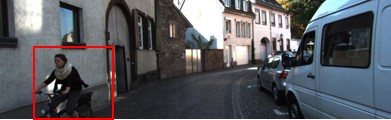}\\
        \includegraphics[width=0.993\textwidth,height=0.5in]{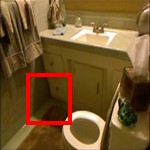}\\
        \includegraphics[width=0.993\textwidth,height=0.5in]{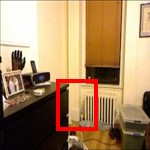}\\
        \footnotesize (a) Image
    \end{minipage}%
    \begin{minipage}{0.24\linewidth}
        \centering
        \includegraphics[width=0.993\textwidth,height=0.5in]{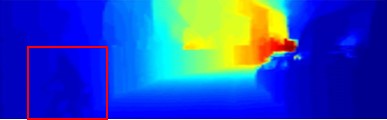}\\
        \includegraphics[width=0.993\textwidth,height=0.5in]{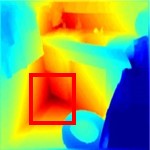}\\
        \includegraphics[width=0.993\textwidth,height=0.5in]{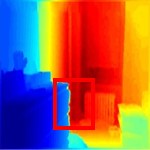}\\     
        \footnotesize (b) GT
    \end{minipage}%
    \begin{minipage}{0.24\linewidth}
        \centering
        \includegraphics[width=0.993\textwidth,height=0.5in]{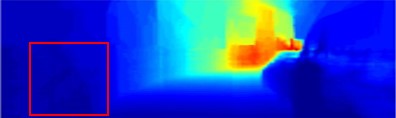}\\
        \includegraphics[width=0.993\textwidth,height=0.5in]{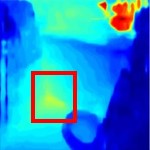}\\
        \includegraphics[width=0.993\textwidth,height=0.5in]{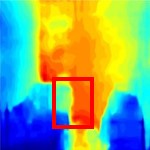}\\ 
        \footnotesize (c) DORN
    \end{minipage}%
    \begin{minipage}{0.24\linewidth}
        \centering
        \includegraphics[width=0.993\textwidth,height=0.5in]{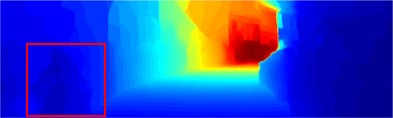}\\
        \includegraphics[width=0.993\textwidth,height=0.5in]{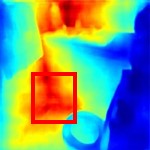}\\
        \includegraphics[width=0.993\textwidth,height=0.5in]{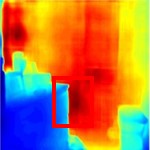}\\
        \footnotesize (d) \method
    \end{minipage}%
 
\centering
\caption{Qualitative examples on failure cases.}
\label{fig:vis_fail}
\vspace{-0.8cm}
\end{figure}

\subsection{Discussion on Failure Cases}

{We show some failure cases on both the KITTI and the NYU datasets in Figure~\ref{fig:vis_fail} to discuss our method's limitations. As can be seen the results on KITTI in the last row, the \method{} fails to provide clear structure details of the cyclist. This issue can be also observed in the DORN method. On the other hand, the failure results shown in the 2nd and the 3rd rows in Figure~\ref{fig:vis_fail} indicate that when there is a sharp corner in the picture, the \method{} cannot predict fine-grained prediction for it. These limitations we believe is mainly due to the restricted representation power of the method in explicitly modeling long-range dependencies due to the intrinsic locality of dominated convolution operations in the employed main network architecture.  
In the future, we plan to investigate Transformer~\cite{dosovitskiy2020image} to solve these limitations.}

\section{Conclusions}\label{sec:conclusion}
In this paper we proposed a novel approach to improve the learning of deep features representations for dense pixel-wise prediction tasks. Our approach seamlessly integrates a novel structured attention model within a probabilistic framework. In particular, we proposed to structure the attention tensors as the sum of $T$ rank tensors, each being the tensor-product of a spatial attention map and a channel attention vector. These two kinds of variables are jointly learned within the probabilistic formulation made tractable thanks to the variational approximation. The proposed structured attention is rich enough to capture complex spatial- and channel-level inter-dependencies, while being efficient to compute. The overall optimisation of the probabilistic model and of the CNN front-end is performed jointly. Extensive experimental evaluations show that \method~outperforms state-of-the-art methods on several datasets, thus confirming the importance of {jointly} structuring the {spatial- and channel-wise} attention variables for {learning effective deep representations for} dense pixel-level prediction tasks. Future works might deal with the relationship with deep unsupervised probabilistic models also used for data fusion~\cite{sadeghi2020audio,sadeghi2021mixture,sadeghi2021switching}.


%


\ifCLASSOPTIONcompsoc
  \section*{Acknowledgments}
\else
  \section*{Acknowledgment}
\fi


This research is supported in part
by China Scholarship Council (CSC) during a visit of Guanglei Yang to University of Trento, the Early Career Scheme of the Research Grants Council (RGC) of the Hong
Kong SAR under grant No.~26202321, HKUST Startup Fund No.~R9253, and by the European Commission through the H2020 SPRING project under GA No.~871245. This  work  has  been  also  supported  by Caritro Foundation and the Deep Learning lab of the ProM Facility.

\ifCLASSOPTIONcaptionsoff
  \newpage
\fi



\bibliographystyle{IEEEtran}
\bibliography{ref}

%

\vspace{-1.2cm}
\begin{IEEEbiography}[{\includegraphics[width=1in,height=1.25in,clip,keepaspectratio]{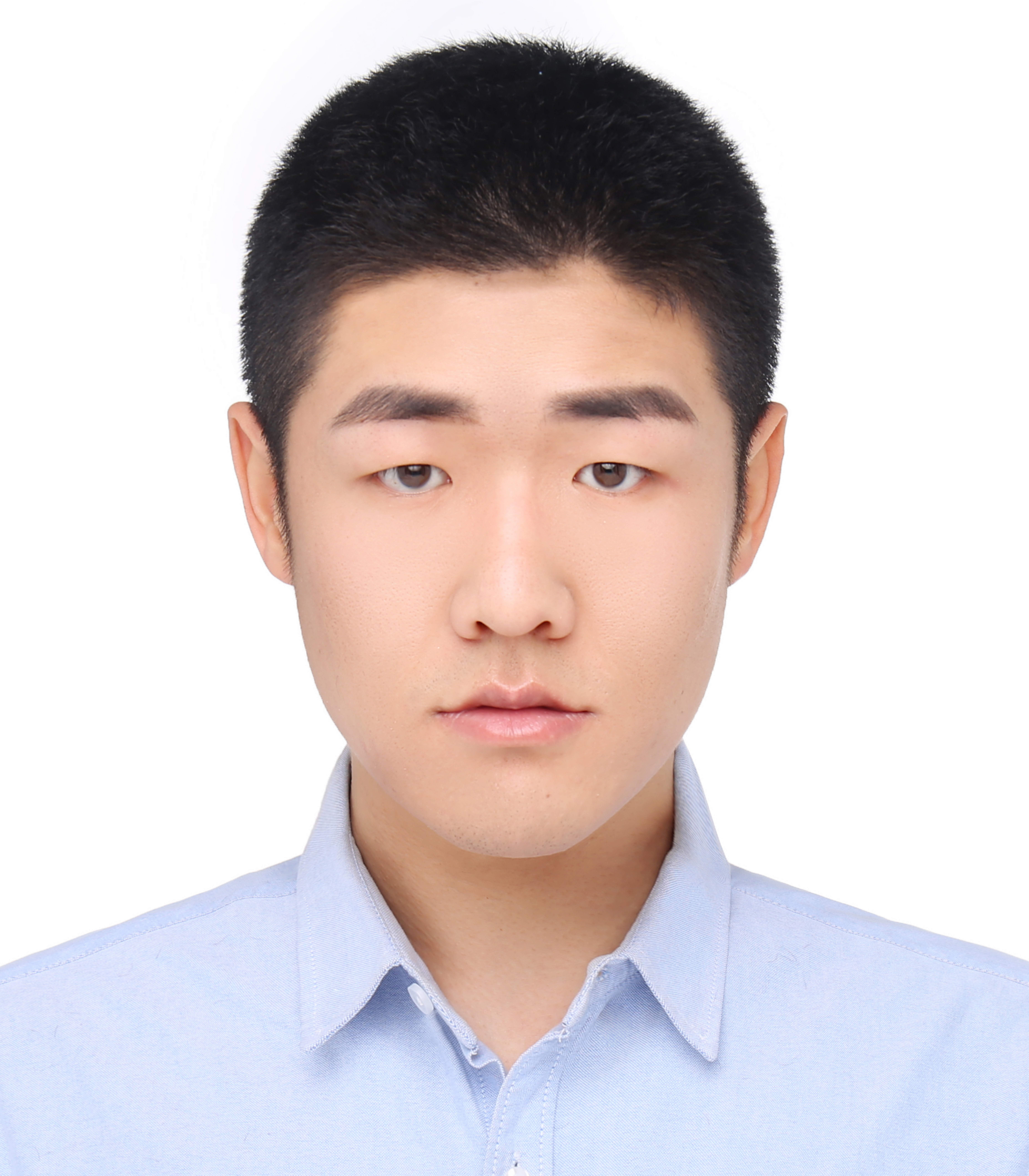}}]{Guanglei Yang}
received the B.S. degree in instrument science and technology from Harbin Institute of Technology (HIT), Harbin, China, in 2016. He is currently pursuing the Ph.D degree in the School of Instrumentation Science and Engineering, Harbin Institute of Technology(HIT), Harbin, China. He is working at University of Trento as a visiting student from 2020 to now. His research interests mainly include domain adaption, pixel-level prediction and attention gate.
\end{IEEEbiography}

\vspace{-0.5cm}

\begin{IEEEbiography}[{\includegraphics[width=1in,height=1.25in,clip,keepaspectratio]{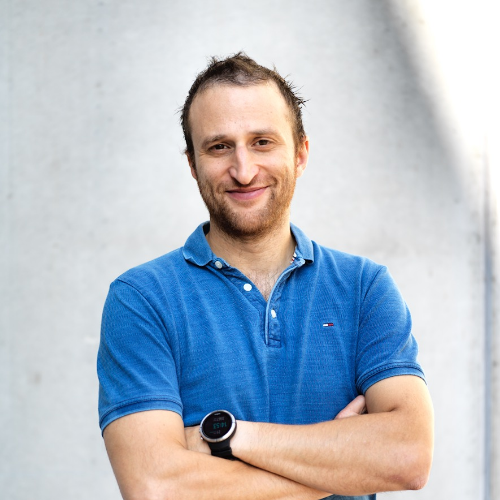}}]
{Paolo Rota}
is an assistant professor (RTDa) at University of Trento (in the MHUG group), working on computer vision and machine learning. He received his PhD in Information and Communication Technologies from the University of Trento in 2015. Prior joining UniTN he worked as Post-doc at the TU Wien and at the Italian Institute of Technology (IIT) of Genova. He is also collaborating with the ProM facility of Rovereto on assisting companies in inserting machine learning in their production chain.
\end{IEEEbiography}

\vspace{-0.5cm}

\begin{IEEEbiography}[{\includegraphics[width=1in,height=1.25in,clip,keepaspectratio]{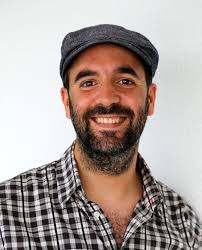}}]
{Xavier Alameda-Pineda} is a Research Scientist at Inria, and the Leader of the RobotLearn Team. He obtained the M.Sc. (equivalent) in Mathematics in 2008, in Telecommunications in 2009 from BarcelonaTech and in Computer Science in 2010 from Université Grenoble-Alpes (UGA). He the worked towards his Ph.D. in Mathematics and Computer Science, and obtained it 2013, from UGA. After a two-year post-doc period at the Multimodal Human Understanding Group, at University of Trento, he was appointed with his current position. Xavier is an active member of SIGMM, a senior member of IEEE and a member of ELLIS. He is the Coordinator of the H2020 Project SPRING: Socially Pertinent Robots in Gerontological Healthcare and is co-chair at the Multidisciplinary Institute of Artificial Intelligence. Xavier’s research interests are at the cross-roads of machine learning, computer vision and audio processing for scene and behavior analysis and human-robot interaction. He served as Area Chair at ICCV’17, of ICIAP’19 and of ACM
MM’19, MM'20 and MM'21 as well as AAAI'21. He will serve as Program Co-Chair at ACM MM'22. He is the recipient of several paper awards and of the ACM SIGMM Rising Star Award in 2018.
\end{IEEEbiography}

\vspace{-0.5cm}

\begin{IEEEbiography}[{\includegraphics[width=1in,height=1.25in,clip,keepaspectratio]{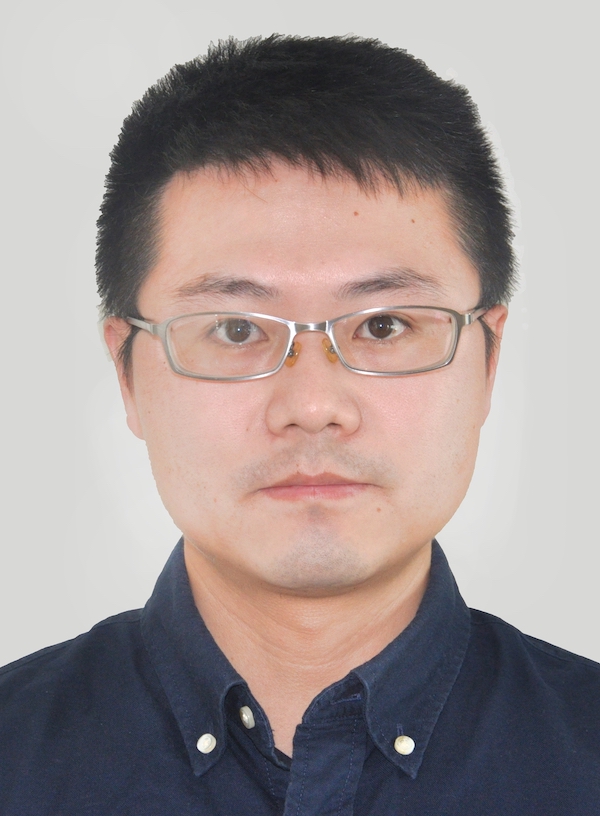}}]
{Dan Xu}
is an Assistant Professor in the Department of Computer Science and Engineering at HKUST. He was a Postdoctoral Research Fellow in VGG at the University of Oxford. He was a Ph.D. student in the Department of Computer Science at the University of Trento. He was also a research assistant of MM Lab at the Chinese University of Hong Kong. He received the best scientific paper award at ICPR 2016, and a Best Paper Nominee at ACM MM 2018. He served as Senior Programme Committee/Area Chair at multiple international conferences including AAAI 2021, ACM MM 2020, 2021, WACV 2021, and ICPR 2020.
\end{IEEEbiography}

\vspace{-0.5cm}
\begin{IEEEbiography}[{\includegraphics[width=1in,height=1.25in,clip,keepaspectratio]{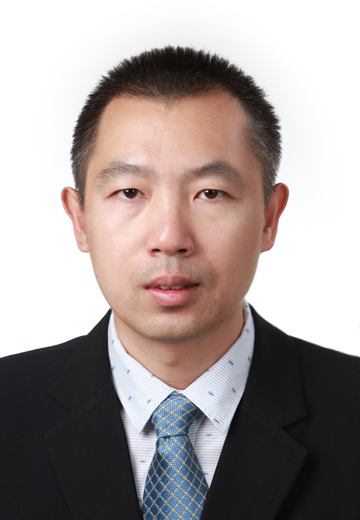}}]{Mingli Ding}
received the B.S., M.S. and Ph.D. degrees in instrument science and technology from Harbin Institute of Technology (HIT), Harbin, China, in 1996, 1997 and 2001, respectively. He worked as a visiting scholar in France from 2009 to 2010. Currently, he is a professor in the  School  of  Instrumentation Science and Engineering at Harbin Institute of Technology. Prof. Ding’s research interests are intelligence tests and information processing, automation test technology, computer vision, and machine learning. He has published over 40 papers in peer-reviewed journals and conferences.
\end{IEEEbiography}

\vspace{-15.5cm}

\begin{IEEEbiography}[{\includegraphics[width=1in,height=1.25in,clip,keepaspectratio]{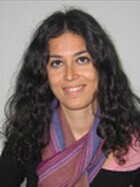}}]
{Elisa Ricci}
received the M.S. (2004) and PhD degree (2008) from the
University of Perugia. She is an associate
professor at the University of Trento and a head of research unit at Fondazione Bruno Kessler. Previously, she
was a post-doctoral researcher at
Idiap research institute and Fondazione Bruno Kessler and a visiting researcher at the
University of Bristol. She
received the Honorable mention award at ICCV
2021 and the Best Paper Award at ACM MM 2021. Her research interests are
mainly in the areas of computer vision and
deep learning. She is an ELLIS fellow. 
\end{IEEEbiography}




\end{document}